%% file: neurips_2026_itr.tex
\documentclass{article}

\PassOptionsToPackage{numbers,sort&compress}{natbib}
\usepackage[main,final]{neurips_2026}

\usepackage[utf8]{inputenc}
\usepackage[T1]{fontenc}
\usepackage{hyperref}
\usepackage{url}
\usepackage{booktabs}
\usepackage{amsfonts}
\usepackage{nicefrac}
\usepackage{microtype}
\usepackage{xcolor}
\usepackage{amsmath,amssymb,amsthm,mathtools}
\usepackage{fix-cm}
\DeclareMathSizes{7.5}{7.5}{5.5}{5}
\DeclareMathSizes{7}{7}{5}{5}
\usepackage{bm}
\usepackage{graphicx}
\usepackage{subcaption}
\usepackage{multirow}
\usepackage{enumitem}
\usepackage{algorithm}
\usepackage{algpseudocode}

\title{Distribution-First Framework for Learning Risk-Sensitive Individualized Treatment Rules}

\author{%
  Zhen Fang \\
Department of Statistics and Operations Research \\
University of North Carolina at Chapel Hill \\
\texttt{zhenfang@unc.edu}
  \And
  Guanting Chen \\
  Department of Statistics and Operations Research \\
  University of North Carolina at Chapel Hill \\
  \texttt{guanting@unc.edu}
  \And
  Yufeng Liu \\
  Department of Statistics \\
  University of Michigan \\
  \texttt{yufliu@umich.edu}
}

\newtheorem{assumption}{Assumption}
\newtheorem{theorem}{Theorem}
\newtheorem{corollary}{Corollary}
\newtheorem{remark}{Remark}
\newtheorem{lemma}{Lemma}
\newtheorem{proposition}{Proposition}

\newcommand{\E}{\mathbb{E}}
\newcommand{\Pp}{\mathbb{P}}
\newcommand{\R}{\mathbb{R}}
\newcommand{\ind}{\mathbf{1}}
\newcommand{\X}{\mathcal{X}}
\newcommand{\A}{\mathcal{A}}

\newcommand{\Fhat}{\widehat{F}}
\newcommand{\Tt}{\mathcal{T}}
\newcommand{\Q}{Q}
\newcommand{\D}{\mathcal{D}}

\newcommand{\Qhat}{\widehat{Q}}
\newcommand{\Reg}{\mathcal{R}}
\newcommand{\pihat}{\widehat{\pi}}

\newcommand{\cvar}{\mathrm{CVaR}}
\newcommand{\var}{\mathrm{Var}}
\newcommand{\abs}[1]{\left|#1\right|}
\newcommand{\1}{\mathbf 1}
\newcommand{\op}{O_{\Pp}}
\newcommand{\norm}[1]{\left\lVert #1\right\rVert}
\newcommand{\itrtablecell}[2]{#1\hspace{0.8pt}(#2)}
\newcommand{\bestcell}[2]{\textbf{#1}\hspace{0.8pt}(#2)}

\begin{document}

\maketitle

\begin{abstract}
Learning a risk-sensitive individualized treatment rule (ITR) with patient covariates is challenging when the treatment criterion depends on the full conditional outcome distribution rather than only its conditional mean. Existing methods are largely criterion-specific, making it difficult to compare multiple treatment criteria within a common pipeline, such as quantiles,
conditional value-at-risk (CVaR), and complex preference-based objectives such as cumulative prospect theory. We propose a distribution-first framework for risk-sensitive ITR learning that first estimates treatment-specific conditional outcome distributions and then constructs the target ITR by plug-in criterion evaluation. The proposed framework unifies treatment rule learning across multiple criteria and accommodates hard-to-optimize objectives within the same estimation pipeline. On the theoretical side, it yields a general error propagation result for locally H\"older-stable criteria, linking first-stage distribution estimation error to downstream criterion estimation and treatment regret, together with sharper regret bounds under standard margin conditions. Empirically, we show that the proposed framework outperforms direct criterion-specific methods in multiple settings and further evaluate it on the ACTG175 HIV clinical trial, a randomized study comparing several antiretroviral regimens in HIV-infected adults.
\end{abstract}

\section{Introduction}

Personalized decision making is a central goal in precision medicine, aiming to move beyond the traditional ``one size fits all'' paradigm and tailor treatment decisions to patient-specific characteristics \citep{kosorok2019precision}. 
A core statistical approach to this problem is learning an individualized treatment rule (ITR), which maps patient covariates to treatment recommendations to optimize a clinically meaningful objective. 
Over the past two decades, a rich literature has developed around optimal ITR learning, including regression-based and value-based approaches that seek treatment rules with the best expected outcome, leading to substantial progress in individualized treatment learning \citep{murphy2003optimal,qian2011performance,zhao2012estimating,zhou2017residual,qi2020multi}.

In many treatment decisions, however, the conditional mean alone does not fully reflect what matters clinically. 
Two treatments may have similar expected outcomes but substantially different lower-tail behavior, variability, or asymmetry, leading to different risk--benefit profiles for different patients.
This limitation has motivated growing interest in risk-sensitive treatment criteria that depend on distributional features beyond the conditional mean, such as conditional quantiles and conditional value-at-risk (CVaR), which are better suited to capture downside risk and tail behavior \citep{wang2018quantile,xiao2019robust,qi2023robustness}. 
More broadly, complex risk-sensitive criteria like Cumulative Prospect Theory (CPT) have also been recognized as meaningful in health-related decision making \citep{attema2013prospect,lipman2019corrective}. Originating in economics, CPT combines a value function with probability distortion to represent asymmetric preferences over gains and losses together with nonlinear attitudes toward unlikely events \citep{tversky1992advances}.

At the same time, existing risk-sensitive ITR methods face two related challenges. 
First, they are typically criterion-specific: a method designed for one treatment criterion, such as a conditional quantile or CVaR objective, generally does not transfer directly to another. 
As a result, adapting treatment recommendations to different patient preferences may require redesigning and retraining a separate learning procedure for each new criterion, which can substantially increase modeling and computational cost in practice. 
Second, direct criterion-specific learning becomes especially difficult for more complex risk-sensitive objectives. 
This is particularly clear for CPT-type criteria, whose functional form involves nonlinear transformations of cumulative probabilities rather than a simple expectation of a pointwise loss. 
Related evidence also appears in other decision-learning settings, such as reinforcement learning and agents \citep{prashanth2022risk,huang2022offpolicy,barakat2025global,smith2025risk}. 
Taken together, these observations suggest that the challenge is broader than any single criterion and arises more generally when treatment learning is tied to a specific risk-sensitive objective.

{\bf Problem description.} These considerations motivate the main question of this paper: \emph{how can we learn risk-sensitive ITRs in a unified and flexible way when the relevant treatment criterion depends on the full conditional outcome distribution rather than only its mean?} 
The challenge is not only to optimize a single pre-specified risk-sensitive criterion, but to develop a general approach that can support multiple treatment criteria and remain applicable when the target objective is difficult to optimize directly. 
Such a formulation would make it possible to compare treatment rules under different risk-sensitive objectives and to accommodate more complex criteria beyond the standard conditional mean-based setting.

{\bf Related work and our method.} 
Our goal is to develop a unified framework for risk-sensitive ITR learning under the potential outcome framework. A natural starting point is that many treatment criteria of interest can be written as functionals of the conditional outcome distribution given patient covariates. This suggests a distribution-first strategy: first estimate the treatment-specific conditional outcome distribution, then obtain the target criterion by plug-in evaluation, and finally construct the treatment rule by comparing criterion values across treatments. This perspective is especially appealing when one wishes to compare treatment rules under different criteria or handle objectives that are difficult to optimize directly.

This view is supported by a broad literature on conditional distribution estimation. Classical approaches include kernel-based conditional distribution estimation  \citep{hall1999nonparametric}, while more recent work has developed random forest-based methods for conditional quantile and distribution learning \citep{meinshausen2006quantile,athey2019generalized,cevid2022distributional}. There is also growing interest in modern generative approaches to conditional sampling and distribution learning, including deep generative methods such as GCDS \citep{zhou2023deep} and related neural density estimators \citep{liu2021density}. Diffusion-based generative models, such as denoising diffusion probabilistic models (DDPMs) \citep{ho2020denoising}, offer another flexible approach to distribution learning.  Recent work has further extended diffusion models to causal inference, such as DiffPO \citep{ma2024diffpo}, which models potential outcome distributions while addressing treatment selection bias. Together, these methods provide increasingly rich options for estimating treatment-specific conditional outcome distributions.

To the best of our knowledge, unlike criterion-specific methods \citep{murphy2003optimal,qian2011performance,zhao2012estimating,zhou2017residual,qi2020multi}, there has been no work that explicitly proposes a unified framework for risk-sensitive ITR learning. The work closest to ours appears in offline policy assessment, where \citet{huang2021offpolicy} studied plug-in estimation of Lipschitz criteria by first estimating the target policy's outcome CDF and then analyzing how CDF estimation error propagates to criterion estimation error. However, their setting differs from ours in several aspects. First, they estimate the policy-induced marginal outcome distribution for policy assessment, whereas we focus on treatment- and covariate-specific conditional outcome distributions. Second, our goal is treatment recommendation, which requires treatment rule comparison and regret analysis rather than criterion estimation alone. Our work builds on this perspective but shifts the focus from criterion estimation to treatment recommendation. Another related work is \citet{naf2024causal}, which develops a distributional forest-based estimator of conditional treatment effects. Its focus is distributional treatment effect estimation and inference rather than criterion-specific treatment rule construction.


{\bf Contributions.} Our main contributions are as follows. 
(1) We formulate a distribution-first framework for risk-sensitive ITR learning to provide a unified way to learn treatment rules under multiple criteria and accommodate complex objectives that are difficult to handle by direct criterion-specific optimization. 
(2) We establish a theoretical analysis for this framework by linking  the first-stage distribution estimation error to downstream criterion estimation and treatment regret in the ITR setting. 
(3) We empirically show that the proposed framework outperforms direct criterion-specific methods in multiple settings, and we further evaluate the framework on a real data application to illustrate a complete pipeline.

\section{Distribution-First Risk-Sensitive ITR Learning}
\label{sec:method}

We observe i.i.d.\ samples $\{(X_i,A_i,Y_i)\}_{i=1}^n$, where $X_i\in\X\subseteq\R^p$ is a vector of covariates, $A_i\in\A=\{-1,+1\}$ is the treatment assignment, and $Y_i\in\R$ is the observed outcome, with larger values preferred. Let $Y_i(a)$ denote the potential outcome under treatment $a\in\A$. We make the following standard identification assumptions\citep{rubin1974estimating}.

\begin{assumption}[Identification]
\label{ass:identification}
For each $a\in\A$: (i) \emph{consistency}, $Y=Y(A)$; (ii) \emph{unconfoundedness}, $Y(a)\perp A\mid X$; and (iii) \emph{positivity}, there exists $\eta>0$ such that
$
\eta \le \Pp(A=a\mid X=x)\le 1-\eta,\ \forall x\in\X$.
\end{assumption}

Under Assumption~\ref{ass:identification}, the treatment-specific conditional distribution of the potential outcome is identified by
\(
F_a(y\mid x)
=
\Pp\{Y(a)\le y\mid X=x\}
=
\Pp\{Y\le y\mid X=x,A=a\}\).
Let $\Tt$ be a real-valued criterion functional on distributions, with larger values preferred. We define the treatment-specific score
$Q_a^{\Tt}(x)=\Tt\{F_a(\cdot\mid x)\}$,
and the corresponding optimal individualized treatment rule
$\pi_{\Tt}(x)\in\arg\max_{a\in\A} Q_a^{\Tt}(x)$.
This formulation makes clear that the target of learning is the full treatment-specific conditional distribution, rather than only the conditional mean, whenever treatment choice depends on distributional features beyond expectation.

Our distribution-first framework proceeds in two stages shown in Algorithm \ref{alg:df}. In Stage 1, we estimate the treatment-specific conditional distributions $F_a(\cdot\mid x)$ for $a\in\A$. Let $\Fhat_a(\cdot\mid x)$ denote a generic first-stage estimator. The framework is agnostic to the particular conditional distribution learner; in our experiments, we consider kernel-based estimators, random forest-based estimators, and modern generative models, which are discussed in Section~\ref{sec:estimators}. In Stage 2, we evaluate the target treatment criterion by plug-in:
$
\Qhat_a^{\Tt}(x)=\Tt\{\Fhat_a(\cdot\mid x)\}$,
and return the estimated treatment rule
$\pihat_{\Tt}(x)\in\arg\max_{a\in\A}\Qhat_a^{\Tt}(x)$.
This modular design allows the same first-stage estimate to support multiple downstream criteria without redesigning the learning procedure for each objective.

\begin{algorithm}[thb]
\caption{Distribution-First Risk-Sensitive ITR Learning}
\label{alg:df}
\begin{algorithmic}[1]
\Require Data $\{(X_i,A_i,Y_i)\}_{i=1}^n$, a criterion functional $\Tt$, and a first-stage conditional distribution learner
\State Estimate treatment-specific conditional distributions $\Fhat_a(\cdot\mid x)$ for each $a\in\A$
\State Compute plug-in scores $\Qhat_a^{\Tt}(x)=\Tt\{\Fhat_a(\cdot\mid x)\}$
\State Return $\pihat_{\Tt}(x)\in\arg\max_{a\in\A}\Qhat_a^{\Tt}(x)$
\end{algorithmic}
\end{algorithm}

\subsection{Treatment criteria}
\label{subsec:criteria}
A main advantage of the proposed framework is that different treatment criteria can be evaluated from the same estimated conditional distribution. These criteria correspond to different notions of utility, risk sensitivity, and personal preference. In this paper, we focus on four representative criteria: the conditional mean, the conditional quantile, CVaR, and CPT.

{\bf Conditional mean.}
The mean criterion is defined by
$\Tt_{\mathrm{mean}}\{F_a(\cdot\mid x)\}
=
\E\{Y(a)\mid X=x\}$.
This is the standard objective in much of the classical ITR literature. It corresponds to risk-neutral decision making, in the sense that treatments are compared solely through their average outcomes. While natural and interpretable, the conditional mean criterion ignores variability, tail behavior, and asymmetry of the outcome distribution.

{\bf Conditional quantile.}
For $\tau\in(0,1)$, let
$Q_{\tau}\{Y(a)\mid X=x\}
=
\inf\{y:F_a(y\mid x)\ge \tau\}$
denote the conditional $\tau$-quantile under treatment $a$. The corresponding criterion is
$\Tt_{\mathrm{Q},\tau}\{F_a(\cdot\mid x)\}
=
Q_{\tau}\{Y(a)\mid X=x\}$.
Quantile-based decision making focuses on a specific location of the outcome distribution rather than its average. When $\tau$ is small, the criterion emphasizes lower-tail performance and yields a more conservative treatment rule.

{\bf Conditional value-at-risk (CVaR).}
For $\alpha\in(0,1)$, we define the lower-tail CVaR by
$\cvar_{\alpha}\{Y(a)\mid X=x\}
=\frac{1}{\alpha}\int_0^\alpha Q_u\{Y(a)\mid X=x\}\,du$.
The associated criterion is
$\Tt_{\mathrm{CVaR},\alpha}\{F_a(\cdot\mid x)\}
=\cvar_{\alpha}\{Y(a)\mid X=x\}$.
CVaR summarizes the average outcome over the worst $\alpha$ fraction of the conditional distribution. Compared with a single lower quantile, it captures the overall severity of adverse outcomes in the lower tail and is therefore a more informative measure of downside risk. In personalized treatment learning, CVaR is particularly relevant when severe but relatively rare unfavorable outcomes must be explicitly controlled.

{\bf Cumulative prospect theory (CPT).}
Unlike the previous criteria, CPT is designed to encode personal preferences under risk, including asymmetric attitudes toward gains and losses and nonlinear perception of probabilities. Let $r(x)$ be a reference point and define the centered outcome
$Z(a)=Y(a)-r(x)$.
We use the value function
\begin{equation*}
u(z)=
\begin{cases}
u^+(z)=z^{\eta_+}, & z\ge 0,\\[3pt]
-u^-(-z)=-\lambda(-z)^{\eta_-}, & z<0,
\end{cases}
\end{equation*}
where $\lambda>1$ captures loss aversion, and $\eta_+,\eta_-\in(0,1]$ control diminishing sensitivity over gains and losses. Let $G_a(\cdot\mid x)$ denote the conditional distribution function of $Z(a)\mid X=x$. Given probability-weighting functions $w^+$ and $w^-$ for gains and losses, the CPT criterion is
\begin{equation*}
\Tt_{\mathrm{CPT}}\{F_a(\cdot\mid x)\}
=
\int_0^\infty w^+\!\left(1-G_a(z\mid x)\right)\, d u^+(z)
-
\int_0^\infty w^-\!\left(G_a(-z\mid x)\right)\, d u^-(z).
\end{equation*}
CPT is useful when treatment choice should reflect not only statistical risk but also behavioral preference. Relative to above criteria, CPT can represent richer patient-specific attitudes, such as stronger aversion to losses than attraction to gains, or overweighting of small probabilities. This makes CPT particularly appealing for personalized medicine when patients or clinicians care about more nuanced risk--reward tradeoffs.


\subsection{First-stage conditional distribution estimation}
\label{sec:estimators}

Our distribution-first framework is modular. 
Instead of designing a separate learner for each decision criterion, we first estimate the treatment-specific conditional outcome distribution and then evaluate downstream functionals by plug-in. 

We consider three broad classes of first-stage estimators: kernel-based methods, random forest-based methods, and generative models. 
These three estimator classes represent distinct modeling principles for first-stage distribution learning. 
Kernel methods emphasize local smoothness and interpretability, random forest-based methods emphasize adaptive partitioning and robustness in heterogeneous tabular data, and generative models emphasize flexibility for complex conditional distributions. By studying all three within the same framework, we can assess whether the downstream gains of distribution-first decision learning are robust across substantially different first-stage estimators.

{\bf Kernel-based estimators.}
Kernel methods estimate the conditional distribution by locally weighting observations near the target covariate value. 
For treatment arm $a$, a generic estimator is
\[
\widehat{F}_a(y\mid x)
=
\frac{\sum_{i=1}^n K_h(X_i-x)\,\ind(A_i=a)\,\ind(Y_i\le y)}
{\sum_{i=1}^n K_h(X_i-x)\,\ind(A_i=a)},
\]
with an analogous density estimator obtained by additional smoothing in $y$. 

The main appeal of kernel methods is their simplicity and interpretability \citep{hall1999nonparametric}. 
They impose relatively weak structural assumptions and are natural when the conditional distribution varies smoothly with covariates, making them a useful baseline for distribution-first learning. 
Their main limitation is that performance can degrade quickly in moderate or high dimensions, where local neighborhoods become sparse and estimation becomes sensitive to the choice of metric and bandwidth.

{\bf Random Forest-based estimators.}
Forest-based methods replace fixed metric neighborhoods with data-adaptive neighborhoods learned through recursive partitioning. 
For a treatment-specific fit, or a joint fit whose weights are zero
outside the queried treatment arm, the fitted forest induces nonnegative
weights $\alpha_i^{(a)}(x)$ satisfying
$\sum_{i:A_i=a}\alpha_i^{(a)}(x)=1$. The resulting estimator is
\[
\widehat{F}_a(y\mid x)
=
\sum_{i:A_i=a}
\alpha_i^{(a)}(x)\,\ind(Y_i\le y).
\]
This formulation is closely related to distributional random forests and their causal extensions \citep{breiman2001random,meinshausen2006quantile,cevid2022distributional}. 
The main advantage of this class is adaptivity. Forest methods can capture nonlinearities, interactions, and heterogeneous structure without relying on a fixed distance in the ambient covariate space. 
This makes them especially attractive for tabular settings with moderate or high-dimensional covariates, where only a subset of variables may be relevant. 
In addition, the learned weights provide a convenient representation of the full conditional distribution, which aligns naturally with our plug-in framework.

{\bf Generative models.}
A more flexible alternative is to model $Y(a)\mid X=x$ generatively and sample
$\widetilde{Y}_{a,1}(x),\ldots,\widetilde{Y}_{a,M}(x)\sim \widehat{P}\{Y(a)\mid X=x\}$,
then construct an empirical distribution. This class includes GANs, normalizing flows, variational models, and conditional diffusion models \citep{goodfellow2014generative,kingma2013auto,papamakarios2021normalizing,ho2020denoising,song2021scorebased,kotelnikov2023tabddpm,ma2024diffpo,tang2025conditional}. The main strength of this class is expressive power. Generative models can in principle represent complex non-Gaussian, skewed, heavy-tailed, or multimodal conditional outcome distributions that may be difficult to capture with local smoothing or recursive partitioning. 
This makes them a natural candidate when the outcome law has substantial distributional complexity. 
At the same time, they typically require heavier computation, more tuning, and larger effective sample sizes for stable training.

A key point is that the best estimator for global distributional fit need not be the best estimator for downstream decision quality under a particular criterion. For example, a method that fits the center of the distribution well may still perform poorly for tail-sensitive objectives if it misses lower-tail behavior. For this reason, we evaluate first-stage estimators primarily through the treatment rules they induce, rather than viewing conditional distribution estimation as an end in itself.

\section{Theoretical Properties}
\label{sec:theory}

We establish a general error propagation guarantee for the proposed framework, connecting first-stage conditional distribution estimation error to downstream criterion estimation error and, ultimately, to treatment policy regret. Our focus is on a broad class of H\"older-continuous functionals of the treatment-specific conditional outcome distribution. This formulation covers the main criteria considered in this work under criterion-specific regularity conditions, including the conditional mean, quantile, mean-variance, CVaR, and CPT.

For any treatment rule $\pi$, define
$V_{\mathcal T}(\pi)
=
\E\!\left[ Q_{\pi(X)}^{\mathcal T}(X)\right],\
\mathcal R_{\mathcal T}(\pi)
=
V_{\mathcal T}(\pi_{\mathcal T})-V_{\mathcal T}(\pi)$.
Define $
e_n^{\mathcal T}(x)
=
\max_{a\in\mathcal A}
|\widehat Q_a^{\mathcal T}(x)-Q_a^{\mathcal T}(x)|$,
and
$\delta_n^{\mathcal T}(x)
=
\max_a d_{\mathcal T}
\{\widehat F_a(\cdot\mid x),F_a(\cdot\mid x)\}$. So correspondingly, we have $e_{n,\infty}^{\mathcal T}
=
\sup_{x\in\mathcal X}e_n^{\mathcal T}(x)$ and $\delta_{n,\infty}^{\mathcal T}
=\sup_{x\in\mathcal X}\delta_n^{\mathcal T}(x)$.
We first characterize how perturbations of the estimated conditional distribution propagate to the criterion value.

\begin{assumption}[H\"older continuity]
\label{ass:holder}
There exist \(L_{\mathcal T}>0\), \(\epsilon_{\mathcal T}>0\), and
\(0<\rho_{\mathcal T}\le1\) such that, uniformly over \(a\) and \(x\),
for every distribution \(G\) satisfying
\(d_{\mathcal T}\{G,F_a(\cdot\mid x)\}\le\epsilon_{\mathcal T},\) we have
\[
|\mathcal T(G)-\mathcal T\{F_a(\cdot\mid x)\}|
\le
L_{\mathcal T}
d_{\mathcal T}\{G,F_a(\cdot\mid x)\}^{\rho_{\mathcal T}}.
\]
\end{assumption}

Here, $L_{\mathcal{T}}$ is a criterion-specific stability constant, $\epsilon_{\mathcal{T}}$ specifies the neighborhood over which the stability condition holds, while $\rho_{\mathcal{T}}$ is the H\"older exponent characterizing the sensitivity of the criterion to local distributional perturbations. 
When $\rho_{\mathcal{T}}=1$, the condition gives local Lipschitz stability, related to the CDF-based Lipschitz formulation of \citet{huang2021offpolicy}. 
Allowing $\rho_{\mathcal{T}}<1$ extends the framework to non-Lipschitz distributional criteria, including the CPT specification considered in this work. 
The semimetric $d_{\mathcal{T}}$ may be chosen according to the criterion and the underlying distribution class, and need not be common across different criteria. Verification of this condition for the criteria considered in this paper is provided in Appendix~\ref{app:holder-verify}.

Define the H\"older-induced score-error bound $\bar e_n^{\mathcal T}
    :=
    L_{\mathcal T}
    \bigl(\delta_{n,\infty}^{\mathcal T}\bigr)^{\rho_{\mathcal T}}$.
Under Assumption~\ref{ass:holder}, whenever
\(\delta_{n,\infty}^{\mathcal T}\le\epsilon_{\mathcal T}\),
the uniform criterion estimation error satisfies
\[
    e_{n,\infty}^{\mathcal T}
    \le \bar e_n^{\mathcal T}.
\]
Thus, \(\bar e_n^{\mathcal T}\) summarizes how first-stage
distribution estimation error propagates through the
criterion-specific H\"older stability condition.

\begin{theorem}[Baseline policy regret]
\label{thm:base-rate}
The plug-in treatment rule satisfies
$\mathcal R_{\mathcal T}(\widehat\pi_{\mathcal T})
\leq
2e_{n,\infty}^{\mathcal T}$.
Under Assumption \ref{ass:holder}, if
$\delta_{n,\infty}^{\mathcal T}\leq\epsilon_{\mathcal T}$, then $
\mathcal R_{\mathcal T}(\widehat\pi_{\mathcal T})
    \le 2\bar e_n^{\mathcal T}$.
\end{theorem}

Theorem \ref{thm:base-rate} establishes a direct link between first-stage conditional distribution estimation and the quality of the resulting treatment rule. A sharper guarantee is available when little probability mass lies near the treatment decision boundary. Let
$Q_\star^{\mathcal T}(x)
=
\max_{a\in\mathcal A}Q_a^{\mathcal T}(x)$,
and define the gap to the best strictly suboptimal treatment by
$\Delta_{\mathcal T}(x)
=
\min_{a:\,Q_a^{\mathcal T}(x)<Q_\star^{\mathcal T}(x)}
\left\{
Q_\star^{\mathcal T}(x)-Q_a^{\mathcal T}(x)
\right\}$,
with $\Delta_{\mathcal T}(x)=+\infty$ if all treatments are optimal.

\begin{assumption}[Margin condition]
\label{ass:margin}
There are \(t_0>0\) and a nondecreasing
\(\Psi_\Tt:[0,t_0]\to[0,1]\), with
\(\Psi_\Tt(t)\to0\) as \(t\downarrow0\), such that
\begin{equation*}
  \Pp_X\{\Delta_\Tt(X)\le t\}
  \le
  \Psi_\Tt(t),
  \qquad 0<t\le t_0.
\end{equation*}
\end{assumption}
The function $\Psi_{\mathcal T}$ characterizes the probability mass near the treatment decision boundary, with smaller values corresponding to fewer covariate values for which a strictly suboptimal treatment is nearly optimal. 
This condition encompasses the standard margin or low-noise condition used to obtain fast rates in plug-in classification and policy learning\citep{audibert2007fast,luedtke2017faster}. In comparison, existing direct-learning approaches obtain sharper rates under method-specific structural conditions, such as approximation-error, sparsity, or function-class assumptions \citep{qian2011performance,qi2023robustness}.

\begin{theorem}[Boundary-refined policy regret]
\label{thm:final-rate}
Under Assumptions \ref{ass:holder}-\ref{ass:margin}, if 
$\delta_{n,\infty}^{\mathcal T}\le\epsilon_{\mathcal T}$ and $2\bar e_n^{\mathcal T}\le t_0$,
then
$\mathcal R_{\mathcal T}(\widehat\pi_{\mathcal T})
    \le
    2\bar e_n^{\mathcal T}
    \Psi_{\mathcal T}\bigl(2\bar e_n^{\mathcal T}\bigr)$.
\end{theorem}

Theorems~\ref{thm:base-rate} and~\ref{thm:final-rate}
separate criterion stability from decision-boundary behavior.
The quantity \(\bar e_n^{\mathcal T}\) summarizes the propagation
of distribution estimation error through the criterion,
while \(\Psi_{\mathcal T}(2\bar e_n^{\mathcal T})\)
provides the boundary-dependent refinement of the baseline bound. Detailed verification for specific criteria, conditional distribution estimators, and explicit convergence-rate consequences under additional assumptions, is provided in Appendix~\ref{app:theory}.

\section{Simulations}
 We conduct simulation studies to evaluate the proposed framework. Throughout, we consider binary randomized trials with equal treatment assignment probabilities. We compare our methods with several representative baselines, including \(\ell_1\)-PLS \citep{qian2011performance}, RWL \citep{zhou2017residual}, CQR \citep{linn2017outcome,xiao2019robust}, and the direct \(\ell_2\)-CVaR method of \citet{qi2023robustness}. Within our distribution-first framework, we further compare several first-stage conditional distribution estimators, including Kernel, Distributional Random Forest (DRF), Conditional DDPM, and GCDS.
All results are based on 100 Monte Carlo replications. Kernel, 
\(\ell_1\)-PLS, RWL, and the direct
\(\ell_2\)-CVaR method use cross-validation for tuning; other methods use prespecified or separately selected hyperparameters as provided in Appendix \ref{app:exp}.
Across settings, the outcome is generated as
$Y = 1 + X_1 + X_2 + A\delta(\mathbf{X}) + \epsilon$.
We report four main metrics:
$
\texttt{Mean}, \ Q_{0.1}, \ \mathrm{CVaR}_{0.1}, \mathrm{CVaR}_{0.25},\ \texttt{Err}.
$
Here \texttt{Mean} measures overall policy value, and \texttt{Err} measures decision error relative to the oracle rule. We also report \(Q_{0.1}\), \(\mathrm{CVaR}_{0.1}\), and  \(\mathrm{CVaR}_{0.25}\) to assess lower-tail performance, since the proposed framework is intended to support risk-sensitive decision making. In particular, \(Q_{0.1}\) captures a lower-tail threshold, whereas \(\mathrm{CVaR}_{0.1}\) summarizes the average outcome among the worst \(10\%\) of cases. These are standard downside-risk summaries and are especially informative in heavy-tailed settings \citep{rockafellar2000optimization,qi2023robustness}. In the tables, for criterion-specific competitors, each row evaluates one native-trained rule under all reported criteria; for Kernel, DRF, Conditional DDPM, and GCDS, each Value/Err pair uses the plug-in rule targeting its column criterion. 

Overall, our framework shows competitive performance
among criterion-specific methods, especially when the outcome distribution skews from baseline settings. Among the candidate first-stage estimators, DRF is the most reliable overall choice.

\subsection{Distribution Shift Examples}
We first examine robustness under distribution shift, following the setup of \citet{qi2023robustness}. In both shifted settings, we use \(\delta(\mathbf{X})=X_1-X_2+X_3\), with \(n=200\) and \(p=20\). In the covariate shift setting, the training covariates are generated from
$
X_j \sim 0.7\,N(0,1)+0.3\,\mathrm{LogNormal}(0,2),
$
whereas the test covariates follow \(X_j\sim N(0,1)\); the noise remains \(\varepsilon\sim N(0,1)\) for both training and test samples. In the outcome shift setting, both training and test covariates follow \(\mathrm{Unif}([-1,1]^p)\), while the training noise is generated from
$
\varepsilon \sim 0.7\,N(0,1)+0.3\,\mathrm{LogNormal}(0,2),
$
and the test noise follows \(\varepsilon\sim N(0,1)\).
For simplicity, we only compare \texttt{Kernel} and \(\ell_2\)-CVaR in this subsection. This follows \citet{qi2023robustness}, where \(\ell_2\)-CVaR is already the strongest criterion-specific robust competitor in this experiment.

Table~\ref{tab:shift_main} shows that Kernel is consistently more stable than \(\ell_2\)-CVaR under both types of shift.
The main reason might be that our method depends more directly on the conditional distribution, instead of a single criterion-specific training objective, and therefore is more robust when the training and test environments differ.
The internal comparison within the framework is also informative. 
In the outcome shift scenario, Kernel plug-in rules based on risk-sensitive criteria substantially outperform the mean-based rule, reflecting that lower-tail criteria are less affected by the heavy-tailed training noise than the conditional mean.

\begin{table*}[t]
\centering
\caption{Mean and CVaR performance under distribution shift. Standard errors are reported in parentheses.}
\label{tab:shift_main}
\fontsize{7pt}{8.5pt}\selectfont
\setlength{\tabcolsep}{1pt}
\begin{tabular*}{\textwidth}{@{\extracolsep{\fill}}llcccccccc@{}}
\toprule
& & \multicolumn{2}{c}{Mean} & \multicolumn{2}{c}{$\mathrm{CVaR}_{0.1}$} & \multicolumn{2}{c}{$\mathrm{CVaR}_{0.25}$} & \multicolumn{2}{c}{$\mathrm{CVaR}_{0.5}$} \\
\cmidrule(lr){3-4}\cmidrule(lr){5-6}\cmidrule(lr){7-8}\cmidrule(l){9-10}
Scenario & Method & Value & Err & Value & Err & Value & Err & Value & Err \\
\midrule
\multirow{5}{*}{\shortstack[l]{Covariate\\shift}} & Oracle & \itrtablecell{2.380}{0.003} & \itrtablecell{0.000}{0.000} & \itrtablecell{\textminus1.061}{0.005} & \itrtablecell{0.000}{0.000} & \itrtablecell{\textminus0.142}{0.004} & \itrtablecell{0.000}{0.000} & \itrtablecell{0.773}{0.003} & \itrtablecell{0.000}{0.000} \\
& $\ell_2$-CVaR (0.1) & \itrtablecell{2.092}{0.025} & \itrtablecell{0.198}{0.009} & \itrtablecell{\textminus1.695}{0.046} & \itrtablecell{0.198}{0.009} & \itrtablecell{\textminus0.656}{0.039} & \itrtablecell{0.198}{0.009} & \itrtablecell{0.360}{0.033} & \itrtablecell{0.198}{0.009} \\
& $\ell_2$-CVaR (0.25) & \itrtablecell{2.145}{0.019} & \itrtablecell{0.179}{0.008} & \itrtablecell{\textminus1.605}{0.038} & \itrtablecell{0.179}{0.008} & \itrtablecell{\textminus0.573}{0.032} & \itrtablecell{0.179}{0.008} & \itrtablecell{0.431}{0.027} & \itrtablecell{0.179}{0.008} \\
& $\ell_2$-CVaR (0.5) & \itrtablecell{2.174}{0.021} & \itrtablecell{0.162}{0.008} & \itrtablecell{\textminus1.547}{0.042} & \itrtablecell{0.162}{0.008} & \itrtablecell{\textminus0.528}{0.035} & \itrtablecell{0.162}{0.008} & \itrtablecell{0.471}{0.029} & \itrtablecell{0.162}{0.008} \\
& Kernel & \bestcell{2.333}{0.004} & \bestcell{0.077}{0.002} & \bestcell{\textminus1.419}{0.023} & \bestcell{0.113}{0.003} & \bestcell{\textminus0.323}{0.011} & \bestcell{0.098}{0.003} & \bestcell{0.671}{0.006} & \bestcell{0.091}{0.003} \\
\midrule
\multirow{5}{*}{\shortstack[l]{Outcome\\shift}} & Oracle & \itrtablecell{1.812}{0.002} & \itrtablecell{0.000}{0.000} & \itrtablecell{\textminus0.702}{0.004} & \itrtablecell{0.000}{0.000} & \itrtablecell{\textminus0.016}{0.003} & \itrtablecell{0.000}{0.000} & \itrtablecell{0.673}{0.002} & \itrtablecell{0.000}{0.000} \\
& $\ell_2$-CVaR (0.1) & \itrtablecell{1.430}{0.019} & \itrtablecell{0.314}{0.009} & \itrtablecell{\textminus1.349}{0.024} & \itrtablecell{0.314}{0.009} & \itrtablecell{\textminus0.594}{0.024} & \itrtablecell{0.314}{0.009} & \itrtablecell{0.169}{0.023} & \itrtablecell{0.314}{0.009} \\
& $\ell_2$-CVaR (0.25) & \itrtablecell{1.463}{0.018} & \itrtablecell{0.298}{0.009} & \itrtablecell{\textminus1.300}{0.024} & \itrtablecell{0.298}{0.009} & \itrtablecell{\textminus0.550}{0.023} & \itrtablecell{0.298}{0.009} & \itrtablecell{0.209}{0.022} & \itrtablecell{0.298}{0.009} \\
& $\ell_2$-CVaR (0.5) & \itrtablecell{1.504}{0.018} & \itrtablecell{0.277}{0.008} & \itrtablecell{\textminus1.250}{0.023} & \itrtablecell{0.277}{0.008} & \itrtablecell{\textminus0.497}{0.023} & \itrtablecell{0.277}{0.008} & \itrtablecell{0.259}{0.022} & \itrtablecell{0.277}{0.008} \\
& Kernel & \bestcell{1.514}{0.021} & \bestcell{0.260}{0.010} & \bestcell{\textminus0.892}{0.014} & \bestcell{0.131}{0.004} & \bestcell{\textminus0.114}{0.008} & \bestcell{0.105}{0.003} & \bestcell{0.621}{0.005} & \bestcell{0.089}{0.003} \\
\bottomrule
\end{tabular*}
\end{table*}

\subsection{Comparison Across Outcome Regimes}
\label{subsec:exp}
We evaluate the proposed framework further via three additional examples. In the main text, we focus on dimension \(p=10\) and $n=1000$ since this setting is close to our real data, with additional results deferred to Appendix \ref{app:exp}. $X_j\overset{\mathrm{iid}}{\sim}\mathrm{Unif}[-1,1]$. The testing sample size is 10,000.
The experimental settings below are from \citep{qi2023robustness}:

\noindent\textbf{Setting 1 (linear Gaussian):}
\(\delta(X)=X_1-X_2+X_3,\ \varepsilon\sim N(0,1)\).

\noindent\textbf{Setting 2 (nonlinear Gaussian):}
\(\delta(X)=\varsigma(X_1)\varsigma(X_2)-c_0,\ 
\varsigma(x)=1+1/(1+\exp\{-20(x-1/3)\}),\ 
\varepsilon\sim N(0,1)\),
with \(c_0=1.87\).

\noindent\textbf{Setting 3 (heavy-tail):}
\(\delta(X)=X_1-X_2+X_3,\ 
\log\varepsilon\sim N(0,2^2)\).

Setting~1 is the standard sparse linear Gaussian benchmark and is oracle for both \(\ell_1\)-PLS and CQR. Setting~2 is used to test adaptation to nonlinear decision boundary. Setting~3 introduces heavy right-tailed lognormal noise, which is used to assess the robustness of different ITRs and is also oracle for CQR.
Because $\varepsilon$ is independent of $(X, A)$ in Settings 1--3, the two treatment-specific conditional outcome distributions differ only by a location shift. All criteria considered here are monotone with respect to first-order stochastic dominance, so they share the same oracle rule $\operatorname{sign}\{\delta(X)\}$. These settings therefore assess how accurately each method recovers this common rule under different outcome regimes. Appendix~\ref{app:treatment-interactive} considers a treatment-interactive design in which the optimal rule genuinely depends on the criterion.

The results are summarized in Tables~\ref{tab:s1_main}--\ref{tab:s3_main}. Overall, our framework shows competitive performance
among criterion-specific methods.
In Setting~1, the proposed framework remains competitive even in this favorable regime for classical linear baselines. Within the distribution-first family, Conditional DDPM performs best, followed by GCDS and DRF.
In Setting~2, DRF becomes the strongest method among all competitors, showing its ability to adapt to nonlinear treatment heterogeneity. Conditional DDPM and GCDS remain
competitive with criterion-specific methods.
In Setting~3, GCDS, Kernel, and DRF all clearly outper-
form the other baselines beyond oracle CQR. In the distribution-first family, DRF has the highest $Q_{0.1}$ and $\mathrm{CVaR}_{0.1}$ Values, whereas Kernel has the highest $\mathrm{CVaR}_{0.25}$ Values and the lowest Err at all four tail criteria. 
The linear mean-based baselines deteriorate sharply in this regime. Since all criteria share the same oracle rule here, this deterioration reflects the sensitivity of squared-loss estimation to heavy right-tailed noise, whereas quantile- and CVaR-based plug-in rules are more robust to such noise.
Together with the treatment-interactive and higher-dimensional experiments in Appendix~\ref{app:exp}, these results show that DRF is the most reliable first-stage estimator in most settings, whereas Kernel degrades quickly as $p$ grows and the generative estimators are less stable, with one GCDS run failing. We therefore recommend DRF as the default first-stage estimator and regard generative models as promising but exploratory.

\begin{table*}[t]
\centering
\caption{Setting 1 (linear Gaussian; $n=1000$, $p=10$). Entries are means (Monte Carlo standard errors) over 100 paired repetitions.}
\label{tab:s1_main}
\fontsize{7pt}{8.5pt}\selectfont
\setlength{\tabcolsep}{1pt}
\begin{tabular*}{\textwidth}{@{\extracolsep{\fill}}lcccccccc@{}}
\toprule
& \multicolumn{2}{c}{Mean} & \multicolumn{2}{c}{$\mathrm{Quantile}_{0.1}$} & \multicolumn{2}{c}{$\mathrm{CVaR}_{0.1}$} & \multicolumn{2}{c}{$\mathrm{CVaR}_{0.25}$} \\
\cmidrule(lr){2-3}\cmidrule(lr){4-5}\cmidrule(lr){6-7}\cmidrule(l){8-9}
Method & Value & Err & Value & Err & Value & Err & Value & Err \\
\midrule
Oracle & \itrtablecell{1.810}{0.002} & \itrtablecell{0.000}{0.000} & \itrtablecell{\textminus0.042}{0.004} & \itrtablecell{0.000}{0.000} & \itrtablecell{\textminus0.699}{0.004} & \itrtablecell{0.000}{0.000} & \itrtablecell{\textminus0.015}{0.003} & \itrtablecell{0.000}{0.000} \\
$\ell_1$-PLS & \itrtablecell{1.805}{0.002} & \itrtablecell{0.034}{0.002} & \bestcell{\textminus0.050}{0.004} & \itrtablecell{0.034}{0.002} & \itrtablecell{\textminus0.712}{0.005} & \bestcell{0.034}{0.002} & \itrtablecell{\textminus0.024}{0.004} & \itrtablecell{0.034}{0.002} \\
RWL & \itrtablecell{1.800}{0.002} & \itrtablecell{0.051}{0.002} & \itrtablecell{\textminus0.060}{0.004} & \itrtablecell{0.051}{0.002} & \itrtablecell{\textminus0.720}{0.004} & \itrtablecell{0.051}{0.002} & \itrtablecell{\textminus0.032}{0.003} & \itrtablecell{0.051}{0.002} \\
\addlinespace[3pt]
CQR (0.1) & \itrtablecell{1.799}{0.002} & \itrtablecell{0.053}{0.001} & \itrtablecell{\textminus0.063}{0.004} & \itrtablecell{0.053}{0.001} & \itrtablecell{\textminus0.723}{0.004} & \itrtablecell{0.053}{0.001} & \itrtablecell{\textminus0.035}{0.003} & \itrtablecell{0.053}{0.001} \\
\addlinespace[3pt]
$\ell_2$-CVaR (0.1) & \itrtablecell{1.776}{0.003} & \itrtablecell{0.092}{0.002} & \itrtablecell{\textminus0.102}{0.005} & \itrtablecell{0.092}{0.002} & \itrtablecell{\textminus0.768}{0.005} & \itrtablecell{0.092}{0.002} & \itrtablecell{\textminus0.075}{0.004} & \itrtablecell{0.092}{0.002} \\
$\ell_2$-CVaR (0.25) & \itrtablecell{1.786}{0.003} & \itrtablecell{0.078}{0.002} & \itrtablecell{\textminus0.088}{0.004} & \itrtablecell{0.078}{0.002} & \itrtablecell{\textminus0.754}{0.005} & \itrtablecell{0.078}{0.002} & \itrtablecell{\textminus0.060}{0.004} & \itrtablecell{0.078}{0.002} \\
\midrule
Kernel & \itrtablecell{1.771}{0.003} & \itrtablecell{0.096}{0.002} & \itrtablecell{\textminus0.221}{0.006} & \itrtablecell{0.148}{0.002} & \itrtablecell{\textminus0.923}{0.007} & \itrtablecell{0.146}{0.002} & \itrtablecell{\textminus0.144}{0.006} & \itrtablecell{0.125}{0.002} \\
DRF & \itrtablecell{1.788}{0.002} & \itrtablecell{0.075}{0.001} & \itrtablecell{\textminus0.096}{0.004} & \itrtablecell{0.084}{0.001} & \itrtablecell{\textminus0.766}{0.005} & \itrtablecell{0.085}{0.001} & \itrtablecell{\textminus0.062}{0.003} & \itrtablecell{0.080}{0.001} \\
Conditional DDPM & \bestcell{1.807}{0.002} & \bestcell{0.032}{0.001} & \bestcell{\textminus0.050}{0.004} & \bestcell{0.033}{0.001} & \bestcell{\textminus0.709}{0.004} & \bestcell{0.034}{0.001} & \bestcell{\textminus0.023}{0.003} & \bestcell{0.033}{0.001} \\
GCDS & \itrtablecell{1.798}{0.003} & \itrtablecell{0.055}{0.001} & \itrtablecell{\textminus0.077}{0.004} & \itrtablecell{0.067}{0.002} & \itrtablecell{\textminus0.756}{0.005} & \itrtablecell{0.076}{0.003} & \itrtablecell{\textminus0.050}{0.003} & \itrtablecell{0.066}{0.002} \\
\bottomrule
\end{tabular*}
\end{table*}

\begin{table*}[t]
\centering
\caption{Setting 2 (nonlinear Gaussian; $n=1000$, $p=10$). Entries are means (Monte Carlo standard errors) over 100 paired repetitions.}
\label{tab:s2_main}
\fontsize{7pt}{8.5pt}\selectfont
\setlength{\tabcolsep}{1pt}
\begin{tabular*}{\textwidth}{@{\extracolsep{\fill}}lcccccccc@{}}
\toprule
& \multicolumn{2}{c}{Mean} & \multicolumn{2}{c}{$\mathrm{Quantile}_{0.1}$} & \multicolumn{2}{c}{$\mathrm{CVaR}_{0.1}$} & \multicolumn{2}{c}{$\mathrm{CVaR}_{0.25}$} \\
\cmidrule(lr){2-3}\cmidrule(lr){4-5}\cmidrule(lr){6-7}\cmidrule(l){8-9}
Method & Value & Err & Value & Err & Value & Err & Value & Err \\
\midrule
Oracle & \itrtablecell{1.654}{0.003} & \itrtablecell{0.000}{0.000} & \itrtablecell{\textminus0.044}{0.003} & \itrtablecell{0.000}{0.000} & \itrtablecell{\textminus0.587}{0.003} & \itrtablecell{0.000}{0.000} & \itrtablecell{\textminus0.029}{0.002} & \itrtablecell{0.000}{0.000} \\
$\ell_1$-PLS & \itrtablecell{1.566}{0.003} & \itrtablecell{0.173}{0.001} & \itrtablecell{\textminus0.156}{0.003} & \itrtablecell{0.173}{0.001} & \itrtablecell{\textminus0.697}{0.003} & \itrtablecell{0.173}{0.001} & \itrtablecell{\textminus0.139}{0.003} & \itrtablecell{0.173}{0.001} \\
RWL & \itrtablecell{1.592}{0.003} & \itrtablecell{0.214}{0.002} & \itrtablecell{\textminus0.118}{0.003} & \itrtablecell{0.214}{0.002} & \itrtablecell{\textminus0.658}{0.003} & \itrtablecell{0.214}{0.002} & \itrtablecell{\textminus0.102}{0.003} & \itrtablecell{0.214}{0.002} \\
\addlinespace[3pt]
CQR (0.1) & \itrtablecell{1.561}{0.003} & \itrtablecell{0.182}{0.001} & \itrtablecell{\textminus0.166}{0.004} & \itrtablecell{0.182}{0.001} & \itrtablecell{\textminus0.709}{0.004} & \itrtablecell{0.182}{0.001} & \itrtablecell{\textminus0.148}{0.003} & \itrtablecell{0.182}{0.001} \\
\addlinespace[3pt]
$\ell_2$-CVaR (0.1) & \itrtablecell{1.559}{0.004} & \itrtablecell{0.253}{0.003} & \itrtablecell{\textminus0.140}{0.004} & \itrtablecell{0.253}{0.003} & \itrtablecell{\textminus0.679}{0.004} & \itrtablecell{0.253}{0.003} & \itrtablecell{\textminus0.124}{0.003} & \itrtablecell{0.253}{0.003} \\
$\ell_2$-CVaR (0.25) & \itrtablecell{1.578}{0.003} & \itrtablecell{0.242}{0.003} & \itrtablecell{\textminus0.125}{0.004} & \itrtablecell{0.242}{0.003} & \itrtablecell{\textminus0.664}{0.004} & \itrtablecell{0.242}{0.003} & \itrtablecell{\textminus0.109}{0.003} & \itrtablecell{0.242}{0.003} \\
\midrule
Kernel & \itrtablecell{1.582}{0.003} & \itrtablecell{0.231}{0.002} & \itrtablecell{\textminus0.237}{0.004} & \itrtablecell{0.412}{0.003} & \itrtablecell{\textminus0.807}{0.004} & \itrtablecell{0.448}{0.003} & \itrtablecell{\textminus0.219}{0.004} & \itrtablecell{0.416}{0.003} \\
DRF & \bestcell{1.638}{0.003} & \bestcell{0.066}{0.004} & \bestcell{\textminus0.076}{0.003} & \bestcell{0.129}{0.007} & \bestcell{\textminus0.626}{0.004} & \bestcell{0.158}{0.008} & \bestcell{\textminus0.060}{0.003} & \bestcell{0.124}{0.007} \\
Conditional DDPM & \itrtablecell{1.596}{0.003} & \itrtablecell{0.200}{0.001} & \itrtablecell{\textminus0.121}{0.003} & \itrtablecell{0.195}{0.001} & \itrtablecell{\textminus0.663}{0.003} & \itrtablecell{0.193}{0.001} & \itrtablecell{\textminus0.105}{0.002} & \itrtablecell{0.195}{0.001} \\
GCDS & \itrtablecell{1.568}{0.003} & \itrtablecell{0.184}{0.001} & \itrtablecell{\textminus0.192}{0.007} & \itrtablecell{0.187}{0.001} & \itrtablecell{\textminus0.769}{0.011} & \itrtablecell{0.195}{0.002} & \itrtablecell{\textminus0.173}{0.006} & \itrtablecell{0.187}{0.001} \\
\bottomrule
\end{tabular*}
\end{table*}

\begin{table*}[t]
\centering
\caption{Setting 3 (heavy-tailed lognormal; $n=1000$, $p=10$). Entries are means (Monte Carlo standard errors) over 100 paired repetitions.}
\label{tab:s3_main}
\fontsize{7pt}{8.5pt}\selectfont
\setlength{\tabcolsep}{1pt}
\begin{tabular*}{\textwidth}{@{\extracolsep{\fill}}lcccccccc@{}}
\toprule
& \multicolumn{2}{c}{Mean} & \multicolumn{2}{c}{$\mathrm{Quantile}_{0.1}$} & \multicolumn{2}{c}{$\mathrm{CVaR}_{0.1}$} & \multicolumn{2}{c}{$\mathrm{CVaR}_{0.25}$} \\
\cmidrule(lr){2-3}\cmidrule(lr){4-5}\cmidrule(lr){6-7}\cmidrule(l){8-9}
Method & Value & Err & Value & Err & Value & Err & Value & Err \\
\midrule
Oracle & \itrtablecell{9.244}{0.075} & \itrtablecell{0.000}{0.000} & \itrtablecell{1.167}{0.003} & \itrtablecell{0.000}{0.000} & \itrtablecell{0.615}{0.003} & \itrtablecell{0.000}{0.000} & \itrtablecell{1.243}{0.002} & \itrtablecell{0.000}{0.000} \\
$\ell_1$-PLS & \itrtablecell{8.670}{0.081} & \itrtablecell{0.432}{0.011} & \itrtablecell{0.267}{0.028} & \itrtablecell{0.432}{0.011} & \itrtablecell{\textminus0.339}{0.025} & \itrtablecell{0.432}{0.011} & \itrtablecell{0.357}{0.027} & \itrtablecell{0.432}{0.011} \\
RWL & \itrtablecell{8.759}{0.089} & \itrtablecell{0.396}{0.009} & \itrtablecell{0.348}{0.024} & \itrtablecell{0.396}{0.009} & \itrtablecell{\textminus0.263}{0.023} & \itrtablecell{0.396}{0.009} & \itrtablecell{0.444}{0.024} & \itrtablecell{0.396}{0.009} \\
\addlinespace[3pt]
CQR (0.1) & \bestcell{9.243}{0.076} & \bestcell{0.009}{0.000} & \bestcell{1.167}{0.003} & \bestcell{0.009}{0.000} & \bestcell{0.614}{0.003} & \bestcell{0.009}{0.000} & \bestcell{1.242}{0.002} & \bestcell{0.009}{0.000} \\
\addlinespace[3pt]
$\ell_2$-CVaR (0.1) & \itrtablecell{9.074}{0.084} & \itrtablecell{0.190}{0.016} & \itrtablecell{0.867}{0.040} & \itrtablecell{0.190}{0.016} & \itrtablecell{0.279}{0.041} & \itrtablecell{0.190}{0.016} & \itrtablecell{0.943}{0.039} & \itrtablecell{0.190}{0.016} \\
$\ell_2$-CVaR (0.25) & \itrtablecell{9.080}{0.086} & \itrtablecell{0.186}{0.014} & \itrtablecell{0.884}{0.036} & \itrtablecell{0.186}{0.014} & \itrtablecell{0.289}{0.037} & \itrtablecell{0.186}{0.014} & \itrtablecell{0.959}{0.035} & \itrtablecell{0.186}{0.014} \\
\midrule
Kernel & \itrtablecell{8.681}{0.082} & \itrtablecell{0.416}{0.007} & \itrtablecell{1.069}{0.004} & \bestcell{0.116}{0.001} & \itrtablecell{0.468}{0.004} & \bestcell{0.109}{0.001} & \bestcell{1.168}{0.003} & \bestcell{0.096}{0.001} \\
DRF & \itrtablecell{8.861}{0.081} & \itrtablecell{0.372}{0.010} & \bestcell{1.088}{0.004} & \itrtablecell{0.118}{0.002} & \bestcell{0.516}{0.005} & \itrtablecell{0.118}{0.002} & \itrtablecell{1.166}{0.003} & \itrtablecell{0.114}{0.002} \\
Conditional DDPM & \itrtablecell{8.994}{0.081} & \itrtablecell{0.266}{0.009} & \itrtablecell{0.897}{0.018} & \itrtablecell{0.194}{0.007} & \itrtablecell{0.267}{0.021} & \itrtablecell{0.201}{0.008} & \itrtablecell{0.976}{0.017} & \itrtablecell{0.193}{0.007} \\
GCDS & \bestcell{9.089}{0.074} & \bestcell{0.163}{0.008} & \itrtablecell{1.024}{0.014} & \itrtablecell{0.138}{0.006} & \itrtablecell{0.411}{0.016} & \itrtablecell{0.142}{0.006} & \itrtablecell{1.097}{0.013} & \itrtablecell{0.136}{0.006} \\
\bottomrule
\end{tabular*}
\end{table*}

\section{Real data analysis}
We evaluate the proposed framework on the ACTG175 HIV clinical trial,  a randomized study
comparing several antiretroviral regimens in HIV-infected adult\citep{hammer1996trial}. We compare ZDV+ddI with ZDV+ZAL among 1,046 participants, and use the cohort and 12 baseline covariates supplied with \citet{qi2023robustness}. The
outcome is the change in CD4 count from baseline to follow-up with larger values preferred.  All methods use the same 200 treatment-stratified 80/20 splits.
The results show that our framework yields a more flexible treatment rule than the
criterion-specific methods, while maintaining strong out-of-sample performance.

We compare \(\ell_1\)-PLS, \(\ell_2\)-CVaR(0.25), and CQR(0.25), with three policies derived
from our distribution-first framework: DRF-Mean, DRF-CVaR(0.25), and DRF-CPT.
For the proposed
approach, we first estimate the conditional outcome distribution under each treatment arm using
DRF, and then construct treatment rules by optimizing different decision criteria with respect to the
estimated distribution. All learned policies are evaluated on held-out test sets using mean value,
$\text{CVaR}_{0.25}$, and CPT. We also report the proportion of subjects assigned to treatment 1 (ZDV+ddI) and runtime.
CPT uses reference point zero (no CD4 change), with \(\eta_+=\eta_-=0.88\), \(\lambda=2.25\), \(\gamma_+=0.61\), and \(\gamma_-=0.69\) \citep{tversky1992advances}. This choice is natural in the present setting, where avoiding poor outcomes and
accounting for asymmetric gain-loss preferences may both be clinically relevant.

Table~\ref{tab:actg175_main} reports all three utilities for each fixed learned rule. DRF-Mean has the highest average held-out mean, while DRF-CPT has the highest average marginal CVaR and CPT among the six policies. DRF-CVaR also has a numerically higher average CVaR than the three baselines, but a lower average mean, illustrating a tradeoff between the evaluated utilities. The three DRF-based policies also produce meaningfully different treatment rates, indicating that different downstream criteria can lead to substantively different treatment recommendations even when they are based on the same estimated conditional outcome distribution. This illustrates the practical value of the distribution-first framework that once a flexible conditional distribution model is fitted, one can construct and compare treatment rules tailored to different decision objectives within a single unified pipeline.

Runtime includes split extraction, initialization, parameter selection where applicable, fitting, prediction, and held-out evaluation. The DRF time covers all three policies and their full cross-utility evaluations. The CVaR pipeline includes five-fold cross-validation over 11 regularization parameters, whereas DRF uses fixed settings without a tuning search. With the split-specific regularization parameter already selected, the separately timed post-selection CVaR workflow takes 1.849 [1.380, 8.533] seconds. Thus, DRF's computational benefit is reusable distribution estimation across criteria, rather than uniformly faster single-criterion learning.

\begin{table}[t]
\centering
\caption{ACTG175 policy evaluation and local computational cost. Utilities and treatment rates are mean (standard deviation) across 200 paired splits; runtimes are median [Q1, Q3] in seconds. Each row evaluates the same learned rule under all three utilities. The three DRF rules share one distribution-estimation stage, and their combined runtime is reported once. Bold indicates the largest average value in each utility column.}
\label{tab:actg175_main}
\fontsize{8}{9.5}\selectfont
\setlength{\tabcolsep}{3pt}
\renewcommand{\arraystretch}{2.1}
\newcommand{\actgcell}[2]{\shortstack{#1\\(#2)}}
\newcommand{\actgbestcell}[2]{\shortstack{\textbf{#1}\\(#2)}}
\begin{tabular*}{\linewidth}{@{\extracolsep{\fill}}llccccc@{}}
\toprule
Method & Criterion & Mean & $\cvar_{0.25}$ & CPT & \shortstack{Treat-1\\rate} & \shortstack{Runtime\\(s)} \\
\midrule
\(\ell_1\)-PLS & Mean
& \actgcell{51.797}{12.731}
& \actgcell{-107.879}{21.484}
& \actgcell{-3.485}{13.110}
& \actgcell{0.887}{0.067}
& \shortstack{0.073\\{}[0.051, 0.083]} \\
\(\ell_2\)-CVaR (0.25) & CVaR (0.25)
& \actgcell{52.878}{13.061}
& \actgcell{-107.955}{21.109}
& \actgcell{-2.946}{12.631}
& \actgcell{0.921}{0.123}
& \shortstack{193.587\\{}[135.079, 268.785]} \\
CQR (0.25) & Quantile (0.25)
& \actgcell{48.280}{12.845}
& \actgcell{-109.971}{20.140}
& \actgcell{-6.121}{12.554}
& \actgcell{0.702}{0.071}
& \shortstack{0.005\\{}[0.004, 0.015]} \\
\midrule
\multirow{3}{*}{\shortstack[l]{DRF\\(shared fit)}} & Mean
& \actgbestcell{54.042}{12.292}
& \actgcell{-108.808}{21.467}
& \actgcell{-3.058}{12.720}
& \actgcell{0.983}{0.021}
& \multirow{3}{*}{\shortstack{4.042\\{}[3.238, 4.662]}} \\
& CVaR (0.25)
& \actgcell{44.950}{12.863}
& \actgcell{-105.383}{17.219}
& \actgcell{-2.555}{11.481}
& \actgcell{0.575}{0.107}
& \\
& CPT
& \actgcell{53.637}{12.150}
& \actgbestcell{-104.632}{19.734}
& \actgbestcell{-1.246}{12.363}
& \actgcell{0.861}{0.076}
& \\
\bottomrule
\end{tabular*}
\end{table}

\section{Discussion}
We study risk-sensitive ITR learning through a distribution-first framework. The key idea is to first estimate the treatment-specific conditional outcome distribution and then construct the treatment rule by plug-in criterion evaluation. This yields a unified pipeline for multiple risk-sensitive criteria, including mean, quantile, CVaR, and CPT-type objectives. Our results support this perspective both theoretically and empirically. The theory provides a general error propagation result from first-stage distribution estimation to downstream criterion estimation and treatment regret. Empirically, performance depends on the outcome regime and first-stage estimator, while the treatment-interactive and real-data analyses illustrate criterion-dependent decisions using shared distribution estimates.

Several directions remain for future work. First, extending the distribution-first approach to doubly robust estimators is a potential next step, especially for observational settings. Second, the current theory focuses on H\"older-continuous criteria; sharper results for specific criteria may yield stronger and more interpretable guarantees. Third, our experiments suggest that gains from generative models depend on the outcome regime and target criterion. Future work should therefore investigate broader data types and response structures where generative models may be better matched to the underlying outcome distribution.

\section*{Acknowledgments and Disclosure of Funding}
The authors thank the reviewers for their helpful and constructive
comments, which have improved the presentation of this paper.
This research was supported in part by NSF grant DMS-2625396.
Any opinions, findings, conclusions, or recommendations expressed
in this material are those of the authors and do not necessarily
reflect the views of the National Science Foundation.
The authors declare no competing interests.

\bibliographystyle{unsrtnat}
\bibliography{ref}

\newpage
\appendix

\section{Additional theoretical details}
\label{app:theory}
\subsection{Verification of the H\"older Continuity Condition}
\label{app:holder-verify}

Assumption~\ref{ass:holder} is formulated at the level of criterion stability and allows the semimetric $d_{\mathcal T}$ to depend on the distributional criterion and the underlying distribution class. The formulation is closely related to the CDF-based Lipschitz risk-functional framework of \citet{huang2021offpolicy}, which establishes plug-in stability for a broad class of Lipschitz functionals and verifies such properties for criteria including the mean, CVaR, variance, mean–variance, and CPT-type objectives. These results correspond to the special case $\rho_{\mathcal T}=1$ of Assumption~\ref{ass:holder}.

The H\"older formulation relaxes the global Lipschitz requirement in two directions. First, it requires stability only locally, which accommodates criteria such as quantiles that are not globally Lipschitz but are locally Lipschitz under standard regularity conditions around the target quantile. Second, it allows H\"older exponents ($\rho_{\mathcal T}<1$), which accommodates functionals that are locally continuous but not Lipschitz. In particular, the Tversky–Kahneman probability distortions used in our CPT specification exhibit singular behavior near the probability endpoints and are generally not Lipschitz on [0,1]. We show below that these distortions are H\"older continuous, yielding a corresponding H\"older stability bound for the CPT functional.

Write
\[
  d_\infty(F,G)=\sup_{y\in\R}\abs{F(y)-G(y)}
\]
and let \(F^{-1}(u)=\inf\{y:F(y)\ge u\}\).  The next proposition records
conditions under which common criteria satisfy
Assumption~\ref{ass:holder}. These conditions are not interchangeable:
for example, CDF sup-norm convergence on an unbounded support does not by
itself control a mean or a tail average.

For CPT, the formal statement below uses a prespecified fixed reference point
\(r_0=0\), matching the ACTG175 analysis.  A covariate-dependent reference
point \(r(x)\) instead defines a family of functionals \(\Tt_x\); the same
argument applies only when the stability modulus and all transformed-outcome
bounds hold uniformly in \(x\).

\begin{proposition}[Representative stability moduli]
\label{prop:functional-stability}
Suppose \(F\) and \(G\) have common support contained in
\([\ell,u]\), and let \(D=u-\ell\) and
\(M=\max\{\abs{\ell},\abs{u}\}\).
\begin{enumerate}
  \item For the mean \(\Tt_{\mathrm{mean}}(F)=\int y\,dF(y)\),
  \[
    \abs{\Tt_{\mathrm{mean}}(F)-\Tt_{\mathrm{mean}}(G)}
    \le D\,d_\infty(F,G).
  \]

  \item For \(\lambda\ge0\) and
  \(\Tt_{\mathrm{MV}}(F)=\Tt_{\mathrm{mean}}(F)-\lambda\Tt_{\var}(F)\),
  \[
    \abs{\Tt_{\mathrm{MV}}(F)-\Tt_{\mathrm{MV}}(G)}
    \le
    D(1+4\lambda M)\,d_\infty(F,G).
  \]

  \item For the lower-tail CVaR of a reward, obtained by
  sign-converting the usual upper-tail loss convention,
  \[
    \Tt_{\cvar,\tau}(F)
    =
    \frac{1}{\tau}\int_0^\tau F^{-1}(v)\,dv,
    \qquad \tau\in(0,1),
  \]
  \[
    \abs{\Tt_{\cvar,\tau}(F)-\Tt_{\cvar,\tau}(G)}
    \le
    \frac{D}{\tau}\,d_\infty(F,G).
  \]

  \item Let \(\Tt_{Q,\tau}(F)=F^{-1}(\tau)\), write
  \(q_\tau=F^{-1}(\tau)\), and suppose \(F(q_\tau)=\tau\).
  Assume that, for some \(r_\tau,c_\tau>0\), \(F\) is absolutely
  continuous on
  \([q_\tau-r_\tau,q_\tau+r_\tau]\), with density satisfying
  \(f(y)\ge c_\tau\) almost everywhere on this interval. If
  \[
  d_\infty(F,G)
  <
  \min\{\tau,1-\tau,c_\tau r_\tau\},
  \]
  then
  \[
    \abs{\Tt_{Q,\tau}(F)-\Tt_{Q,\tau}(G)}
    \le
    c_\tau^{-1}d_\infty(F,G).
  \]
  For a bound uniform in treatments and covariates, the same
  \(r_\tau\), \(c_\tau\), and crossing condition
  \(F_a(q_{\tau,a}(x)\mid x)=\tau\) must hold uniformly over \(a\) and \(x\).

  \item Consider a cumulative-prospect functional
  \citep{tversky1992advances} with a nondecreasing value
  function \(v\), the fixed reference point normalized to zero so that
  \(v(0)=0\), and
  transformed gains and losses
  \(v^+(y)=\max\{v(y),0\}\) and
  \(v^-(y)=\max\{-v(y),0\}\) satisfying
  \(0\le v^+(Y)\le M_+\) and \(0\le v^-(Y)\le M_-\), and probability
  distortions satisfying
  \[
    \abs{w_\pm(s)-w_\pm(t)}
    \le L_\pm\abs{s-t}^{\rho_\pm}.
  \]
  Define its gain-minus-loss Choquet tail representation by
  \begin{align*}
    \Tt_{\mathrm{CPT}}(F)
    &=
    \int_0^{M_+}
    w_+\!\left(
      \Pp_F\{v^+(Y)>z\}
    \right)\,dz
    \\
    &\quad-
    \int_0^{M_-}
    w_-\!\left(
      \Pp_F\{v^-(Y)>z\}
    \right)\,dz.
  \end{align*}
  With \(\rho=\min(\rho_+,\rho_-)\), this functional satisfies
  \[
    \abs{\Tt_{\mathrm{CPT}}(F)-\Tt_{\mathrm{CPT}}(G)}
    \le
    (L_+M_++L_-M_-)\,
    d_\infty(F,G)^\rho.
  \]
\end{enumerate}
\end{proposition}

\begin{proof}
In one dimension,
\[
  W_1(F,G)
  =
  \int_\ell^u\abs{F(y)-G(y)}\,dy
  =
  \int_0^1\abs{F^{-1}(v)-G^{-1}(v)}\,dv,
\]
so \(W_1(F,G)\le Dd_\infty(F,G)\).  The mean is
1-Lipschitz in \(W_1\), proving the first claim.  Integration by parts
for the bounded-variation function \(y^2\) gives
\[
  \abs{\E_F(Y^2)-\E_G(Y^2)}
  \le2MD\,d_\infty(F,G).
\]
Moreover,
\[
  \abs{\mu(F)^2-\mu(G)^2}
  \le2M\abs{\mu(F)-\mu(G)}
  \le2MD\,d_\infty(F,G),
\]
which proves the stated mean-variance bound.  For lower-tail CVaR,
\[
  \abs{\cvar_\tau(F)-\cvar_\tau(G)}
  \le
  \tau^{-1}W_1(F,G).
\]
For the quantile statement, let
\(\delta=d_\infty(F,G)\) and \(q_\tau=F^{-1}(\tau)\).
The assumed bound on \(\delta\) ensures that
\(\tau-\delta,\tau+\delta\in(0,1)\) and
\(\delta/c_\tau<r_\tau\). The generalized inverse inequalities give
\[
  F^{-1}(\tau-\delta)
  \le
  G^{-1}(\tau)
  \le
  F^{-1}(\tau+\delta)
\].
Since \(F(q_\tau)=\tau\) and \(f\ge c_\tau\) almost everywhere on the
specified neighborhood,
\[
F(q_\tau-s)\le\tau-c_\tau s,
\qquad
F(q_\tau+s)\ge\tau+c_\tau s,
\qquad 0\le s\le r_\tau.
\]
Taking \(s=\delta/c_\tau\) therefore yields
\[
q_\tau-\frac{\delta}{c_\tau}
\le
F^{-1}(\tau-\delta)
\le
G^{-1}(\tau)
\le
F^{-1}(\tau+\delta)
\le
q_\tau+\frac{\delta}{c_\tau},
\]
which proves the claim.

For CPT, write
\[
p_F^+(z)=P_F\{v^+(Y)>z\},
\qquad
p_F^-(z)=P_F\{v^-(Y)>z\},
\]
and define $p_G^+(z)$ and $p_G^-(z)$ analogously. Since $v$ is nondecreasing, each event appearing above is an upper or lower threshold event in Y. Hence, using CDF values or their left limits as appropriate,
\[
|p_F^\pm(z)-p_G^\pm(z)|
\le
d_\infty(F,G)
=
\delta.
\]
This remains true when $v$ has flat regions.

By the H\"older continuity of the probability distortions,
\[
|w_\pm(p_F^\pm(z))-w_\pm(p_G^\pm(z))|
\le
L_\pm\delta^{\rho_\pm}.
\]
Therefore, using the gain-minus-loss tail-integral representation,
\begin{align}
    |\Tt_{\mathrm{CPT}}(F)-\Tt_{\mathrm{CPT}}(G)|
&\le
\int_0^{M_+}
|w_+(p_F^+(z))-w_+(p_G^+(z))|\,dz \nonumber\\
&\quad+
\int_0^{M_-}
|w_-(p_F^-(z))-w_-(p_G^-(z))|\,dz \nonumber\\
&\le
L_+M_+\delta^{\rho_+}
+
L_-M_-\delta^{\rho_-}.\nonumber
\end{align}
Letting $\rho=\min(\rho_+,\rho_-)$, and noting that $0\le\delta\le1$, we have
$\delta^{\rho_+}\le\delta^\rho,
\delta^{\rho_-}\le\delta^\rho$.
Thus,
\[
|\Tt_{\mathrm{CPT}}(F)-\Tt_{\mathrm{CPT}}(G)|
\le
(L_+M_+ + L_-M_-)\delta^\rho,
\]
which proves the claim.
\end{proof}

\begin{lemma}[H\"older continuity of the standard CPT distortion]
\label{lem:cpt-exponent}
For the commonly used Tversky--Kahneman distortion
\[
  w_\gamma(p)
  =
  \frac{p^\gamma}{
    \{p^\gamma+(1-p)^\gamma\}^{1/\gamma}},
  \qquad 0<\gamma<1,
\]
the endpoint behavior is H\"older.  With gain and loss
exponents \(\gamma_+\) and \(\gamma_-\), respectively,
Proposition~\ref{prop:functional-stability} uses
\(\rho_\Tt=\min(\gamma_+,\gamma_-)\).
More generally, for every \(0<\gamma<1\), there is a finite
\(C_\gamma\) such that
\[
  \abs{w_\gamma(p)-w_\gamma(q)}
  \le
  C_\gamma\abs{p-q}^{\gamma},
  \qquad p,q\in[0,1].
\]
\end{lemma}

\begin{proof}
Let \(D_\gamma(p)=p^\gamma+(1-p)^\gamma\).  Because
\(D_\gamma(p)\ge1\), direct differentiation on \((0,1)\) gives
\[
  w_\gamma'(p)
  =
  D_\gamma(p)^{-1/\gamma-1}
  \left[
    (\gamma-1)p^{2\gamma-1}
    +\gamma p^{\gamma-1}(1-p)^\gamma
    +p^\gamma(1-p)^{\gamma-1}
  \right].
\]
Since \(0<p<1\), the triangle inequality therefore gives
\[
  \abs{w_\gamma'(p)}
  \le
  p^{\gamma-1}+(1-p)^{\gamma-1}.
\]
For \(0\le p<q\le1\), integration therefore yields
\[
  \abs{w_\gamma(q)-w_\gamma(p)}
  \le
  \frac{1}{\gamma}
  \left[
    q^\gamma-p^\gamma
    +(1-p)^\gamma-(1-q)^\gamma
  \right].
\]
The subadditivity of \(t^\gamma\) implies that each bracketed difference is
at most \((q-p)^\gamma\).  Thus the claimed bound holds with
\(C_\gamma=2/\gamma\) on \((0,1)\), and continuity of \(w_\gamma\) extends
it to the endpoints.
\end{proof}

\begin{remark}[Unbounded outcomes]
\label{rem:tails}
For Gaussian, lognormal, or other unbounded outcomes, the bounded-support
statements in Proposition~\ref{prop:functional-stability} require either
outcome clipping that is also used by the algorithm, or explicit tail and
moment assumptions with a truncation remainder.  Alternatively, a
criterion-matched transport metric may be used: \(W_1\) controls the mean and
lower-tail CVaR under the required integrability, whereas variance needs
second-moment control (for example through \(W_2\) plus uniform moment
conditions), and unbounded CPT needs conditions matched to the growth of its
value function.  CDF sup-norm convergence alone cannot control a mean when a
vanishing amount of mass may move arbitrarily far into the tail.  Consequently,
the bounded-support CDF theory does not by itself cover the Gaussian or
lognormal simulations.
\end{remark}

\subsection{Proofs of the Main Theoretical Results}
\label{app:proofs}

We first establish a pointwise oracle inequality for the plug-in treatment rule, which directly yields Theorem \ref{thm:base-rate} and serves as the key step in proving Theorem \ref{thm:final-rate}.

\begin{lemma}[Pointwise oracle inequality]
\label{lem:pointwise-oracle}
For every \(x\),
\begin{equation*}
  0
  \le
  \Q_\star^\Tt(x)-\Q_{\pihat_\Tt(x)}^\Tt(x)
  \le
  2e_n^\Tt(x).
\end{equation*}
This statement permits arbitrary ties among optimal treatments.
\end{lemma}

\begin{proof}
Choose any \(a^\star\in\arg\max_a Q_a^{\mathcal T}(x)\).  Since
\(\Qhat_{\pihat_\Tt(x)}^\Tt(x)\ge\Qhat_{a^\star}^\Tt(x)\),
\begin{align*}
  \Q_\star^\Tt(x)-\Q_{\pihat_\Tt(x)}^\Tt(x)
  &=
  \Q_{a^\star}^\Tt(x)-\Q_{\pihat_\Tt(x)}^\Tt(x)
  \\
  &\le
  \abs{\Qhat_{a^\star}^\Tt(x)-\Q_{a^\star}^\Tt(x)}
  +
  \abs{
    \Qhat_{\pihat_\Tt(x)}^\Tt(x)
    -
    \Q_{\pihat_\Tt(x)}^\Tt(x)
  }
  \\
  &\le2e_n^\Tt(x).
\end{align*}
\end{proof}

\begin{proof}[Proof of Theorem \ref{thm:base-rate}] By the definition of policy regret and Lemma~\ref{lem:pointwise-oracle},
\[
\mathcal R_{\mathcal T}(\widehat\pi_{\mathcal T})
=
\E_X\left[
Q_\star^{\mathcal T}(X)
-
Q_{\widehat\pi_{\mathcal T}(X)}^{\mathcal T}(X)
\right]\le
2e_{n,\infty}^{\mathcal T}.
\]
Under Assumption~\ref{ass:holder}, on the event
$\delta_{n,\infty}^{\mathcal T}\le\epsilon_{\mathcal T}$,
$e_{n,\infty}^{\mathcal T}
\le
L_{\mathcal T}
(\delta_{n,\infty}^{\mathcal T})^{\rho_{\mathcal T}}$.
Combining the two inequalities gives
$\mathcal R_{\mathcal T}(\widehat\pi_{\mathcal T})
\le
2L_{\mathcal T}
(\delta_{n,\infty}^{\mathcal T})^{\rho_{\mathcal T}}$,
as claimed.
\end{proof}

The following lemma gives a general transfer from pointwise criterion estimation error to policy regret under the local boundary-mass condition. It serves as the common basis for both Theorem~\ref{thm:final-rate} and the integrated-error extension below.

\begin{lemma}[General margin transfer]
\label{lem:general-boundary-transfer}
Under Assumption~\ref{ass:margin}, for every realization of the training
sample and every \(0<t\le t_0\),
\begin{equation}
  \Reg_\Tt(\pihat_\Tt)
  \le
  t\Psi_\Tt(t)
  +
  2\E_X\!\left[
    e_n^\Tt(X)
    \1\{e_n^\Tt(X)>t/2\}
  \right].
  \label{eq:general-lq-transfer}
\end{equation}
\end{lemma}

\begin{proof}
Let
\[
  G_n(X)
  =
  \Q_\star^\Tt(X)-\Q_{\pihat_\Tt(X)}^\Tt(X).
\]
On \(\{0<G_n\le t\}\), the selected treatment is strictly suboptimal and
\(\Delta_\Tt(X)\le G_n(X)\le t\).  Hence
\[
  \E[G_n\1\{0<G_n\le t\}]
  \le
  t\Pp\{\Delta_\Tt(X)\le t\}
  \le
  t\Psi_\Tt(t).
\]
On \(\{G_n>t\}\), Lemma~\ref{lem:pointwise-oracle} implies
\(G_n\le2e_n^\Tt(X)\) and \(e_n^\Tt(X)>t/2\).  Adding the two pieces proves
\eqref{eq:general-lq-transfer}.
\end{proof}

\begin{proof}[Proof of Theorem~\ref{thm:final-rate}]
Under Assumption~\ref{ass:holder}, on the event
$\delta_{n,\infty}^{\mathcal T}\leq\epsilon_{\mathcal T},
e_{n,\infty}^{\mathcal T}
\leq
L_{\mathcal T}
(\delta_{n,\infty}^{\mathcal T})^{\rho_{\mathcal T}}$.

If \(\delta_{n, \infty}^{\mathcal T}=0\), Theorem \ref{thm:base-rate} gives zero regret directly.
Otherwise, apply Lemma~\ref{lem:general-boundary-transfer} with
$t
=
2L_{\mathcal T}
(\delta_{n,\infty}^{\mathcal T})^{\rho_{\mathcal T}}$.
By the assumed condition
$2L_{\mathcal T}
(\delta_{n,\infty}^{\mathcal T})^{\rho_{\mathcal T}}
\leq t_0$,
this is an admissible choice of $t$. Moreover,
$e_{n}^{\mathcal T}(X)
\leq
e_{n,\infty}^{\mathcal T}
\leq
L_{\mathcal T}
(\delta_{n,\infty}^{\mathcal T})^{\rho_{\mathcal T}}
=
t/2$
for every $X$. Hence,
$\mathbf 1
\left\{
e_n^{\mathcal T}(X)>t/2
\right\}
=0$,
and the second term in Lemma~\ref{lem:general-boundary-transfer} vanishes. 
Therefore,
$\mathcal R_{\mathcal T}(\widehat\pi_{\mathcal T})
\leq
2L_{\mathcal T}
(\delta_{n,\infty}^{\mathcal T})^{\rho_{\mathcal T}}
\Psi_{\mathcal T}
\left\{
2L_{\mathcal T}
(\delta_{n,\infty}^{\mathcal T})^{\rho_{\mathcal T}}
\right\}$,
which proves Theorem~\ref{thm:final-rate}.
\end{proof}

The main-text results connect uniform conditional distribution estimation error to policy regret through the H\"older stability condition. Independently of this distribution-to-score transfer, Lemma~\ref{lem:general-boundary-transfer} also yields regret guarantees directly from integrated criterion estimation errors. We record these score-level consequences under the familiar polynomial special case of the boundary-mass condition.

\begin{corollary}[Polynomial margin and \(L_q\) score errors]
\label{cor:polynomial-margin}
Suppose that, for \(0<t\le t_0\),
\begin{equation}
  \Psi_\Tt(t)\le C_\Delta t^\kappa,
  \qquad \kappa\ge0.
  \label{eq:polynomial-margin}
\end{equation}
Then the following statements hold.
\begin{enumerate}
  \item For \(1<q<\infty\), with
  \(m_{n,q}=\norm{e_{n}^\Tt}_{L_q(P_X)}\),
  \begin{equation}
    \Reg_\Tt(\pihat_\Tt)
    \le
    C_{q,\kappa,C_\Delta}
    m_{n,q}^{\,q(1+\kappa)/(q+\kappa)}
    \label{eq:lq-rate}
  \end{equation}
  whenever \(m_{n,q}\) is sufficiently small.  Consequently,
  \(m_{n,q}=\op(\rho_n)\) implies
  \[
    \Reg_\Tt(\pihat_\Tt)
    =
    \op\!\left(
      \rho_n^{\,q(1+\kappa)/(q+\kappa)}
    \right).
  \]

  \item If \(2m_{n,\infty}\le t_0\) and
  \(m_{n,\infty}=\norm{e_{n}^\Tt}_{L_\infty(P_X)}=e_{n,\infty}^\Tt\), then
  \begin{equation}
    \Reg_\Tt(\pihat_\Tt)
    \le
    2m_{n,\infty}\Psi_\Tt(2m_{n,\infty})
    \le
    2^{1+\kappa}C_\Delta
    m_{n,\infty}^{1+\kappa}.
    \label{eq:linfty-margin}
  \end{equation}

  \item With only an \(L_1(P_X)\) score bound, the always-valid conclusion
  remains
  \[
    \Reg_\Tt(\pihat_\Tt)
    \le2m_{n,1}.
  \]
\end{enumerate}
\end{corollary}

\begin{proof}
For \(1<q<\infty\), H\"older's inequality followed by Markov's inequality
gives
\[
  \E[e_n^\Tt(X)\1\{e_n^\Tt(X)>t/2\}]
  \le
  2^{q-1}m_{n,q}^q t^{1-q}.
\]
Thus Lemma~\ref{lem:general-boundary-transfer} gives
\[
  \Reg_\Tt(\pihat_\Tt)
  \le
  C_\Delta t^{1+\kappa}
  +
  2^q m_{n,q}^q t^{1-q}.
\]
Taking \(t\) proportional to
\(m_{n,q}^{q/(q+\kappa)}\) proves \eqref{eq:lq-rate}.
For \(q=\infty\), a strictly suboptimal selection implies both
\(G_n(X)\le2m_{n,\infty}\) and
\(\Delta_\Tt(X)\le2m_{n,\infty}\), yielding
\eqref{eq:linfty-margin}. 
\end{proof}

\begin{remark}[Interpretation and limiting cases]
\label{rem:margin-cases}
When \(\kappa=0\), the polynomial condition supplies no rate acceleration,
and all displays reduce to the first-stage score rate.  Under a hard gap,
\(\Psi_\Tt(t)=0\) for \(t<t_0\), a uniformly accurate plug-in rule has zero
regret once \(2m_{n,\infty}<t_0\).  Without any restriction on mass near the
decision boundary, a score perturbation of size \(\varepsilon\) may change
the selected treatment on a region with gaps of order \(\varepsilon\), so a
general improvement over order \(\varepsilon\) is impossible.  This is the
same role played by low-noise assumptions in classical plug-in
classification \citep{audibert2007fast}.
\end{remark}

\subsection{Estimator-Specific Guarantees}

\subsubsection{A generic weighted conditional-CDF estimator}
\label{app:weighted-cdf}

Both kernel smoothers and forest estimators can be written as weighted
empirical CDFs.  For a treatment \(a\), let
\(\mathcal I_a=\{i:A_i=a\}\), and consider
\begin{equation}
  \Fhat_{a,W}(y\mid x)
  =
  \sum_{i\in\mathcal I_a}
  W_{i,a}(x)\1(Y_i\le y),
  \qquad
  W_{i,a}(x)\ge0,
  \qquad
  \sum_{i\in\mathcal I_a}W_{i,a}(x)=1.
  \label{eq:weighted-cdf}
\end{equation}
This definition is used on
\(\mathcal E_{n,A}=\{\mathcal I_a\ne\varnothing\text{ for all }a\}\);
outside that event, set each unavailable-arm estimate to a prespecified fixed
CDF belonging to the domain of \(\Tt\).  Under overlap,
\[
  \Pp(\mathcal E_{n,A}^c)
  \le
  2(1-\eta)^n
  \longrightarrow0.
\]
All weighted-estimator statements below are proved on
\(\mathcal E_{n,A}\).  For bookkeeping, set the weights, localization bias,
and squared weight mass to arbitrary bounded values off this event.  Its
exponentially small probability leaves the unconditional \(O_{\Pp}\)
conclusions unchanged.
Let \(\mathcal H_n\) contain all covariates, treatment labels, algorithmic
randomization, structure-sample observations, and other information used to
construct the weights, but not the estimation outcomes appearing in
\eqref{eq:weighted-cdf}.

\begin{assumption}[Global honesty for the estimation outcomes]
\label{ass:global-honesty}
Conditionally on \(\mathcal H_n\), the weights in
\eqref{eq:weighted-cdf} are fixed, and the estimation outcomes appearing in
that equation are independent with
\[
  \E[\1(Y_i\le y)\mid\mathcal H_n]
  =
  F_a(y\mid X_i),
  \qquad i\in\mathcal I_a.
\]
\end{assumption}

For kernels, Assumption~\ref{ass:global-honesty} holds because the weights
depend only on \(X_i\) and \(A_i\).  For forests, it holds for a global
sample-splitting construction in which a structure sample builds all
partitions and an independent estimation sample supplies the weighted
outcomes.  Standard per-tree honesty does not automatically imply
Assumption~\ref{ass:global-honesty} for the final aggregate weights, because
one observation can be used for splitting in some trees and for leaf
estimation in others.  This fixed-weight conditional-independence requirement
is therefore stronger than the per-tree honesty used in generalized forest
constructions \citep{athey2019generalized}.

Define the oracle weighted target, localization bias, squared weight mass,
and effective sample size:
\begin{align*}
  F_{a,W}(y\mid x)
  &=
  \sum_{i\in\mathcal I_a}
  W_{i,a}(x)F_a(y\mid X_i),
  \\
  B_{a,n}(x)
  &=
  \sum_{i\in\mathcal I_a}
  W_{i,a}(x)
  \sup_y\abs{F_a(y\mid X_i)-F_a(y\mid x)},
  \\
  S_{a,n}(x)
  &=
  \sum_{i\in\mathcal I_a}W_{i,a}^2(x),
  \qquad
  n_{\mathrm{eff},a}(x)=S_{a,n}^{-1}(x).
\end{align*}
The exact decomposition
\begin{equation}
  \Fhat_{a,W}-F_a
  =
  (\Fhat_{a,W}-F_{a,W})
  +
  (F_{a,W}-F_a)
  \label{eq:weighted-decomposition}
\end{equation}
separates stochastic fluctuation from localization bias.

\begin{proposition}[Fixed-query weighted-CDF concentration]
\label{prop:weighted-concentration}
On \(\mathcal E_{n,A}\), under Assumption~\ref{ass:global-honesty}, for every
fixed \(a,x\) and every finite \(1\le r<\infty\), there is a constant
\(C_r<\infty\) such that
\begin{equation}
  \left(
  \E\!\left[
    \left.
    \sup_y
    \abs{\Fhat_{a,W}(y\mid x)-F_{a,W}(y\mid x)}^r
    \right|\mathcal H_n
  \right]
  \right)^{1/r}
  \le
  C_r S_{a,n}^{1/2}(x).
  \label{eq:weighted-moment}
\end{equation}
Consequently,
\begin{equation}
  \sup_y
  \abs{\Fhat_{a,W}(y\mid x)-F_a(y\mid x)}
  =
  O_{\Pp}\!\left(
    B_{a,n}(x)+S_{a,n}^{1/2}(x)
  \right)
  \label{eq:weighted-pointwise}
\end{equation}
conditionally on the weight construction.
\end{proposition}

\begin{proof}
The bias part of \eqref{eq:weighted-decomposition} is bounded by
\(B_{a,n}(x)\).  For the stochastic part, write
\[
  Z_i(y)=\1(Y_i\le y)-F_a(y\mid X_i).
\]
Conditional \(L_r\) symmetrization \citep{vanderVaart1996weak} gives
\[
  \left\|
    \sup_y\abs{\sum_iW_{i,a}(x)Z_i(y)}
  \right\|_{L_r(\Pp\mid\mathcal H_n)}
  \le
  2\left\|
    \sup_y\abs{
      \sum_iW_{i,a}(x)\epsilon_i\1(Y_i\le y)
    }
  \right\|_{L_r(\Pp\times\Pp_\epsilon\mid\mathcal H_n)},
\]
where the \(\epsilon_i\)'s are independent Rademacher variables.  Conditional
on the realized outcomes, sort them increasingly.  Threshold sets are
prefixes of this ordering (ties only reduce the collection), so the last
supremum is bounded by the maximum absolute partial sum of the independent
increments \(W_{(i),a}(x)\epsilon_{(i)}\).  For \(r>1\), Doob's maximal
inequality followed by the Khintchine inequality bounds its \(L_r\) norm by
\(C_r\{\sum_iW_{i,a}^2(x)\}^{1/2}\); for \(r=1\), use the \(L_2\) bound and
Jensen's inequality.  This proves \eqref{eq:weighted-moment}, and
\eqref{eq:weighted-pointwise} follows from
\eqref{eq:weighted-decomposition}.
\end{proof}

\begin{theorem}[Integrated weighted-CDF error]
\label{thm:weighted-lq}
Let \(X\sim P_X\) be independent of the training data, let
\[
  \delta_{n,W}(x)
  =
  \max_{a\in\A}\sup_y
  \abs{\Fhat_{a,W}(y\mid x)-F_a(y\mid x)},
\]
and, for finite \(1\le r<\infty\), define
\begin{align*}
  b_{n,r}
  &=
  \norm{\max_{a\in\A}B_{a,n}}_{L_r(P_X)},
  &
  s_{n,r}
  &=
  \norm{\max_{a\in\A}S_{a,n}^{1/2}}_{L_r(P_X)}.
\end{align*}
If
\[
  b_{n,r}=\op(\bar b_{n,r}),
  \qquad
  s_{n,r}=\op(\bar s_{n,r}),
\]
then, under Assumption~\ref{ass:global-honesty},
\begin{equation}
  \norm{\delta_{n,W}}_{L_r(P_X)}
  =
  \op(\bar b_{n,r}+\bar s_{n,r}).
  \label{eq:integrated-cdf-rate}
\end{equation}
\end{theorem}

\begin{proof}
Apply Proposition~\ref{prop:weighted-concentration} conditionally at each
\(x\), integrate its \(r\)-th moment with respect to \(P_X\), and use the
finite-action inequality
\(\max_a z_a^r\le\sum_a z_a^r\).  Conditional Markov's inequality yields
the stochastic order in \eqref{eq:integrated-cdf-rate}; the localization
bias is handled by the triangle inequality.
\end{proof}

\begin{corollary}[Weighted estimator regret]
\label{cor:weighted-regret}
Suppose Assumption~\ref{ass:holder} holds with
\(d_\Tt=d_\infty\). Assume additionally that there exists
\(M_\Tt<\infty\) such that
\[
  e_n^\Tt(x)\le M_\Tt
\]
uniformly in \(x\), almost surely. Assume that all rate sequences used
below converge to zero. Let \(1\le q<\infty\) satisfy
\(\rho_\Tt q\ge1\), and apply
Theorem~\ref{thm:weighted-lq} with \(r=\rho_\Tt q\).  Then
\begin{equation}
  \norm{e_n^\Tt}_{L_q(P_X)}
  =
  \op\!\left[
    \{\bar b_{n,\rho_\Tt q}
      +\bar s_{n,\rho_\Tt q}\}^{\rho_\Tt}
  \right].
  \label{eq:weighted-score-rate}
\end{equation}
Without a margin condition, applying Theorem~\ref{thm:weighted-lq} with
\(r=1\), Jensen's inequality, and \(0\le\delta_{n,W}\le1\) gives
\begin{equation}
  \Reg_\Tt(\pihat_\Tt)
  =
  \op\!\left[
    \{\bar b_{n,1}+\bar s_{n,1}\}^{\rho_\Tt}
  \right].
  \label{eq:weighted-no-margin-rate}
\end{equation}
Under \eqref{eq:polynomial-margin} and \(1<q<\infty\),
\begin{equation}
  \Reg_\Tt(\pihat_\Tt)
  =
  \op\!\left[
    \{\bar b_{n,\rho_\Tt q}
      +\bar s_{n,\rho_\Tt q}\}^{
       \rho_\Tt q(1+\kappa)/(q+\kappa)}
  \right].
  \label{eq:weighted-margin-rate}
\end{equation}
A uniform-in-\(x\) analogue requires a separate entropy or complexity
condition for \(\{W_a(x):x\in\X\}\) and does not follow by setting
\(q=\infty\) in the finite-\(r\) integrated theorem.
\end{corollary}

\begin{proof}
Write \(\delta_n=\delta_{n,W}\). On
\(\{\delta_n(X)\le\epsilon_\Tt\}\),
Assumption~\ref{ass:holder} gives
\[
  e_n^\Tt(X)
  \le
  L_\Tt\delta_n(X)^{\rho_\Tt}.
\]
On the complementary event, use \(e_n^\Tt(X)\le M_\Tt\).
Therefore, whenever \(\rho_\Tt q\ge1\),
\[
\begin{aligned}
  \norm{e_n^\Tt}_{L_q(P_X)}^q
  &\le
  L_\Tt^q
  \E_X\!\left[\delta_n(X)^{\rho_\Tt q}\right]
  +
  M_\Tt^q
  \Pp_X\{\delta_n(X)>\epsilon_\Tt\} \\
  &\le
  \left(
    L_\Tt^q+
    M_\Tt^q\epsilon_\Tt^{-\rho_\Tt q}
  \right)
  \norm{\delta_n}_{L_{\rho_\Tt q}(P_X)}^{\rho_\Tt q}.
\end{aligned}
\]
This proves \eqref{eq:weighted-score-rate}. Moreover, the pointwise
oracle inequality, Jensen's inequality, and Markov's inequality give
\[
\begin{aligned}
  \Reg_\Tt(\pihat_\Tt)
  &\le
  2\norm{e_n^\Tt}_{L_1(P_X)} \\
  &\le
  2L_\Tt
  \norm{\delta_n}_{L_1(P_X)}^{\rho_\Tt}
  +
  2M_\Tt\epsilon_\Tt^{-1}
  \norm{\delta_n}_{L_1(P_X)},
\end{aligned}
\]
which proves \eqref{eq:weighted-no-margin-rate}. Finally, combining
\eqref{eq:weighted-score-rate} with
Corollary~\ref{cor:polynomial-margin} proves
\eqref{eq:weighted-margin-rate}.
\end{proof}

\subsubsection{Kernel conditional-CDF estimation}
\label{app:kernel}

For \(K_h(u)=h^{-p}K(u/h)\), consider the local-constant estimator
\begin{equation}
  \Fhat_a^K(y\mid x)
  =
  \frac{
    \sum_{i=1}^n
    K_h(X_i-x)\1(A_i=a)\1(Y_i\le y)
  }{
    \sum_{i=1}^n
    K_h(X_i-x)\1(A_i=a)
  }.
  \label{eq:kernel-cdf}
\end{equation}
If the denominator is zero, define the estimate to be a prespecified fixed CDF
belonging to the domain of \(\Tt\).
The proof below shows that this fallback is unused uniformly on \(\X_0\) with
probability tending to one.

\begin{assumption}[Uniform kernel conditions]
\label{ass:kernel}
The following conditions hold.
\begin{enumerate}
  \item \(\X_0\subset\X\) is compact and
  \(\Pp_X(X\in\X_0)=1\).  The support is regular enough that the
  population kernel denominator is uniformly positive as specified below.

  \item For each \(a\), the joint design density
  \(g_a(x)=f_X(x)e_a(x)\) is bounded above, and, for some
  \(c_D,h_0>0\),
  \[
    \inf_{\substack{a\in\A,\,x\in\X_0\\0<h\le h_0}}
    \int K(v)g_a(x+hv)\1(x+hv\in\X_0)\,dv
    \ge c_D.
  \]
  A density bounded away from zero on a regular compact support is a
  standard sufficient condition.

  \item \(K\) is bounded, nonnegative, compactly supported,
  \(\int K(u)\,du=1\), and its translates and dilations form a VC-type
  class.  The corresponding product classes
  \[
    (X,A,Y)\mapsto
    K_h(X-x)\1(A=a)\1(Y\le y)
  \]
  and the denominator class without \(\1(Y\le y)\) are also assumed
  VC-type uniformly over \((a,x,y,h)\).  This follows from standard
  permanence properties for product kernels of bounded variation and the
  threshold class.

  \item For some \(0<\beta\le1\) and \(L_F<\infty\),
  \begin{equation}
    \sup_{a\in\A}\sup_{y\in\R}
    \abs{F_a(y\mid x)-F_a(y\mid x')}
    \le
    L_F\norm{x-x'}_2^\beta
    \label{eq:cdf-holder}
  \end{equation}
  for all \(x,x'\in\X_0\).

  \item \(h=h_n\to0\) and
  \(nh_n^p/\log n\to\infty\).
\end{enumerate}
\end{assumption}

The condition \(\beta\le1\) corresponds to the local-constant estimator in
\eqref{eq:kernel-cdf}.  Exploiting smoothness beyond first order requires an
appropriate higher-order or local-polynomial conditional-distribution
estimator and its own shape-validity argument.

\begin{theorem}[Uniform kernel conditional-CDF rate]
\label{thm:kernel-cdf}
Under Assumptions~\ref{ass:identification} and \ref{ass:kernel},
\begin{equation}
  \max_{a\in\A}
  \sup_{x\in\X_0}\sup_{y\in\R}
  \abs{\Fhat_a^K(y\mid x)-F_a(y\mid x)}
  =
  \op\!\left(
    h_n^\beta
    +
    \sqrt{\frac{\log n}{nh_n^p}}
  \right).
  \label{eq:kernel-uniform-rate}
\end{equation}
\end{theorem}

\begin{proof}
For each \(a\), write the numerator and denominator of
\eqref{eq:kernel-cdf}, divided by \(n\), as \(N_{n,a}(x,y)\) and
\(D_{n,a}(x)\).  Their population counterparts are
\begin{align*}
  \bar N_{a,h}(x,y)
  &=
  \int_{x+hv\in\X_0}
  K(v)g_a(x+hv)F_a(y\mid x+hv)\,dv,
  \\
  \bar D_{a,h}(x)
  &=
  \int_{x+hv\in\X_0}K(v)g_a(x+hv)\,dv.
\end{align*}
The design-density lower bound and nonnegativity of \(K\) imply
\(\inf_{a,x}\bar D_{a,h}(x)>c>0\).  Moreover,
\eqref{eq:cdf-holder} gives, uniformly in \(a,x,y\),
\[
  \abs{
    \bar N_{a,h}(x,y)/\bar D_{a,h}(x)
    -
    F_a(y\mid x)
  }
  \le
  C h^\beta.
\]
The VC-type assumption and the bounded, compactly supported kernel imply the
uniform empirical-process bounds
\begin{align*}
  \sup_{a,x,y}
  \abs{N_{n,a}(x,y)-\bar N_{a,h}(x,y)}
  &=
  \op\!\left(
    \sqrt{\frac{\log n}{nh^p}}
  \right),
  \\
  \sup_{a,x}
  \abs{D_{n,a}(x)-\bar D_{a,h}(x)}
  &=
  \op\!\left(
    \sqrt{\frac{\log n}{nh^p}}
  \right).
\end{align*}
More explicitly, the corresponding maximal inequality also contributes an
envelope term of order \(\log n/(nh^p)\), which is dominated by the displayed
square-root term under \(nh^p/\log n\to\infty\).  These are standard
uniform-in-bandwidth kernel empirical-process bounds
\citep{einmahl2005uniform}.
The second display and \(nh^p/\log n\to\infty\) ensure that the empirical
denominator is uniformly bounded away from zero with probability tending to
one.  A ratio expansion and the uniform bias bound prove
\eqref{eq:kernel-uniform-rate}.  This is the standard bias-stochastic
structure for kernel conditional-distribution estimation
\citep{hall1999nonparametric}.
\end{proof}

\begin{corollary}[Kernel score and regret rates]
\label{cor:kernel-regret}
Suppose Assumption~\ref{ass:holder} holds with
\(d_\Tt=d_\infty\).
Let
\[
  r_{n,K}
  =
  h_n^\beta
  +
  \sqrt{\frac{\log n}{nh_n^p}}.
\]
Then
\begin{align*}
  \norm{e_n^\Tt}_{L_\infty(P_X)}
  &=
  \op(r_{n,K}^{\rho_\Tt}),
  \\
  \Reg_\Tt(\pihat_\Tt^K)
  &=
  \op(r_{n,K}^{\rho_\Tt})
  \quad\text{without a margin condition}.
\end{align*}
Under \eqref{eq:polynomial-margin},
\begin{equation*}
  \Reg_\Tt(\pihat_\Tt^K)
  =
  \op(r_{n,K}^{\rho_\Tt(1+\kappa)}).
\end{equation*}
The log-optimized bandwidth and resulting rate are
\begin{align*}
  h_n
  &\asymp
  \left(\frac{\log n}{n}\right)^{1/(2\beta+p)},
  \\
  r_{n,K}
  &\asymp
  \left(\frac{\log n}{n}\right)^{\beta/(2\beta+p)},
  \\
  \Reg_\Tt(\pihat_\Tt^K)
  &=
  \op\!\left[
    \left(\frac{\log n}{n}\right)^{
      \beta\rho_\Tt(1+\kappa)/(2\beta+p)}
  \right].
\end{align*}
\end{corollary}

\begin{remark}[An interior-only kernel bound]
\label{rem:kernel-boundary}
If Theorem~\ref{thm:kernel-cdf} is proved only on
\(\X_0\) but \(\Pp(X\notin\X_0)>0\), it does not by itself imply a
population regret rate.  If all scores are bounded by \(M_\Tt\), one may
instead add the explicit remainder
\[
  \Reg_{\Tt,\X_0}(\pihat_\Tt)
  :=
  \E_X\!\left[
    \{\Q_\star^\Tt(X)-\Q_{\pihat_\Tt(X)}^\Tt(X)\}
    \1(X\in\X_0)
    \mid\D_n
  \right],
\]
and
\[
  \Reg_\Tt(\pihat_\Tt)
  \le
  \Reg_{\Tt,\X_0}(\pihat_\Tt)
  +
  2M_\Tt\Pp(X\notin\X_0).
\]
For a sequence \(\X_{0,n}\), the outside probability must be negligible
relative to the claimed rate.
\end{remark}

\subsubsection{Distributional Random Forest}
\label{app:forest}

A forest conditional-CDF estimator has the form
\begin{equation}
  \Fhat_a^{F}(y\mid x)
  =
  \sum_{i:A_i=a}w_{i,a}(x)\1(Y_i\le y),
  \qquad
  w_{i,a}(x)\ge0,
  \qquad
  \sum_{i:A_i=a}w_{i,a}(x)=1.
  \label{eq:forest-cdf}
\end{equation}
The formal result below covers treatment-specific fits, or a joint fit whose
query at treatment \(a\) produces weights that are zero outside arm \(a\) and
are normalized within that arm.  A generic joint fit that assigns cross-arm
weights requires a separate formulation on the joint \((X,A)\) space.  Once
estimated, the treatment-specific distributions can be reused for every
downstream criterion without refitting criterion-specific objectives.

Suppose a global structure sample constructs all partitions and an
independent estimation sample supplies the outcomes in
\eqref{eq:forest-cdf}.  Then Assumption~\ref{ass:global-honesty} holds and
Theorem~\ref{thm:weighted-lq} applies.  In particular, define
\begin{align*}
  B_{a,n}^{F}(x)
  &=
  \sum_{i:A_i=a}w_{i,a}(x)
  \sup_y\abs{F_a(y\mid X_i)-F_a(y\mid x)},
  \\
  S_{a,n}^{F}(x)
  &=
  \sum_{i:A_i=a}w_{i,a}^2(x).
\end{align*}
If \eqref{eq:cdf-holder} holds, the more informative weighted localization
bound is
\begin{equation*}
  B_{a,n}^{F}(x)
  \le
  L_F
  \sum_{i:A_i=a}
  w_{i,a}(x)\norm{X_i-x}_2^\beta.
\end{equation*}
This is preferable to the maximum radius of all observations with positive
weight, which can be driven by a negligible forest weight.

\begin{corollary}[Integrated forest regret]
\label{cor:forest-regret}
Assume global sample splitting, binary treatment, CDF-supnorm criterion
stability with exponent \(\rho_\Tt\), and the integrated weight controls
below for the required finite values of \(r\):
\[
  \norm{\max_aB_{a,n}^{F}}_{L_r(P_X)}
  =
  \op(\bar b_{n,r}),
  \qquad
  \norm{\max_a(S_{a,n}^{F})^{1/2}}_{L_r(P_X)}
  =
  \op(\bar s_{n,r}).
\]
Then Corollary~\ref{cor:weighted-regret} gives an explicit regret bound.
In particular, for a Lipschitz criterion and \(1<q<\infty\),
\begin{equation*}
  \Reg_\Tt(\pihat_\Tt^{F})
  =
  \op\!\left[
    \{\bar b_{n,q}+\bar s_{n,q}\}^{
      q(1+\kappa)/(q+\kappa)}
  \right]
\end{equation*}
under \eqref{eq:polynomial-margin}; the no-margin bound is
\eqref{eq:weighted-no-margin-rate} with the forest weight quantities.
\end{corollary}

The published DRF theory of \citet{cevid2022distributional} establishes a
fixed-\(x\) polynomial rate for a conditional RKHS mean embedding under
subsampling, honesty, regularity, symmetry, and random-splitting conditions.
Renaming their subsample exponent to \(\theta\), if
\(s_n\asymp n^\theta\), their Theorem~2 gives
\begin{equation*}
  \norm{\widehat\mu_n(x)-\mu(x)}_{\mathcal H}
  =
  \op(n^{-\gamma_{\mathrm{DRF}}}),
  \qquad
  \gamma_{\mathrm{DRF}}
  =
  \frac12\min\!\left\{
    1-\theta,
    \frac{\log\{(1-\alpha_{\mathrm{reg}})^{-1}\}}
         {\log(\alpha_{\mathrm{reg}}^{-1})}
    \frac{\pi_{\mathrm{split}}}{p}\theta
  \right\},
\end{equation*}
for each fixed test point, under their assumptions on the design density,
bounded kernel, conditional embedding, and forest construction.  Here
\(\alpha_{\mathrm{reg}}\) and \(\pi_{\mathrm{split}}\) denote their
tree-regularity and random-split constants.  Their CDF and quantile
corollaries establish consistency at appropriate continuity points.  Those
results do not directly provide a
uniform-in-\((x,y)\) CDF sup-norm rate, nor do they establish automatic
adaptation to an unknown intrinsic dimension.

Therefore, if the implementation uses standard per-tree honest DRF
rather than the globally split construction above, define the
pointwise criterion-matched distribution error by
\[
  \delta_{n,\mathrm{DRF}}^\Tt(x)
  :=
  \max_{a\in\A}
  d_\Tt\!\left\{
    \Fhat_a^{\mathrm{DRF}}(\cdot\mid x),
    F_a(\cdot\mid x)
  \right\}.
\]
A defensible generic assumption, for some finite \(1\le r<\infty\),
is
\begin{equation*}
  \norm{\delta_{n,\mathrm{DRF}}^\Tt}_{L_r(P_X)}
  =
  \op(r_{n,\mathrm{DRF}}),
  \qquad
  r_{n,\mathrm{DRF}}\to0.
\end{equation*}
This distribution-error condition does not by itself imply a regret
rate. It must first be transferred to a score-error rate through the
criterion-stability condition, after which
Lemma~\ref{lem:pointwise-oracle} and
Corollary~\ref{cor:polynomial-margin} can be applied.

More explicitly, suppose Assumption~\ref{ass:holder} holds and that
the bounded score-error condition used in
Corollary~\ref{cor:weighted-regret} is satisfied. If
\(r\ge\rho_\Tt\), the same good-set/bad-set argument gives
\[
  \norm{e_n^\Tt}_{L_{r/\rho_\Tt}(P_X)}
  =
  \op(r_{n,\mathrm{DRF}}^{\rho_\Tt}).
\]
Consequently,
\[
  \Reg_\Tt(\pihat_\Tt^{\mathrm{DRF}})
  =
  \op(r_{n,\mathrm{DRF}}^{\rho_\Tt})
\]
without a margin condition. If \(r>\rho_\Tt\) and
\eqref{eq:polynomial-margin} holds, then
\[
  \Reg_\Tt(\pihat_\Tt^{\mathrm{DRF}})
  =
  \op\!\left(
    r_{n,\mathrm{DRF}}^{
      \rho_\Tt r(1+\kappa)/(r+\rho_\Tt\kappa)}
  \right).
\]
A separately established genuine uniform bound
\[
  \sup_x\delta_{n,\mathrm{DRF}}^\Tt(x)
  =
  \op(r_{n,\mathrm{DRF}})
\]
can instead be combined directly with
Theorems~\ref{thm:base-rate}--\ref{thm:final-rate}. Under the
polynomial margin condition, this gives
\[
  \Reg_\Tt(\pihat_\Tt^{\mathrm{DRF}})
  =
  \op\!\left(
    r_{n,\mathrm{DRF}}^{\rho_\Tt(1+\kappa)}
  \right).
\]
The rate \(r_{n,\mathrm{DRF}}\) should not be replaced by a closed-form
intrinsic-dimension expression unless that localization result is
separately proved for the actual forest construction.

\begin{remark}[Uniform forest bounds need weight-class complexity]
\label{rem:forest-uniform}
Pointwise control through \(B_{a,n}^{F}(x)\) and \(S_{a,n}^{F}(x)\) does
not automatically yield a supremum over \(x\).  A uniform result additionally
requires a complexity bound for the collection of weight vectors indexed by
\(x\).  For example, if a conditional entropy or log-cardinality
\(\ell_n\) controls that class, a typical high-level uniform CDF bound has
the form
\[
  \sup_{a,x,y}
  \abs{\Fhat_a^{F}(y\mid x)-F_a(y\mid x)}
  =
  \op\!\left(
    b_n+\sqrt{s_n\ell_n}
  \right),
\]
where \(b_n\) bounds the uniform localization bias and \(s_n\) bounds the
uniform squared weight mass.  Taking \(\ell_n=O(\log n)\) requires a
separate complexity argument; it does not follow from honesty alone.
\end{remark}

\section{Additional experiment details}
\label{app:exp}
\subsection{Compute resource} 
All simulations were run on
UNC's Longleaf cluster.
Non-generative methods were fit in R 4.5.0
(\texttt{drf} 1.3.1, \texttt{quantreg} 6.1, \texttt{DynTxRegime} 4.16,
\texttt{glmnet} 4.1-10) with 8 CPU cores and 32\,GB of memory per job.
Conditional DDPM was trained on a single NVIDIA A100 (40\,GB) GPU with
PyTorch 2.6.0 (CUDA 12.4), and GCDS was run with 4 CPU cores and
32\,GB of memory using TensorFlow 2.16.2; both generative pipelines
were driven from R 4.4.0.

The treatment-interactive experiment was run locally on macOS using R 4.4.1. The ACTG175 evaluation and runtime measurements were performed on an Apple M3 Pro using R 4.4.1, with numerical-library and DRF thread counts set to one.

\subsection{Full Experimental Results}
\label{app:full-simulation-results}
Tables~\ref{tab:s1_p10_full}, \ref{tab:s2_p10_full}, and \ref{tab:s3_p10_full} report all five criteria and all CQR and $\ell_2$-CVaR training levels for the same 100 paired replications as the main tables. Higher-dimensional results with $p=20,50,100,300$ are also reported below.

In Setting~1, $\ell_1$-PLS has the highest Value and lowest Err among non-oracle methods for all reported criteria at $p=20,50,100,300$. Within the distribution-first family, Conditional DDPM performs best at $p=20$, whereas DRF performs best at $p=50,100,300$. In Setting~2, DRF has the highest Value and lowest Err for all five criteria at $p=20$. At $p=300$, DRF has the highest Mean Value, while $\ell_1$-PLS has the lowest Err for all criteria and the highest quantile and CVaR Values. Within the distribution-first family at $p=300$, DRF has the highest Values and lowest mean-targeted Err, whereas GCDS has the lowest tail-targeted Err among its evaluable repetitions. All methods have 100 evaluable repetitions except GCDS in Setting~2 at $p=300$, which has 99; the failure record and reporting protocol are described in Appendix~\ref{app:simulation-diagnostics}. 
Core code for the experiments is available at
\url{https://github.com/zhenf530/Distributio_Risk_ITR}.


\begin{table*}[!htbp]
\centering
\caption{Setting 1 (linear Gaussian; $n=1000$, $p=10$): full results for Mean, $\mathrm{Quantile}_{0.1}$, and CVaR. Entries are means (Monte Carlo standard errors) over 100 paired repetitions.}
\label{tab:s1_p10_full}
\fontsize{7.5pt}{9pt}\selectfont
\setlength{\tabcolsep}{1pt}
\begin{tabular*}{\textwidth}{@{\extracolsep{\fill}}llccccc@{}}
\toprule
Method & Metric & Mean & $\mathrm{Quantile}_{0.1}$ & $\mathrm{CVaR}_{0.1}$ & $\mathrm{CVaR}_{0.25}$ & $\mathrm{CVaR}_{0.5}$ \\
\midrule
\multirow{2}{*}{Oracle} & Value & \itrtablecell{1.810}{0.002} & \itrtablecell{\textminus0.042}{0.004} & \itrtablecell{\textminus0.699}{0.004} & \itrtablecell{\textminus0.015}{0.003} & \itrtablecell{0.672}{0.002} \\
 & Err & \itrtablecell{0.000}{0.000} & \itrtablecell{0.000}{0.000} & \itrtablecell{0.000}{0.000} & \itrtablecell{0.000}{0.000} & \itrtablecell{0.000}{0.000} \\
\multirow{2}{*}{$\ell_1$-PLS} & Value & \itrtablecell{1.805}{0.002} & \bestcell{\textminus0.050}{0.004} & \itrtablecell{\textminus0.712}{0.005} & \itrtablecell{\textminus0.024}{0.004} & \bestcell{0.666}{0.003} \\
 & Err & \itrtablecell{0.034}{0.002} & \itrtablecell{0.034}{0.002} & \bestcell{0.034}{0.002} & \itrtablecell{0.034}{0.002} & \itrtablecell{0.034}{0.002} \\
\multirow{2}{*}{RWL} & Value & \itrtablecell{1.800}{0.002} & \itrtablecell{\textminus0.060}{0.004} & \itrtablecell{\textminus0.720}{0.004} & \itrtablecell{\textminus0.032}{0.003} & \itrtablecell{0.658}{0.003} \\
 & Err & \itrtablecell{0.051}{0.002} & \itrtablecell{0.051}{0.002} & \itrtablecell{0.051}{0.002} & \itrtablecell{0.051}{0.002} & \itrtablecell{0.051}{0.002} \\
\addlinespace[3pt]
\multirow{2}{*}{CQR (0.1)} & Value & \itrtablecell{1.799}{0.002} & \itrtablecell{\textminus0.063}{0.004} & \itrtablecell{\textminus0.723}{0.004} & \itrtablecell{\textminus0.035}{0.003} & \itrtablecell{0.656}{0.003} \\
 & Err & \itrtablecell{0.053}{0.001} & \itrtablecell{0.053}{0.001} & \itrtablecell{0.053}{0.001} & \itrtablecell{0.053}{0.001} & \itrtablecell{0.053}{0.001} \\
\multirow{2}{*}{CQR (0.25)} & Value & \itrtablecell{1.803}{0.003} & \itrtablecell{\textminus0.057}{0.004} & \itrtablecell{\textminus0.717}{0.004} & \itrtablecell{\textminus0.029}{0.003} & \itrtablecell{0.661}{0.002} \\
 & Err & \itrtablecell{0.043}{0.001} & \itrtablecell{0.043}{0.001} & \itrtablecell{0.043}{0.001} & \itrtablecell{0.043}{0.001} & \itrtablecell{0.043}{0.001} \\
\multirow{2}{*}{CQR (0.5)} & Value & \itrtablecell{1.805}{0.002} & \itrtablecell{\textminus0.054}{0.004} & \itrtablecell{\textminus0.712}{0.004} & \itrtablecell{\textminus0.025}{0.003} & \itrtablecell{0.664}{0.002} \\
 & Err & \itrtablecell{0.039}{0.001} & \itrtablecell{0.039}{0.001} & \itrtablecell{0.039}{0.001} & \itrtablecell{0.039}{0.001} & \itrtablecell{0.039}{0.001} \\
\addlinespace[3pt]
\multirow{2}{*}{$\ell_2$-CVaR (0.1)} & Value & \itrtablecell{1.776}{0.003} & \itrtablecell{\textminus0.102}{0.005} & \itrtablecell{\textminus0.768}{0.005} & \itrtablecell{\textminus0.075}{0.004} & \itrtablecell{0.623}{0.004} \\
 & Err & \itrtablecell{0.092}{0.002} & \itrtablecell{0.092}{0.002} & \itrtablecell{0.092}{0.002} & \itrtablecell{0.092}{0.002} & \itrtablecell{0.092}{0.002} \\
\multirow{2}{*}{$\ell_2$-CVaR (0.25)} & Value & \itrtablecell{1.786}{0.003} & \itrtablecell{\textminus0.088}{0.004} & \itrtablecell{\textminus0.754}{0.005} & \itrtablecell{\textminus0.060}{0.004} & \itrtablecell{0.636}{0.003} \\
 & Err & \itrtablecell{0.078}{0.002} & \itrtablecell{0.078}{0.002} & \itrtablecell{0.078}{0.002} & \itrtablecell{0.078}{0.002} & \itrtablecell{0.078}{0.002} \\
\multirow{2}{*}{$\ell_2$-CVaR (0.5)} & Value & \itrtablecell{1.792}{0.003} & \itrtablecell{\textminus0.078}{0.004} & \itrtablecell{\textminus0.742}{0.005} & \itrtablecell{\textminus0.049}{0.004} & \itrtablecell{0.645}{0.003} \\
 & Err & \itrtablecell{0.069}{0.002} & \itrtablecell{0.069}{0.002} & \itrtablecell{0.069}{0.002} & \itrtablecell{0.069}{0.002} & \itrtablecell{0.069}{0.002} \\
\midrule
\multirow{2}{*}{Kernel} & Value & \itrtablecell{1.771}{0.003} & \itrtablecell{\textminus0.221}{0.006} & \itrtablecell{\textminus0.923}{0.007} & \itrtablecell{\textminus0.144}{0.006} & \itrtablecell{0.593}{0.004} \\
 & Err & \itrtablecell{0.096}{0.002} & \itrtablecell{0.148}{0.002} & \itrtablecell{0.146}{0.002} & \itrtablecell{0.125}{0.002} & \itrtablecell{0.111}{0.002} \\
\multirow{2}{*}{DRF} & Value & \itrtablecell{1.788}{0.002} & \itrtablecell{\textminus0.096}{0.004} & \itrtablecell{\textminus0.766}{0.005} & \itrtablecell{\textminus0.062}{0.003} & \itrtablecell{0.636}{0.003} \\
 & Err & \itrtablecell{0.075}{0.001} & \itrtablecell{0.084}{0.001} & \itrtablecell{0.085}{0.001} & \itrtablecell{0.080}{0.001} & \itrtablecell{0.077}{0.001} \\
\multirow{2}{*}{Conditional DDPM} & Value & \bestcell{1.807}{0.002} & \bestcell{\textminus0.050}{0.004} & \bestcell{\textminus0.709}{0.004} & \bestcell{\textminus0.023}{0.003} & \bestcell{0.666}{0.002} \\
 & Err & \bestcell{0.032}{0.001} & \bestcell{0.033}{0.001} & \bestcell{0.034}{0.001} & \bestcell{0.033}{0.001} & \bestcell{0.033}{0.001} \\
\multirow{2}{*}{GCDS} & Value & \itrtablecell{1.798}{0.003} & \itrtablecell{\textminus0.077}{0.004} & \itrtablecell{\textminus0.756}{0.005} & \itrtablecell{\textminus0.050}{0.003} & \itrtablecell{0.651}{0.003} \\
 & Err & \itrtablecell{0.055}{0.001} & \itrtablecell{0.067}{0.002} & \itrtablecell{0.076}{0.003} & \itrtablecell{0.066}{0.002} & \itrtablecell{0.058}{0.002} \\
\bottomrule
\end{tabular*}
\end{table*}

\begin{table*}[!htbp]
\centering
\caption{Setting~1 (linear Gaussian; $n=1000$, $p=20$): full results for Mean, $\mathrm{Quantile}_{0.1}$, and CVaR. Entries are means (Monte Carlo standard errors) over 100 paired repetitions.}
\label{tab:s1_p20_full}
\fontsize{7.5pt}{9pt}\selectfont
\setlength{\tabcolsep}{1pt}
\begin{tabular*}{\textwidth}{@{\extracolsep{\fill}}llccccc@{}}
\toprule
Method & Metric & Mean & $\mathrm{Quantile}_{0.1}$ & $\mathrm{CVaR}_{0.1}$ & $\mathrm{CVaR}_{0.25}$ & $\mathrm{CVaR}_{0.5}$ \\
\midrule
\multirow{2}{*}{Oracle} & Value & \itrtablecell{1.811}{0.003} & \itrtablecell{\textminus0.040}{0.003} & \itrtablecell{\textminus0.699}{0.004} & \itrtablecell{\textminus0.013}{0.003} & \itrtablecell{0.674}{0.002} \\
 & Err & \itrtablecell{0.000}{0.000} & \itrtablecell{0.000}{0.000} & \itrtablecell{0.000}{0.000} & \itrtablecell{0.000}{0.000} & \itrtablecell{0.000}{0.000} \\
\multirow{2}{*}{$\ell_1$-PLS} & Value & \bestcell{1.805}{0.003} & \bestcell{\textminus0.052}{0.004} & \bestcell{\textminus0.716}{0.004} & \bestcell{\textminus0.025}{0.003} & \bestcell{0.666}{0.003} \\
 & Err & \bestcell{0.035}{0.002} & \bestcell{0.035}{0.002} & \bestcell{0.035}{0.002} & \bestcell{0.035}{0.002} & \bestcell{0.035}{0.002} \\
\multirow{2}{*}{RWL} & Value & \itrtablecell{1.793}{0.003} & \itrtablecell{\textminus0.072}{0.004} & \itrtablecell{\textminus0.739}{0.004} & \itrtablecell{\textminus0.046}{0.004} & \itrtablecell{0.648}{0.003} \\
 & Err & \itrtablecell{0.065}{0.002} & \itrtablecell{0.065}{0.002} & \itrtablecell{0.065}{0.002} & \itrtablecell{0.065}{0.002} & \itrtablecell{0.065}{0.002} \\
\addlinespace[3pt]
\multirow{2}{*}{CQR (0.1)} & Value & \itrtablecell{1.788}{0.003} & \itrtablecell{\textminus0.083}{0.004} & \itrtablecell{\textminus0.751}{0.005} & \itrtablecell{\textminus0.055}{0.004} & \itrtablecell{0.641}{0.003} \\
 & Err & \itrtablecell{0.074}{0.001} & \itrtablecell{0.074}{0.001} & \itrtablecell{0.074}{0.001} & \itrtablecell{0.074}{0.001} & \itrtablecell{0.074}{0.001} \\
\multirow{2}{*}{CQR (0.25)} & Value & \itrtablecell{1.796}{0.003} & \itrtablecell{\textminus0.067}{0.004} & \itrtablecell{\textminus0.732}{0.004} & \itrtablecell{\textminus0.040}{0.003} & \itrtablecell{0.653}{0.003} \\
 & Err & \itrtablecell{0.058}{0.001} & \itrtablecell{0.058}{0.001} & \itrtablecell{0.058}{0.001} & \itrtablecell{0.058}{0.001} & \itrtablecell{0.058}{0.001} \\
\multirow{2}{*}{CQR (0.5)} & Value & \itrtablecell{1.797}{0.003} & \itrtablecell{\textminus0.064}{0.004} & \itrtablecell{\textminus0.729}{0.004} & \itrtablecell{\textminus0.038}{0.003} & \itrtablecell{0.655}{0.003} \\
 & Err & \itrtablecell{0.053}{0.001} & \itrtablecell{0.053}{0.001} & \itrtablecell{0.053}{0.001} & \itrtablecell{0.053}{0.001} & \itrtablecell{0.053}{0.001} \\
\addlinespace[3pt]
\multirow{2}{*}{$\ell_2$-CVaR (0.1)} & Value & \itrtablecell{1.750}{0.004} & \itrtablecell{\textminus0.148}{0.005} & \itrtablecell{\textminus0.828}{0.006} & \itrtablecell{\textminus0.121}{0.005} & \itrtablecell{0.587}{0.004} \\
 & Err & \itrtablecell{0.120}{0.002} & \itrtablecell{0.120}{0.002} & \itrtablecell{0.120}{0.002} & \itrtablecell{0.120}{0.002} & \itrtablecell{0.120}{0.002} \\
\multirow{2}{*}{$\ell_2$-CVaR (0.25)} & Value & \itrtablecell{1.771}{0.003} & \itrtablecell{\textminus0.110}{0.004} & \itrtablecell{\textminus0.790}{0.005} & \itrtablecell{\textminus0.086}{0.004} & \itrtablecell{0.617}{0.003} \\
 & Err & \itrtablecell{0.096}{0.002} & \itrtablecell{0.096}{0.002} & \itrtablecell{0.096}{0.002} & \itrtablecell{0.096}{0.002} & \itrtablecell{0.096}{0.002} \\
\multirow{2}{*}{$\ell_2$-CVaR (0.5)} & Value & \itrtablecell{1.781}{0.003} & \itrtablecell{\textminus0.092}{0.004} & \itrtablecell{\textminus0.764}{0.005} & \itrtablecell{\textminus0.066}{0.003} & \itrtablecell{0.632}{0.003} \\
 & Err & \itrtablecell{0.083}{0.002} & \itrtablecell{0.083}{0.002} & \itrtablecell{0.083}{0.002} & \itrtablecell{0.083}{0.002} & \itrtablecell{0.083}{0.002} \\
\midrule
\multirow{2}{*}{Kernel} & Value & \itrtablecell{1.735}{0.004} & \itrtablecell{\textminus0.342}{0.008} & \itrtablecell{\textminus1.070}{0.009} & \itrtablecell{\textminus0.251}{0.008} & \itrtablecell{0.527}{0.006} \\
 & Err & \itrtablecell{0.133}{0.003} & \itrtablecell{0.200}{0.003} & \itrtablecell{0.203}{0.003} & \itrtablecell{0.173}{0.003} & \itrtablecell{0.153}{0.003} \\
\multirow{2}{*}{DRF} & Value & \itrtablecell{1.781}{0.003} & \itrtablecell{\textminus0.103}{0.004} & \itrtablecell{\textminus0.780}{0.005} & \itrtablecell{\textminus0.069}{0.003} & \itrtablecell{0.630}{0.003} \\
 & Err & \itrtablecell{0.081}{0.001} & \itrtablecell{0.090}{0.002} & \itrtablecell{0.092}{0.002} & \itrtablecell{0.086}{0.001} & \itrtablecell{0.083}{0.001} \\
\multirow{2}{*}{Conditional DDPM} & Value & \bestcell{1.800}{0.003} & \bestcell{\textminus0.058}{0.004} & \bestcell{\textminus0.722}{0.004} & \bestcell{\textminus0.031}{0.003} & \bestcell{0.661}{0.003} \\
 & Err & \bestcell{0.048}{0.001} & \bestcell{0.048}{0.001} & \bestcell{0.049}{0.001} & \bestcell{0.048}{0.001} & \bestcell{0.048}{0.001} \\
\multirow{2}{*}{GCDS} & Value & \itrtablecell{1.779}{0.003} & \itrtablecell{\textminus0.099}{0.004} & \itrtablecell{\textminus0.782}{0.006} & \itrtablecell{\textminus0.075}{0.004} & \itrtablecell{0.628}{0.003} \\
 & Err & \itrtablecell{0.084}{0.002} & \itrtablecell{0.087}{0.002} & \itrtablecell{0.090}{0.002} & \itrtablecell{0.087}{0.002} & \itrtablecell{0.085}{0.002} \\
\bottomrule
\end{tabular*}
\end{table*}

\begin{table*}[!htbp]
\centering
\caption{Setting~1 (linear Gaussian; $n=1000$, $p=50$): full results for Mean, $\mathrm{Quantile}_{0.1}$, and CVaR. Entries are means (Monte Carlo standard errors) over 100 paired repetitions.}
\label{tab:s1_p50_full}
\fontsize{7.5pt}{9pt}\selectfont
\setlength{\tabcolsep}{1pt}
\begin{tabular*}{\textwidth}{@{\extracolsep{\fill}}llccccc@{}}
\toprule
Method & Metric & Mean & $\mathrm{Quantile}_{0.1}$ & $\mathrm{CVaR}_{0.1}$ & $\mathrm{CVaR}_{0.25}$ & $\mathrm{CVaR}_{0.5}$ \\
\midrule
\multirow{2}{*}{Oracle} & Value & \itrtablecell{1.814}{0.003} & \itrtablecell{\textminus0.045}{0.003} & \itrtablecell{\textminus0.702}{0.004} & \itrtablecell{\textminus0.019}{0.003} & \itrtablecell{0.671}{0.002} \\
 & Err & \itrtablecell{0.000}{0.000} & \itrtablecell{0.000}{0.000} & \itrtablecell{0.000}{0.000} & \itrtablecell{0.000}{0.000} & \itrtablecell{0.000}{0.000} \\
\multirow{2}{*}{$\ell_1$-PLS} & Value & \bestcell{1.806}{0.003} & \bestcell{\textminus0.062}{0.004} & \bestcell{\textminus0.727}{0.004} & \bestcell{\textminus0.036}{0.003} & \bestcell{0.658}{0.003} \\
 & Err & \bestcell{0.044}{0.002} & \bestcell{0.044}{0.002} & \bestcell{0.044}{0.002} & \bestcell{0.044}{0.002} & \bestcell{0.044}{0.002} \\
\multirow{2}{*}{RWL} & Value & \itrtablecell{1.777}{0.003} & \itrtablecell{\textminus0.112}{0.005} & \itrtablecell{\textminus0.787}{0.005} & \itrtablecell{\textminus0.088}{0.004} & \itrtablecell{0.616}{0.003} \\
 & Err & \itrtablecell{0.095}{0.002} & \itrtablecell{0.095}{0.002} & \itrtablecell{0.095}{0.002} & \itrtablecell{0.095}{0.002} & \itrtablecell{0.095}{0.002} \\
\addlinespace[3pt]
\multirow{2}{*}{CQR (0.1)} & Value & \itrtablecell{1.758}{0.003} & \itrtablecell{\textminus0.142}{0.004} & \itrtablecell{\textminus0.824}{0.005} & \itrtablecell{\textminus0.118}{0.004} & \itrtablecell{0.591}{0.003} \\
 & Err & \itrtablecell{0.115}{0.001} & \itrtablecell{0.115}{0.001} & \itrtablecell{0.115}{0.001} & \itrtablecell{0.115}{0.001} & \itrtablecell{0.115}{0.001} \\
\multirow{2}{*}{CQR (0.25)} & Value & \itrtablecell{1.779}{0.003} & \itrtablecell{\textminus0.109}{0.004} & \itrtablecell{\textminus0.780}{0.004} & \itrtablecell{\textminus0.083}{0.003} & \itrtablecell{0.619}{0.003} \\
 & Err & \itrtablecell{0.093}{0.001} & \itrtablecell{0.093}{0.001} & \itrtablecell{0.093}{0.001} & \itrtablecell{0.093}{0.001} & \itrtablecell{0.093}{0.001} \\
\multirow{2}{*}{CQR (0.5)} & Value & \itrtablecell{1.784}{0.003} & \itrtablecell{\textminus0.101}{0.004} & \itrtablecell{\textminus0.773}{0.004} & \itrtablecell{\textminus0.075}{0.003} & \itrtablecell{0.626}{0.003} \\
 & Err & \itrtablecell{0.086}{0.001} & \itrtablecell{0.086}{0.001} & \itrtablecell{0.086}{0.001} & \itrtablecell{0.086}{0.001} & \itrtablecell{0.086}{0.001} \\
\addlinespace[3pt]
\multirow{2}{*}{$\ell_2$-CVaR (0.1)} & Value & \itrtablecell{1.695}{0.004} & \itrtablecell{\textminus0.254}{0.006} & \itrtablecell{\textminus0.950}{0.007} & \itrtablecell{\textminus0.226}{0.006} & \itrtablecell{0.501}{0.005} \\
 & Err & \itrtablecell{0.172}{0.002} & \itrtablecell{0.172}{0.002} & \itrtablecell{0.172}{0.002} & \itrtablecell{0.172}{0.002} & \itrtablecell{0.172}{0.002} \\
\multirow{2}{*}{$\ell_2$-CVaR (0.25)} & Value & \itrtablecell{1.732}{0.004} & \itrtablecell{\textminus0.183}{0.006} & \itrtablecell{\textminus0.877}{0.006} & \itrtablecell{\textminus0.161}{0.005} & \itrtablecell{0.557}{0.004} \\
 & Err & \itrtablecell{0.141}{0.002} & \itrtablecell{0.141}{0.002} & \itrtablecell{0.141}{0.002} & \itrtablecell{0.141}{0.002} & \itrtablecell{0.141}{0.002} \\
\multirow{2}{*}{$\ell_2$-CVaR (0.5)} & Value & \itrtablecell{1.754}{0.003} & \itrtablecell{\textminus0.144}{0.005} & \itrtablecell{\textminus0.832}{0.006} & \itrtablecell{\textminus0.123}{0.004} & \itrtablecell{0.588}{0.004} \\
 & Err & \itrtablecell{0.119}{0.002} & \itrtablecell{0.119}{0.002} & \itrtablecell{0.119}{0.002} & \itrtablecell{0.119}{0.002} & \itrtablecell{0.119}{0.002} \\
\midrule
\multirow{2}{*}{Kernel} & Value & \itrtablecell{1.664}{0.004} & \itrtablecell{\textminus0.522}{0.007} & \itrtablecell{\textminus1.284}{0.009} & \itrtablecell{\textminus0.411}{0.007} & \itrtablecell{0.411}{0.005} \\
 & Err & \itrtablecell{0.191}{0.002} & \itrtablecell{0.269}{0.002} & \itrtablecell{0.281}{0.003} & \itrtablecell{0.240}{0.002} & \itrtablecell{0.213}{0.002} \\
\multirow{2}{*}{DRF} & Value & \bestcell{1.777}{0.003} & \bestcell{\textminus0.123}{0.005} & \bestcell{\textminus0.811}{0.007} & \bestcell{\textminus0.089}{0.004} & \bestcell{0.615}{0.003} \\
 & Err & \bestcell{0.094}{0.001} & \bestcell{0.103}{0.002} & \bestcell{0.105}{0.002} & \bestcell{0.098}{0.001} & \bestcell{0.095}{0.001} \\
\multirow{2}{*}{Conditional DDPM} & Value & \itrtablecell{1.726}{0.003} & \itrtablecell{\textminus0.197}{0.005} & \itrtablecell{\textminus0.892}{0.006} & \itrtablecell{\textminus0.174}{0.005} & \itrtablecell{0.546}{0.004} \\
 & Err & \itrtablecell{0.147}{0.002} & \itrtablecell{0.148}{0.002} & \itrtablecell{0.149}{0.002} & \itrtablecell{0.148}{0.002} & \itrtablecell{0.147}{0.002} \\
\multirow{2}{*}{GCDS} & Value & \itrtablecell{1.626}{0.006} & \itrtablecell{\textminus0.388}{0.012} & \itrtablecell{\textminus1.121}{0.014} & \itrtablecell{\textminus0.358}{0.011} & \itrtablecell{0.403}{0.009} \\
 & Err & \itrtablecell{0.213}{0.004} & \itrtablecell{0.223}{0.004} & \itrtablecell{0.230}{0.005} & \itrtablecell{0.222}{0.004} & \itrtablecell{0.217}{0.004} \\
\bottomrule
\end{tabular*}
\end{table*}

\begin{table*}[!htbp]
\centering
\caption{Setting~1 (linear Gaussian; $n=1000$, $p=100$): full results for Mean, $\mathrm{Quantile}_{0.1}$, and CVaR. Entries are means (Monte Carlo standard errors) over 100 paired repetitions.}
\label{tab:s1_p100_full}
\fontsize{7.5pt}{9pt}\selectfont
\setlength{\tabcolsep}{1pt}
\begin{tabular*}{\textwidth}{@{\extracolsep{\fill}}llccccc@{}}
\toprule
Method & Metric & Mean & $\mathrm{Quantile}_{0.1}$ & $\mathrm{CVaR}_{0.1}$ & $\mathrm{CVaR}_{0.25}$ & $\mathrm{CVaR}_{0.5}$ \\
\midrule
\multirow{2}{*}{Oracle} & Value & \itrtablecell{1.817}{0.003} & \itrtablecell{\textminus0.046}{0.003} & \itrtablecell{\textminus0.700}{0.003} & \itrtablecell{\textminus0.016}{0.003} & \itrtablecell{0.673}{0.003} \\
 & Err & \itrtablecell{0.000}{0.000} & \itrtablecell{0.000}{0.000} & \itrtablecell{0.000}{0.000} & \itrtablecell{0.000}{0.000} & \itrtablecell{0.000}{0.000} \\
\multirow{2}{*}{$\ell_1$-PLS} & Value & \bestcell{1.807}{0.003} & \bestcell{\textminus0.063}{0.004} & \bestcell{\textminus0.723}{0.005} & \bestcell{\textminus0.033}{0.003} & \bestcell{0.660}{0.003} \\
 & Err & \bestcell{0.046}{0.002} & \bestcell{0.046}{0.002} & \bestcell{0.046}{0.002} & \bestcell{0.046}{0.002} & \bestcell{0.046}{0.002} \\
\multirow{2}{*}{RWL} & Value & \itrtablecell{1.744}{0.004} & \itrtablecell{\textminus0.170}{0.005} & \itrtablecell{\textminus0.852}{0.006} & \itrtablecell{\textminus0.141}{0.005} & \itrtablecell{0.573}{0.004} \\
 & Err & \itrtablecell{0.130}{0.002} & \itrtablecell{0.130}{0.002} & \itrtablecell{0.130}{0.002} & \itrtablecell{0.130}{0.002} & \itrtablecell{0.130}{0.002} \\
\addlinespace[3pt]
\multirow{2}{*}{CQR (0.1)} & Value & \itrtablecell{1.703}{0.003} & \itrtablecell{\textminus0.240}{0.005} & \itrtablecell{\textminus0.936}{0.006} & \itrtablecell{\textminus0.212}{0.005} & \itrtablecell{0.514}{0.004} \\
 & Err & \itrtablecell{0.166}{0.001} & \itrtablecell{0.166}{0.001} & \itrtablecell{0.166}{0.001} & \itrtablecell{0.166}{0.001} & \itrtablecell{0.166}{0.001} \\
\multirow{2}{*}{CQR (0.25)} & Value & \itrtablecell{1.737}{0.003} & \itrtablecell{\textminus0.180}{0.005} & \itrtablecell{\textminus0.863}{0.005} & \itrtablecell{\textminus0.151}{0.004} & \itrtablecell{0.564}{0.003} \\
 & Err & \itrtablecell{0.136}{0.001} & \itrtablecell{0.136}{0.001} & \itrtablecell{0.136}{0.001} & \itrtablecell{0.136}{0.001} & \itrtablecell{0.136}{0.001} \\
\multirow{2}{*}{CQR (0.5)} & Value & \itrtablecell{1.750}{0.003} & \itrtablecell{\textminus0.162}{0.004} & \itrtablecell{\textminus0.840}{0.004} & \itrtablecell{\textminus0.132}{0.003} & \itrtablecell{0.580}{0.003} \\
 & Err & \itrtablecell{0.125}{0.001} & \itrtablecell{0.125}{0.001} & \itrtablecell{0.125}{0.001} & \itrtablecell{0.125}{0.001} & \itrtablecell{0.125}{0.001} \\
\addlinespace[3pt]
\multirow{2}{*}{$\ell_2$-CVaR (0.1)} & Value & \itrtablecell{1.632}{0.004} & \itrtablecell{\textminus0.357}{0.006} & \itrtablecell{\textminus1.062}{0.007} & \itrtablecell{\textminus0.325}{0.006} & \itrtablecell{0.416}{0.005} \\
 & Err & \itrtablecell{0.215}{0.002} & \itrtablecell{0.215}{0.002} & \itrtablecell{0.215}{0.002} & \itrtablecell{0.215}{0.002} & \itrtablecell{0.215}{0.002} \\
\multirow{2}{*}{$\ell_2$-CVaR (0.25)} & Value & \itrtablecell{1.679}{0.004} & \itrtablecell{\textminus0.281}{0.006} & \itrtablecell{\textminus0.978}{0.006} & \itrtablecell{\textminus0.250}{0.005} & \itrtablecell{0.482}{0.005} \\
 & Err & \itrtablecell{0.183}{0.002} & \itrtablecell{0.183}{0.002} & \itrtablecell{0.183}{0.002} & \itrtablecell{0.183}{0.002} & \itrtablecell{0.183}{0.002} \\
\multirow{2}{*}{$\ell_2$-CVaR (0.5)} & Value & \itrtablecell{1.707}{0.004} & \itrtablecell{\textminus0.232}{0.006} & \itrtablecell{\textminus0.927}{0.006} & \itrtablecell{\textminus0.205}{0.005} & \itrtablecell{0.520}{0.004} \\
 & Err & \itrtablecell{0.162}{0.002} & \itrtablecell{0.162}{0.002} & \itrtablecell{0.162}{0.002} & \itrtablecell{0.162}{0.002} & \itrtablecell{0.162}{0.002} \\
\midrule
\multirow{2}{*}{Kernel} & Value & \itrtablecell{1.578}{0.005} & \itrtablecell{\textminus0.676}{0.008} & \itrtablecell{\textminus1.424}{0.009} & \itrtablecell{\textminus0.556}{0.007} & \itrtablecell{0.291}{0.006} \\
 & Err & \itrtablecell{0.243}{0.002} & \itrtablecell{0.323}{0.002} & \itrtablecell{0.334}{0.003} & \itrtablecell{0.293}{0.002} & \itrtablecell{0.265}{0.002} \\
\multirow{2}{*}{DRF} & Value & \bestcell{1.770}{0.003} & \bestcell{\textminus0.152}{0.007} & \bestcell{\textminus0.850}{0.012} & \bestcell{\textminus0.110}{0.006} & \bestcell{0.604}{0.003} \\
 & Err & \bestcell{0.104}{0.002} & \bestcell{0.117}{0.003} & \bestcell{0.119}{0.003} & \bestcell{0.110}{0.002} & \bestcell{0.108}{0.002} \\
\multirow{2}{*}{Conditional DDPM} & Value & \itrtablecell{1.342}{0.014} & \itrtablecell{\textminus0.756}{0.016} & \itrtablecell{\textminus1.471}{0.016} & \itrtablecell{\textminus0.720}{0.016} & \itrtablecell{0.052}{0.015} \\
 & Err & \itrtablecell{0.359}{0.006} & \itrtablecell{0.360}{0.006} & \itrtablecell{0.361}{0.006} & \itrtablecell{0.360}{0.006} & \itrtablecell{0.360}{0.006} \\
\multirow{2}{*}{GCDS} & Value & \itrtablecell{1.689}{0.004} & \itrtablecell{\textminus0.264}{0.005} & \itrtablecell{\textminus0.957}{0.006} & \itrtablecell{\textminus0.233}{0.005} & \itrtablecell{0.496}{0.004} \\
 & Err & \itrtablecell{0.176}{0.001} & \itrtablecell{0.176}{0.001} & \itrtablecell{0.176}{0.001} & \itrtablecell{0.176}{0.001} & \itrtablecell{0.176}{0.001} \\
\bottomrule
\end{tabular*}
\end{table*}

\begin{table*}[!htbp]
\centering
\caption{Setting~1 (linear Gaussian; $n=1000$, $p=300$): full results for Mean, $\mathrm{Quantile}_{0.1}$, and CVaR. Entries are means (Monte Carlo standard errors) over 100 paired repetitions.}
\label{tab:s1_p300_full}
\fontsize{7.5pt}{9pt}\selectfont
\setlength{\tabcolsep}{1pt}
\begin{tabular*}{\textwidth}{@{\extracolsep{\fill}}llccccc@{}}
\toprule
Method & Metric & Mean & $\mathrm{Quantile}_{0.1}$ & $\mathrm{CVaR}_{0.1}$ & $\mathrm{CVaR}_{0.25}$ & $\mathrm{CVaR}_{0.5}$ \\
\midrule
\multirow{2}{*}{Oracle} & Value & \itrtablecell{1.816}{0.002} & \itrtablecell{\textminus0.044}{0.003} & \itrtablecell{\textminus0.703}{0.004} & \itrtablecell{\textminus0.016}{0.003} & \itrtablecell{0.674}{0.003} \\
 & Err & \itrtablecell{0.000}{0.000} & \itrtablecell{0.000}{0.000} & \itrtablecell{0.000}{0.000} & \itrtablecell{0.000}{0.000} & \itrtablecell{0.000}{0.000} \\
\multirow{2}{*}{$\ell_1$-PLS} & Value & \bestcell{1.799}{0.003} & \bestcell{\textminus0.075}{0.004} & \bestcell{\textminus0.750}{0.005} & \bestcell{\textminus0.049}{0.004} & \bestcell{0.648}{0.003} \\
 & Err & \bestcell{0.058}{0.002} & \bestcell{0.058}{0.002} & \bestcell{0.058}{0.002} & \bestcell{0.058}{0.002} & \bestcell{0.058}{0.002} \\
\multirow{2}{*}{RWL} & Value & \itrtablecell{1.600}{0.004} & \itrtablecell{\textminus0.404}{0.006} & \itrtablecell{\textminus1.115}{0.006} & \itrtablecell{\textminus0.372}{0.006} & \itrtablecell{0.375}{0.005} \\
 & Err & \itrtablecell{0.233}{0.002} & \itrtablecell{0.233}{0.002} & \itrtablecell{0.233}{0.002} & \itrtablecell{0.233}{0.002} & \itrtablecell{0.233}{0.002} \\
\addlinespace[3pt]
\multirow{2}{*}{CQR (0.1)} & Value & \itrtablecell{1.518}{0.004} & \itrtablecell{\textminus0.524}{0.006} & \itrtablecell{\textminus1.246}{0.006} & \itrtablecell{\textminus0.494}{0.005} & \itrtablecell{0.268}{0.005} \\
 & Err & \itrtablecell{0.276}{0.001} & \itrtablecell{0.276}{0.001} & \itrtablecell{0.276}{0.001} & \itrtablecell{0.276}{0.001} & \itrtablecell{0.276}{0.001} \\
\multirow{2}{*}{CQR (0.25)} & Value & \itrtablecell{1.533}{0.004} & \itrtablecell{\textminus0.501}{0.005} & \itrtablecell{\textminus1.223}{0.005} & \itrtablecell{\textminus0.471}{0.005} & \itrtablecell{0.288}{0.004} \\
 & Err & \itrtablecell{0.268}{0.002} & \itrtablecell{0.268}{0.002} & \itrtablecell{0.268}{0.002} & \itrtablecell{0.268}{0.002} & \itrtablecell{0.268}{0.002} \\
\multirow{2}{*}{CQR (0.5)} & Value & \itrtablecell{1.560}{0.004} & \itrtablecell{\textminus0.460}{0.005} & \itrtablecell{\textminus1.177}{0.006} & \itrtablecell{\textminus0.430}{0.005} & \itrtablecell{0.323}{0.004} \\
 & Err & \itrtablecell{0.254}{0.001} & \itrtablecell{0.254}{0.001} & \itrtablecell{0.254}{0.001} & \itrtablecell{0.254}{0.001} & \itrtablecell{0.254}{0.001} \\
\addlinespace[3pt]
\multirow{2}{*}{$\ell_2$-CVaR (0.1)} & Value & \itrtablecell{1.490}{0.004} & \itrtablecell{\textminus0.563}{0.006} & \itrtablecell{\textminus1.284}{0.007} & \itrtablecell{\textminus0.531}{0.006} & \itrtablecell{0.233}{0.005} \\
 & Err & \itrtablecell{0.290}{0.002} & \itrtablecell{0.290}{0.002} & \itrtablecell{0.290}{0.002} & \itrtablecell{0.290}{0.002} & \itrtablecell{0.290}{0.002} \\
\multirow{2}{*}{$\ell_2$-CVaR (0.25)} & Value & \itrtablecell{1.542}{0.004} & \itrtablecell{\textminus0.490}{0.006} & \itrtablecell{\textminus1.208}{0.006} & \itrtablecell{\textminus0.459}{0.006} & \itrtablecell{0.299}{0.005} \\
 & Err & \itrtablecell{0.264}{0.002} & \itrtablecell{0.264}{0.002} & \itrtablecell{0.264}{0.002} & \itrtablecell{0.264}{0.002} & \itrtablecell{0.264}{0.002} \\
\multirow{2}{*}{$\ell_2$-CVaR (0.5)} & Value & \itrtablecell{1.592}{0.004} & \itrtablecell{\textminus0.417}{0.006} & \itrtablecell{\textminus1.129}{0.006} & \itrtablecell{\textminus0.387}{0.005} & \itrtablecell{0.363}{0.005} \\
 & Err & \itrtablecell{0.238}{0.002} & \itrtablecell{0.238}{0.002} & \itrtablecell{0.238}{0.002} & \itrtablecell{0.238}{0.002} & \itrtablecell{0.238}{0.002} \\
\midrule
\multirow{2}{*}{Kernel} & Value & \itrtablecell{1.408}{0.005} & \itrtablecell{\textminus0.865}{0.007} & \itrtablecell{\textminus1.595}{0.007} & \itrtablecell{\textminus0.773}{0.007} & \itrtablecell{0.063}{0.007} \\
 & Err & \itrtablecell{0.330}{0.002} & \itrtablecell{0.396}{0.002} & \itrtablecell{0.402}{0.002} & \itrtablecell{0.375}{0.002} & \itrtablecell{0.352}{0.002} \\
\multirow{2}{*}{DRF} & Value & \bestcell{1.751}{0.004} & \bestcell{\textminus0.240}{0.012} & \bestcell{\textminus1.004}{0.019} & \bestcell{\textminus0.193}{0.010} & \bestcell{0.557}{0.005} \\
 & Err & \bestcell{0.128}{0.002} & \bestcell{0.161}{0.005} & \bestcell{0.176}{0.006} & \bestcell{0.149}{0.004} & \bestcell{0.138}{0.003} \\
\multirow{2}{*}{Conditional DDPM} & Value & \itrtablecell{1.055}{0.006} & \itrtablecell{\textminus1.079}{0.007} & \itrtablecell{\textminus1.774}{0.006} & \itrtablecell{\textminus1.039}{0.006} & \itrtablecell{\textminus0.271}{0.007} \\
 & Err & \itrtablecell{0.479}{0.002} & \itrtablecell{0.478}{0.002} & \itrtablecell{0.478}{0.002} & \itrtablecell{0.478}{0.002} & \itrtablecell{0.478}{0.002} \\
\multirow{2}{*}{GCDS} & Value & \itrtablecell{1.229}{0.006} & \itrtablecell{\textminus0.900}{0.007} & \itrtablecell{\textminus1.607}{0.007} & \itrtablecell{\textminus0.859}{0.007} & \itrtablecell{\textminus0.083}{0.007} \\
 & Err & \itrtablecell{0.407}{0.002} & \itrtablecell{0.407}{0.002} & \itrtablecell{0.407}{0.002} & \itrtablecell{0.407}{0.002} & \itrtablecell{0.407}{0.002} \\
\bottomrule
\end{tabular*}
\end{table*}

\begin{table*}[!htbp]
\centering
\caption{Setting 2 (nonlinear Gaussian; $n=1000$, $p=10$): full results for Mean, $\mathrm{Quantile}_{0.1}$, and CVaR. Entries are means (Monte Carlo standard errors) over 100 paired repetitions.}
\label{tab:s2_p10_full}
\fontsize{7.5pt}{9pt}\selectfont
\setlength{\tabcolsep}{1pt}
\begin{tabular*}{\textwidth}{@{\extracolsep{\fill}}llccccc@{}}
\toprule
Method & Metric & Mean & $\mathrm{Quantile}_{0.1}$ & $\mathrm{CVaR}_{0.1}$ & $\mathrm{CVaR}_{0.25}$ & $\mathrm{CVaR}_{0.5}$ \\
\midrule
\multirow{2}{*}{Oracle} & Value & \itrtablecell{1.654}{0.003} & \itrtablecell{\textminus0.044}{0.003} & \itrtablecell{\textminus0.587}{0.003} & \itrtablecell{\textminus0.029}{0.002} & \itrtablecell{0.533}{0.002} \\
 & Err & \itrtablecell{0.000}{0.000} & \itrtablecell{0.000}{0.000} & \itrtablecell{0.000}{0.000} & \itrtablecell{0.000}{0.000} & \itrtablecell{0.000}{0.000} \\
\multirow{2}{*}{$\ell_1$-PLS} & Value & \itrtablecell{1.566}{0.003} & \itrtablecell{\textminus0.156}{0.003} & \itrtablecell{\textminus0.697}{0.003} & \itrtablecell{\textminus0.139}{0.003} & \itrtablecell{0.425}{0.002} \\
 & Err & \itrtablecell{0.173}{0.001} & \itrtablecell{0.173}{0.001} & \itrtablecell{0.173}{0.001} & \itrtablecell{0.173}{0.001} & \itrtablecell{0.173}{0.001} \\
\multirow{2}{*}{RWL} & Value & \itrtablecell{1.592}{0.003} & \itrtablecell{\textminus0.118}{0.003} & \itrtablecell{\textminus0.658}{0.003} & \itrtablecell{\textminus0.102}{0.003} & \itrtablecell{0.460}{0.002} \\
 & Err & \itrtablecell{0.214}{0.002} & \itrtablecell{0.214}{0.002} & \itrtablecell{0.214}{0.002} & \itrtablecell{0.214}{0.002} & \itrtablecell{0.214}{0.002} \\
\addlinespace[3pt]
\multirow{2}{*}{CQR (0.1)} & Value & \itrtablecell{1.561}{0.003} & \itrtablecell{\textminus0.166}{0.004} & \itrtablecell{\textminus0.709}{0.004} & \itrtablecell{\textminus0.148}{0.003} & \itrtablecell{0.418}{0.003} \\
 & Err & \itrtablecell{0.182}{0.001} & \itrtablecell{0.182}{0.001} & \itrtablecell{0.182}{0.001} & \itrtablecell{0.182}{0.001} & \itrtablecell{0.182}{0.001} \\
\multirow{2}{*}{CQR (0.25)} & Value & \itrtablecell{1.568}{0.003} & \itrtablecell{\textminus0.155}{0.003} & \itrtablecell{\textminus0.696}{0.004} & \itrtablecell{\textminus0.137}{0.003} & \itrtablecell{0.428}{0.002} \\
 & Err & \itrtablecell{0.180}{0.001} & \itrtablecell{0.180}{0.001} & \itrtablecell{0.180}{0.001} & \itrtablecell{0.180}{0.001} & \itrtablecell{0.180}{0.001} \\
\multirow{2}{*}{CQR (0.5)} & Value & \itrtablecell{1.568}{0.003} & \itrtablecell{\textminus0.155}{0.003} & \itrtablecell{\textminus0.696}{0.003} & \itrtablecell{\textminus0.137}{0.003} & \itrtablecell{0.427}{0.002} \\
 & Err & \itrtablecell{0.178}{0.001} & \itrtablecell{0.178}{0.001} & \itrtablecell{0.178}{0.001} & \itrtablecell{0.178}{0.001} & \itrtablecell{0.178}{0.001} \\
\addlinespace[3pt]
\multirow{2}{*}{$\ell_2$-CVaR (0.1)} & Value & \itrtablecell{1.559}{0.004} & \itrtablecell{\textminus0.140}{0.004} & \itrtablecell{\textminus0.679}{0.004} & \itrtablecell{\textminus0.124}{0.003} & \itrtablecell{0.436}{0.003} \\
 & Err & \itrtablecell{0.253}{0.003} & \itrtablecell{0.253}{0.003} & \itrtablecell{0.253}{0.003} & \itrtablecell{0.253}{0.003} & \itrtablecell{0.253}{0.003} \\
\multirow{2}{*}{$\ell_2$-CVaR (0.25)} & Value & \itrtablecell{1.578}{0.003} & \itrtablecell{\textminus0.125}{0.004} & \itrtablecell{\textminus0.664}{0.004} & \itrtablecell{\textminus0.109}{0.003} & \itrtablecell{0.450}{0.003} \\
 & Err & \itrtablecell{0.242}{0.003} & \itrtablecell{0.242}{0.003} & \itrtablecell{0.242}{0.003} & \itrtablecell{0.242}{0.003} & \itrtablecell{0.242}{0.003} \\
\multirow{2}{*}{$\ell_2$-CVaR (0.5)} & Value & \itrtablecell{1.584}{0.003} & \itrtablecell{\textminus0.120}{0.003} & \itrtablecell{\textminus0.659}{0.003} & \itrtablecell{\textminus0.105}{0.003} & \itrtablecell{0.455}{0.002} \\
 & Err & \itrtablecell{0.237}{0.002} & \itrtablecell{0.237}{0.002} & \itrtablecell{0.237}{0.002} & \itrtablecell{0.237}{0.002} & \itrtablecell{0.237}{0.002} \\
\midrule
\multirow{2}{*}{Kernel} & Value & \itrtablecell{1.582}{0.003} & \itrtablecell{\textminus0.237}{0.004} & \itrtablecell{\textminus0.807}{0.004} & \itrtablecell{\textminus0.219}{0.004} & \itrtablecell{0.381}{0.004} \\
 & Err & \itrtablecell{0.231}{0.002} & \itrtablecell{0.412}{0.003} & \itrtablecell{0.448}{0.003} & \itrtablecell{0.416}{0.003} & \itrtablecell{0.366}{0.003} \\
\multirow{2}{*}{DRF} & Value & \bestcell{1.638}{0.003} & \bestcell{\textminus0.076}{0.003} & \bestcell{\textminus0.626}{0.004} & \bestcell{\textminus0.060}{0.003} & \bestcell{0.508}{0.003} \\
 & Err & \bestcell{0.066}{0.004} & \bestcell{0.129}{0.007} & \bestcell{0.158}{0.008} & \bestcell{0.124}{0.007} & \bestcell{0.098}{0.006} \\
\multirow{2}{*}{Conditional DDPM} & Value & \itrtablecell{1.596}{0.003} & \itrtablecell{\textminus0.121}{0.003} & \itrtablecell{\textminus0.663}{0.003} & \itrtablecell{\textminus0.105}{0.002} & \itrtablecell{0.460}{0.002} \\
 & Err & \itrtablecell{0.200}{0.001} & \itrtablecell{0.195}{0.001} & \itrtablecell{0.193}{0.001} & \itrtablecell{0.195}{0.001} & \itrtablecell{0.197}{0.001} \\
\multirow{2}{*}{GCDS} & Value & \itrtablecell{1.568}{0.003} & \itrtablecell{\textminus0.192}{0.007} & \itrtablecell{\textminus0.769}{0.011} & \itrtablecell{\textminus0.173}{0.006} & \itrtablecell{0.413}{0.004} \\
 & Err & \itrtablecell{0.184}{0.001} & \itrtablecell{0.187}{0.001} & \itrtablecell{0.195}{0.002} & \itrtablecell{0.187}{0.001} & \itrtablecell{0.183}{0.001} \\
\bottomrule
\end{tabular*}
\end{table*}

\begin{table*}[!htbp]
\centering
\caption{Setting~2 (nonlinear Gaussian; $n=1000$, $p=20$): full results for Mean, $\mathrm{Quantile}_{0.1}$, and CVaR. Entries are means (Monte Carlo standard errors) over 100 paired repetitions.}
\label{tab:s2_p20_full}
\fontsize{7.5pt}{9pt}\selectfont
\setlength{\tabcolsep}{1pt}
\begin{tabular*}{\textwidth}{@{\extracolsep{\fill}}llccccc@{}}
\toprule
Method & Metric & Mean & $\mathrm{Quantile}_{0.1}$ & $\mathrm{CVaR}_{0.1}$ & $\mathrm{CVaR}_{0.25}$ & $\mathrm{CVaR}_{0.5}$ \\
\midrule
\multirow{2}{*}{Oracle} & Value & \itrtablecell{1.650}{0.003} & \itrtablecell{\textminus0.045}{0.003} & \itrtablecell{\textminus0.590}{0.003} & \itrtablecell{\textminus0.029}{0.002} & \itrtablecell{0.531}{0.002} \\
 & Err & \itrtablecell{0.000}{0.000} & \itrtablecell{0.000}{0.000} & \itrtablecell{0.000}{0.000} & \itrtablecell{0.000}{0.000} & \itrtablecell{0.000}{0.000} \\
\multirow{2}{*}{$\ell_1$-PLS} & Value & \itrtablecell{1.558}{0.003} & \itrtablecell{\textminus0.163}{0.003} & \itrtablecell{\textminus0.708}{0.003} & \itrtablecell{\textminus0.145}{0.003} & \itrtablecell{0.418}{0.002} \\
 & Err & \itrtablecell{0.174}{0.001} & \itrtablecell{0.174}{0.001} & \itrtablecell{0.174}{0.001} & \itrtablecell{0.174}{0.001} & \itrtablecell{0.174}{0.001} \\
\multirow{2}{*}{RWL} & Value & \itrtablecell{1.580}{0.003} & \itrtablecell{\textminus0.134}{0.004} & \itrtablecell{\textminus0.674}{0.004} & \itrtablecell{\textminus0.115}{0.003} & \itrtablecell{0.446}{0.003} \\
 & Err & \itrtablecell{0.221}{0.002} & \itrtablecell{0.221}{0.002} & \itrtablecell{0.221}{0.002} & \itrtablecell{0.221}{0.002} & \itrtablecell{0.221}{0.002} \\
\addlinespace[3pt]
\multirow{2}{*}{CQR (0.1)} & Value & \itrtablecell{1.545}{0.003} & \itrtablecell{\textminus0.188}{0.004} & \itrtablecell{\textminus0.737}{0.004} & \itrtablecell{\textminus0.169}{0.003} & \itrtablecell{0.399}{0.003} \\
 & Err & \itrtablecell{0.193}{0.001} & \itrtablecell{0.193}{0.001} & \itrtablecell{0.193}{0.001} & \itrtablecell{0.193}{0.001} & \itrtablecell{0.193}{0.001} \\
\multirow{2}{*}{CQR (0.25)} & Value & \itrtablecell{1.555}{0.003} & \itrtablecell{\textminus0.171}{0.003} & \itrtablecell{\textminus0.718}{0.003} & \itrtablecell{\textminus0.153}{0.003} & \itrtablecell{0.412}{0.002} \\
 & Err & \itrtablecell{0.185}{0.001} & \itrtablecell{0.185}{0.001} & \itrtablecell{0.185}{0.001} & \itrtablecell{0.185}{0.001} & \itrtablecell{0.185}{0.001} \\
\multirow{2}{*}{CQR (0.5)} & Value & \itrtablecell{1.557}{0.003} & \itrtablecell{\textminus0.168}{0.003} & \itrtablecell{\textminus0.713}{0.004} & \itrtablecell{\textminus0.150}{0.003} & \itrtablecell{0.415}{0.002} \\
 & Err & \itrtablecell{0.184}{0.001} & \itrtablecell{0.184}{0.001} & \itrtablecell{0.184}{0.001} & \itrtablecell{0.184}{0.001} & \itrtablecell{0.184}{0.001} \\
\addlinespace[3pt]
\multirow{2}{*}{$\ell_2$-CVaR (0.1)} & Value & \itrtablecell{1.515}{0.005} & \itrtablecell{\textminus0.185}{0.005} & \itrtablecell{\textminus0.729}{0.006} & \itrtablecell{\textminus0.167}{0.005} & \itrtablecell{0.394}{0.004} \\
 & Err & \itrtablecell{0.269}{0.003} & \itrtablecell{0.269}{0.003} & \itrtablecell{0.269}{0.003} & \itrtablecell{0.269}{0.003} & \itrtablecell{0.269}{0.003} \\
\multirow{2}{*}{$\ell_2$-CVaR (0.25)} & Value & \itrtablecell{1.550}{0.003} & \itrtablecell{\textminus0.154}{0.004} & \itrtablecell{\textminus0.695}{0.004} & \itrtablecell{\textminus0.136}{0.003} & \itrtablecell{0.423}{0.003} \\
 & Err & \itrtablecell{0.255}{0.002} & \itrtablecell{0.255}{0.002} & \itrtablecell{0.255}{0.002} & \itrtablecell{0.255}{0.002} & \itrtablecell{0.255}{0.002} \\
\multirow{2}{*}{$\ell_2$-CVaR (0.5)} & Value & \itrtablecell{1.564}{0.003} & \itrtablecell{\textminus0.143}{0.003} & \itrtablecell{\textminus0.684}{0.004} & \itrtablecell{\textminus0.126}{0.003} & \itrtablecell{0.434}{0.003} \\
 & Err & \itrtablecell{0.247}{0.002} & \itrtablecell{0.247}{0.002} & \itrtablecell{0.247}{0.002} & \itrtablecell{0.247}{0.002} & \itrtablecell{0.247}{0.002} \\
\midrule
\multirow{2}{*}{Kernel} & Value & \itrtablecell{1.542}{0.003} & \itrtablecell{\textminus0.305}{0.003} & \itrtablecell{\textminus0.848}{0.003} & \itrtablecell{\textminus0.294}{0.003} & \itrtablecell{0.287}{0.003} \\
 & Err & \itrtablecell{0.262}{0.003} & \itrtablecell{0.469}{0.002} & \itrtablecell{0.487}{0.001} & \itrtablecell{0.474}{0.001} & \itrtablecell{0.436}{0.002} \\
\multirow{2}{*}{DRF} & Value & \bestcell{1.633}{0.003} & \bestcell{\textminus0.083}{0.004} & \bestcell{\textminus0.632}{0.004} & \bestcell{\textminus0.064}{0.003} & \bestcell{0.503}{0.003} \\
 & Err & \bestcell{0.069}{0.005} & \bestcell{0.130}{0.008} & \bestcell{0.165}{0.008} & \bestcell{0.125}{0.008} & \bestcell{0.096}{0.006} \\
\multirow{2}{*}{Conditional DDPM} & Value & \itrtablecell{1.575}{0.003} & \itrtablecell{\textminus0.148}{0.003} & \itrtablecell{\textminus0.697}{0.004} & \itrtablecell{\textminus0.130}{0.003} & \itrtablecell{0.436}{0.002} \\
 & Err & \itrtablecell{0.191}{0.001} & \itrtablecell{0.187}{0.001} & \itrtablecell{0.185}{0.001} & \itrtablecell{0.187}{0.001} & \itrtablecell{0.188}{0.001} \\
\multirow{2}{*}{GCDS} & Value & \itrtablecell{1.542}{0.003} & \itrtablecell{\textminus0.162}{0.004} & \itrtablecell{\textminus0.712}{0.006} & \itrtablecell{\textminus0.143}{0.004} & \itrtablecell{0.420}{0.003} \\
 & Err & \itrtablecell{0.197}{0.001} & \itrtablecell{0.251}{0.005} & \itrtablecell{0.276}{0.007} & \itrtablecell{0.249}{0.005} & \itrtablecell{0.223}{0.003} \\
\bottomrule
\end{tabular*}
\end{table*}

\begin{table*}[!htbp]
\centering
\caption{Setting~2 (nonlinear Gaussian; $n=1000$, $p=300$): full results for Mean, $\mathrm{Quantile}_{0.1}$, and CVaR. Entries are means (Monte Carlo standard errors) over 100 paired repetitions, except GCDS, which has 99 evaluable repetitions.}
\label{tab:s2_p300_full}
\fontsize{7.5pt}{9pt}\selectfont
\setlength{\tabcolsep}{1pt}
\begin{tabular*}{\textwidth}{@{\extracolsep{\fill}}llccccc@{}}
\toprule
Method & Metric & Mean & $\mathrm{Quantile}_{0.1}$ & $\mathrm{CVaR}_{0.1}$ & $\mathrm{CVaR}_{0.25}$ & $\mathrm{CVaR}_{0.5}$ \\
\midrule
\multirow{2}{*}{Oracle} & Value & \itrtablecell{1.650}{0.003} & \itrtablecell{\textminus0.049}{0.003} & \itrtablecell{\textminus0.590}{0.003} & \itrtablecell{\textminus0.031}{0.002} & \itrtablecell{0.530}{0.002} \\
 & Err & \itrtablecell{0.000}{0.000} & \itrtablecell{0.000}{0.000} & \itrtablecell{0.000}{0.000} & \itrtablecell{0.000}{0.000} & \itrtablecell{0.000}{0.000} \\
\multirow{2}{*}{$\ell_1$-PLS} & Value & \itrtablecell{1.551}{0.003} & \bestcell{\textminus0.175}{0.003} & \bestcell{\textminus0.717}{0.003} & \bestcell{\textminus0.157}{0.003} & \bestcell{0.408}{0.002} \\
 & Err & \bestcell{0.174}{0.001} & \bestcell{0.174}{0.001} & \bestcell{0.174}{0.001} & \bestcell{0.174}{0.001} & \bestcell{0.174}{0.001} \\
\multirow{2}{*}{RWL} & Value & \itrtablecell{1.385}{0.005} & \itrtablecell{\textminus0.358}{0.007} & \itrtablecell{\textminus0.963}{0.009} & \itrtablecell{\textminus0.350}{0.007} & \itrtablecell{0.245}{0.006} \\
 & Err & \itrtablecell{0.318}{0.002} & \itrtablecell{0.318}{0.002} & \itrtablecell{0.318}{0.002} & \itrtablecell{0.318}{0.002} & \itrtablecell{0.318}{0.002} \\
\addlinespace[3pt]
\multirow{2}{*}{CQR (0.1)} & Value & \itrtablecell{1.281}{0.003} & \itrtablecell{\textminus0.564}{0.004} & \itrtablecell{\textminus1.235}{0.005} & \itrtablecell{\textminus0.558}{0.004} & \itrtablecell{0.088}{0.003} \\
 & Err & \itrtablecell{0.355}{0.001} & \itrtablecell{0.355}{0.001} & \itrtablecell{0.355}{0.001} & \itrtablecell{0.355}{0.001} & \itrtablecell{0.355}{0.001} \\
\multirow{2}{*}{CQR (0.25)} & Value & \itrtablecell{1.290}{0.003} & \itrtablecell{\textminus0.552}{0.004} & \itrtablecell{\textminus1.222}{0.005} & \itrtablecell{\textminus0.547}{0.004} & \itrtablecell{0.096}{0.003} \\
 & Err & \itrtablecell{0.351}{0.001} & \itrtablecell{0.351}{0.001} & \itrtablecell{0.351}{0.001} & \itrtablecell{0.351}{0.001} & \itrtablecell{0.351}{0.001} \\
\multirow{2}{*}{CQR (0.5)} & Value & \itrtablecell{1.313}{0.004} & \itrtablecell{\textminus0.520}{0.005} & \itrtablecell{\textminus1.182}{0.006} & \itrtablecell{\textminus0.515}{0.004} & \itrtablecell{0.123}{0.003} \\
 & Err & \itrtablecell{0.340}{0.002} & \itrtablecell{0.340}{0.002} & \itrtablecell{0.340}{0.002} & \itrtablecell{0.340}{0.002} & \itrtablecell{0.340}{0.002} \\
\addlinespace[3pt]
\multirow{2}{*}{$\ell_2$-CVaR (0.1)} & Value & \itrtablecell{1.245}{0.005} & \itrtablecell{\textminus0.529}{0.007} & \itrtablecell{\textminus1.185}{0.010} & \itrtablecell{\textminus0.525}{0.007} & \itrtablecell{0.103}{0.005} \\
 & Err & \itrtablecell{0.388}{0.002} & \itrtablecell{0.388}{0.002} & \itrtablecell{0.388}{0.002} & \itrtablecell{0.388}{0.002} & \itrtablecell{0.388}{0.002} \\
\multirow{2}{*}{$\ell_2$-CVaR (0.25)} & Value & \itrtablecell{1.277}{0.004} & \itrtablecell{\textminus0.482}{0.006} & \itrtablecell{\textminus1.128}{0.007} & \itrtablecell{\textminus0.480}{0.005} & \itrtablecell{0.139}{0.004} \\
 & Err & \itrtablecell{0.372}{0.002} & \itrtablecell{0.372}{0.002} & \itrtablecell{0.372}{0.002} & \itrtablecell{0.372}{0.002} & \itrtablecell{0.372}{0.002} \\
\multirow{2}{*}{$\ell_2$-CVaR (0.5)} & Value & \itrtablecell{1.305}{0.004} & \itrtablecell{\textminus0.421}{0.008} & \itrtablecell{\textminus1.040}{0.012} & \itrtablecell{\textminus0.415}{0.008} & \itrtablecell{0.187}{0.006} \\
 & Err & \itrtablecell{0.364}{0.002} & \itrtablecell{0.364}{0.002} & \itrtablecell{0.364}{0.002} & \itrtablecell{0.364}{0.002} & \itrtablecell{0.364}{0.002} \\
\midrule
\multirow{2}{*}{Kernel} & Value & \itrtablecell{1.225}{0.004} & \itrtablecell{\textminus0.336}{0.003} & \itrtablecell{\textminus0.860}{0.003} & \itrtablecell{\textminus0.322}{0.002} & \itrtablecell{0.207}{0.002} \\
 & Err & \itrtablecell{0.416}{0.003} & \itrtablecell{0.497}{0.001} & \itrtablecell{0.498}{0.001} & \itrtablecell{0.498}{0.001} & \itrtablecell{0.496}{0.001} \\
\multirow{2}{*}{DRF} & Value & \bestcell{1.604}{0.003} & \bestcell{\textminus0.331}{0.003} & \bestcell{\textminus0.858}{0.003} & \bestcell{\textminus0.320}{0.002} & \bestcell{0.267}{0.007} \\
 & Err & \bestcell{0.186}{0.009} & \itrtablecell{0.498}{0.001} & \itrtablecell{0.499}{0.001} & \itrtablecell{0.498}{0.001} & \itrtablecell{0.465}{0.003} \\
\multirow{2}{*}{Conditional DDPM} & Value & \itrtablecell{1.095}{0.004} & \itrtablecell{\textminus0.411}{0.022} & \itrtablecell{\textminus0.965}{0.026} & \itrtablecell{\textminus0.399}{0.021} & \itrtablecell{0.154}{0.016} \\
 & Err & \itrtablecell{0.491}{0.002} & \itrtablecell{0.490}{0.002} & \itrtablecell{0.490}{0.002} & \itrtablecell{0.490}{0.002} & \itrtablecell{0.490}{0.002} \\
\multirow{2}{*}{GCDS$^{\dagger}$} & Value & \itrtablecell{1.331}{0.005} & \itrtablecell{\textminus0.337}{0.008} & \itrtablecell{\textminus0.920}{0.012} & \itrtablecell{\textminus0.330}{0.008} & \itrtablecell{0.246}{0.005} \\
 & Err & \itrtablecell{0.361}{0.004} & \bestcell{0.362}{0.004} & \bestcell{0.362}{0.004} & \bestcell{0.362}{0.004} & \bestcell{0.362}{0.004} \\
\bottomrule
\end{tabular*}
\end{table*}

\begin{table*}[!htbp]
\centering
\caption{Setting 3 (heavy-tailed lognormal; $n=1000$, $p=10$): full results for Mean, $\mathrm{Quantile}_{0.1}$, and CVaR. Entries are means (Monte Carlo standard errors) over 100 paired repetitions.}
\label{tab:s3_p10_full}
\fontsize{7.5pt}{9pt}\selectfont
\setlength{\tabcolsep}{1pt}
\begin{tabular*}{\textwidth}{@{\extracolsep{\fill}}llccccc@{}}
\toprule
Method & Metric & Mean & $\mathrm{Quantile}_{0.1}$ & $\mathrm{CVaR}_{0.1}$ & $\mathrm{CVaR}_{0.25}$ & $\mathrm{CVaR}_{0.5}$ \\
\midrule
\multirow{2}{*}{Oracle} & Value & \itrtablecell{9.244}{0.075} & \itrtablecell{1.167}{0.003} & \itrtablecell{0.615}{0.003} & \itrtablecell{1.243}{0.002} & \itrtablecell{1.952}{0.002} \\
 & Err & \itrtablecell{0.000}{0.000} & \itrtablecell{0.000}{0.000} & \itrtablecell{0.000}{0.000} & \itrtablecell{0.000}{0.000} & \itrtablecell{0.000}{0.000} \\
\multirow{2}{*}{$\ell_1$-PLS} & Value & \itrtablecell{8.670}{0.081} & \itrtablecell{0.267}{0.028} & \itrtablecell{\textminus0.339}{0.025} & \itrtablecell{0.357}{0.027} & \itrtablecell{1.193}{0.027} \\
 & Err & \itrtablecell{0.432}{0.011} & \itrtablecell{0.432}{0.011} & \itrtablecell{0.432}{0.011} & \itrtablecell{0.432}{0.011} & \itrtablecell{0.432}{0.011} \\
\multirow{2}{*}{RWL} & Value & \itrtablecell{8.759}{0.089} & \itrtablecell{0.348}{0.024} & \itrtablecell{\textminus0.263}{0.023} & \itrtablecell{0.444}{0.024} & \itrtablecell{1.288}{0.023} \\
 & Err & \itrtablecell{0.396}{0.009} & \itrtablecell{0.396}{0.009} & \itrtablecell{0.396}{0.009} & \itrtablecell{0.396}{0.009} & \itrtablecell{0.396}{0.009} \\
\addlinespace[3pt]
\multirow{2}{*}{CQR (0.1)} & Value & \itrtablecell{9.243}{0.076} & \bestcell{1.167}{0.003} & \bestcell{0.614}{0.003} & \bestcell{1.242}{0.002} & \bestcell{1.951}{0.002} \\
 & Err & \bestcell{0.009}{0.000} & \bestcell{0.009}{0.000} & \bestcell{0.009}{0.000} & \bestcell{0.009}{0.000} & \bestcell{0.009}{0.000} \\
\multirow{2}{*}{CQR (0.25)} & Value & \bestcell{9.252}{0.074} & \itrtablecell{1.164}{0.003} & \itrtablecell{0.610}{0.003} & \itrtablecell{1.240}{0.002} & \itrtablecell{1.949}{0.002} \\
 & Err & \itrtablecell{0.021}{0.000} & \itrtablecell{0.021}{0.000} & \itrtablecell{0.021}{0.000} & \itrtablecell{0.021}{0.000} & \itrtablecell{0.021}{0.000} \\
\multirow{2}{*}{CQR (0.5)} & Value & \itrtablecell{9.200}{0.075} & \itrtablecell{1.131}{0.004} & \itrtablecell{0.559}{0.005} & \itrtablecell{1.203}{0.003} & \itrtablecell{1.922}{0.003} \\
 & Err & \itrtablecell{0.073}{0.002} & \itrtablecell{0.073}{0.002} & \itrtablecell{0.073}{0.002} & \itrtablecell{0.073}{0.002} & \itrtablecell{0.073}{0.002} \\
\addlinespace[3pt]
\multirow{2}{*}{$\ell_2$-CVaR (0.1)} & Value & \itrtablecell{9.074}{0.084} & \itrtablecell{0.867}{0.040} & \itrtablecell{0.279}{0.041} & \itrtablecell{0.943}{0.039} & \itrtablecell{1.701}{0.034} \\
 & Err & \itrtablecell{0.190}{0.016} & \itrtablecell{0.190}{0.016} & \itrtablecell{0.190}{0.016} & \itrtablecell{0.190}{0.016} & \itrtablecell{0.190}{0.016} \\
\multirow{2}{*}{$\ell_2$-CVaR (0.25)} & Value & \itrtablecell{9.080}{0.086} & \itrtablecell{0.884}{0.036} & \itrtablecell{0.289}{0.037} & \itrtablecell{0.959}{0.035} & \itrtablecell{1.721}{0.030} \\
 & Err & \itrtablecell{0.186}{0.014} & \itrtablecell{0.186}{0.014} & \itrtablecell{0.186}{0.014} & \itrtablecell{0.186}{0.014} & \itrtablecell{0.186}{0.014} \\
\multirow{2}{*}{$\ell_2$-CVaR (0.5)} & Value & \itrtablecell{9.098}{0.090} & \itrtablecell{0.882}{0.038} & \itrtablecell{0.287}{0.039} & \itrtablecell{0.959}{0.037} & \itrtablecell{1.719}{0.033} \\
 & Err & \itrtablecell{0.187}{0.015} & \itrtablecell{0.187}{0.015} & \itrtablecell{0.187}{0.015} & \itrtablecell{0.187}{0.015} & \itrtablecell{0.187}{0.015} \\
\midrule
\multirow{2}{*}{Kernel} & Value & \itrtablecell{8.681}{0.082} & \itrtablecell{1.069}{0.004} & \itrtablecell{0.468}{0.004} & \bestcell{1.168}{0.003} & \bestcell{1.896}{0.003} \\
 & Err & \itrtablecell{0.416}{0.007} & \bestcell{0.116}{0.001} & \bestcell{0.109}{0.001} & \bestcell{0.096}{0.001} & \bestcell{0.098}{0.001} \\
\multirow{2}{*}{DRF} & Value & \itrtablecell{8.861}{0.081} & \bestcell{1.088}{0.004} & \bestcell{0.516}{0.005} & \itrtablecell{1.166}{0.003} & \itrtablecell{1.886}{0.003} \\
 & Err & \itrtablecell{0.372}{0.010} & \itrtablecell{0.118}{0.002} & \itrtablecell{0.118}{0.002} & \itrtablecell{0.114}{0.002} & \itrtablecell{0.114}{0.002} \\
\multirow{2}{*}{Conditional DDPM} & Value & \itrtablecell{8.994}{0.081} & \itrtablecell{0.897}{0.018} & \itrtablecell{0.267}{0.021} & \itrtablecell{0.976}{0.017} & \itrtablecell{1.752}{0.013} \\
 & Err & \itrtablecell{0.266}{0.009} & \itrtablecell{0.194}{0.007} & \itrtablecell{0.201}{0.008} & \itrtablecell{0.193}{0.007} & \itrtablecell{0.192}{0.006} \\
\multirow{2}{*}{GCDS} & Value & \bestcell{9.089}{0.074} & \itrtablecell{1.024}{0.014} & \itrtablecell{0.411}{0.016} & \itrtablecell{1.097}{0.013} & \itrtablecell{1.847}{0.010} \\
 & Err & \bestcell{0.163}{0.008} & \itrtablecell{0.138}{0.006} & \itrtablecell{0.142}{0.006} & \itrtablecell{0.136}{0.006} & \itrtablecell{0.133}{0.006} \\
\bottomrule
\end{tabular*}
\end{table*}

\subsection{Discussion about GCDS}
\label{app:gcds}
We further consider two additional scenarios that are more favorable to GCDS \citep{zhou2023deep}. Since the goal here is internal method selection rather than external benchmarking, we mainly compare these models with DRF, which was the strongest distribution-first estimator in the earlier experiments.
Setting 4 uses multiplicative non-Gaussian noise:
$
Y=\bigl(1+X_1+X_2+A(X_1-X_2+X_3)\bigr)\,Z,\ Z=\exp(0.5\varepsilon),
$
where \(\varepsilon\) follows a two-component Gaussian mixture. Setting 5 uses a mix Gaussian distribution:
$
Y\mid X,A \sim 0.5\,N(m(X)+A\delta(X)-2,0.5^2)
+0.5\,N(m(X)+A\delta(X)+2,0.5^2).
$

Tables~\ref{tab:s4_p10_drf_gcds}--\ref{tab:s5_p20_drf_gcds}
show that the gains remain local. Under Setting~4, GCDS is only marginally better than DRF at \(p=10\), and this advantage disappears already at \(p=20\). Under Setting~5, GCDS has a clear advantage at \(p=10\), but this advantage again disappears at \(p=20\). Hence, even in scenarios designed to favor generative estimators, the advantage is limited and unstable.
Combined with the earlier benchmark results, this leads us to a cautious view of generative estimators in the current tabular ITR problem: they are feasible within the distribution-first framework and may provide gains in selected scenarios, but their behavior is not stable enough to justify using them as default choices. Given their substantially higher training and tuning cost, we currently regard generative models as promising but still exploratory extensions.

\begin{table}[!htbp]
\centering
\caption{Setting~4 (multiplicative non-Gaussian noise; $n=1000$, $p=10$): full results for Mean, $\mathrm{Quantile}_{0.1}$, and CVaR. Entries are means (Monte Carlo standard errors) over 100 paired repetitions.}
\label{tab:s4_p10_drf_gcds}
\label{tab:gcds_targeted}
\fontsize{7.5pt}{9pt}\selectfont
\setlength{\tabcolsep}{1pt}
\begin{tabular*}{\textwidth}{@{\extracolsep{\fill}}llccccc@{}}
\toprule
Method & Metric & Mean & $\mathrm{Quantile}_{0.1}$ & $\mathrm{CVaR}_{0.1}$ & $\mathrm{CVaR}_{0.25}$ & $\mathrm{CVaR}_{0.5}$ \\
\midrule
\multirow{2}{*}{Oracle} & Value & \itrtablecell{3.167}{0.006} & \itrtablecell{0.200}{0.001} & \itrtablecell{\textminus0.115}{0.003} & \itrtablecell{0.179}{0.001} & \itrtablecell{0.530}{0.001} \\
 & Err & \itrtablecell{0.000}{0.000} & \itrtablecell{0.000}{0.000} & \itrtablecell{0.000}{0.000} & \itrtablecell{0.000}{0.000} & \itrtablecell{0.000}{0.000} \\
\multirow{2}{*}{DRF} & Value & \itrtablecell{3.099}{0.007} & \bestcell{0.183}{0.001} & \bestcell{\textminus0.171}{0.005} & \bestcell{0.152}{0.002} & \bestcell{0.505}{0.001} \\
 & Err & \itrtablecell{0.091}{0.002} & \bestcell{0.083}{0.002} & \bestcell{0.080}{0.001} & \bestcell{0.080}{0.001} & \bestcell{0.087}{0.002} \\
\multirow{2}{*}{GCDS} & Value & \bestcell{3.130}{0.007} & \itrtablecell{0.168}{0.002} & \itrtablecell{\textminus0.257}{0.013} & \itrtablecell{0.116}{0.004} & \itrtablecell{0.493}{0.002} \\
 & Err & \bestcell{0.072}{0.002} & \itrtablecell{0.129}{0.003} & \itrtablecell{0.145}{0.004} & \itrtablecell{0.122}{0.003} & \itrtablecell{0.097}{0.002} \\
\bottomrule
\end{tabular*}
\end{table}

\begin{table}[!htbp]
\centering
\caption{Setting~4 (multiplicative non-Gaussian noise; $n=1000$, $p=20$): full results for Mean, $\mathrm{Quantile}_{0.1}$, and CVaR. Entries are means (Monte Carlo standard errors) over 100 paired repetitions.}
\label{tab:s4_p20_drf_gcds}
\fontsize{7.5pt}{9pt}\selectfont
\setlength{\tabcolsep}{1pt}
\begin{tabular*}{\textwidth}{@{\extracolsep{\fill}}llccccc@{}}
\toprule
Method & Metric & Mean & $\mathrm{Quantile}_{0.1}$ & $\mathrm{CVaR}_{0.1}$ & $\mathrm{CVaR}_{0.25}$ & $\mathrm{CVaR}_{0.5}$ \\
\midrule
\multirow{2}{*}{Oracle} & Value & \itrtablecell{3.166}{0.007} & \itrtablecell{0.199}{0.001} & \itrtablecell{\textminus0.117}{0.003} & \itrtablecell{0.178}{0.001} & \itrtablecell{0.530}{0.002} \\
 & Err & \itrtablecell{0.000}{0.000} & \itrtablecell{0.000}{0.000} & \itrtablecell{0.000}{0.000} & \itrtablecell{0.000}{0.000} & \itrtablecell{0.000}{0.000} \\
\multirow{2}{*}{DRF} & Value & \bestcell{3.074}{0.008} & \bestcell{0.176}{0.001} & \bestcell{\textminus0.183}{0.004} & \bestcell{0.145}{0.002} & \bestcell{0.497}{0.002} \\
 & Err & \bestcell{0.106}{0.002} & \bestcell{0.097}{0.002} & \bestcell{0.097}{0.002} & \bestcell{0.094}{0.002} & \bestcell{0.100}{0.002} \\
\multirow{2}{*}{GCDS} & Value & \itrtablecell{3.059}{0.008} & \itrtablecell{0.152}{0.003} & \itrtablecell{\textminus0.314}{0.010} & \itrtablecell{0.079}{0.005} & \itrtablecell{0.461}{0.004} \\
 & Err & \itrtablecell{0.116}{0.003} & \itrtablecell{0.113}{0.003} & \itrtablecell{0.114}{0.003} & \itrtablecell{0.112}{0.003} & \itrtablecell{0.111}{0.003} \\
\bottomrule
\end{tabular*}
\end{table}

\begin{table}[!htbp]
\centering
\caption{Setting~5 (additive Gaussian-mixture noise; $n=1000$, $p=10$): full results for Mean, $\mathrm{Quantile}_{0.1}$, and CVaR. Entries are means (Monte Carlo standard errors) over 100 paired repetitions.}
\label{tab:s5_p10_drf_gcds}
\fontsize{7.5pt}{9pt}\selectfont
\setlength{\tabcolsep}{1pt}
\begin{tabular*}{\textwidth}{@{\extracolsep{\fill}}llccccc@{}}
\toprule
Method & Metric & Mean & $\mathrm{Quantile}_{0.1}$ & $\mathrm{CVaR}_{0.1}$ & $\mathrm{CVaR}_{0.25}$ & $\mathrm{CVaR}_{0.5}$ \\
\midrule
\multirow{2}{*}{Oracle} & Value & \itrtablecell{1.808}{0.003} & \itrtablecell{\textminus1.193}{0.004} & \itrtablecell{\textminus1.817}{0.003} & \itrtablecell{\textminus1.106}{0.003} & \itrtablecell{\textminus0.213}{0.003} \\
 & Err & \itrtablecell{0.000}{0.000} & \itrtablecell{0.000}{0.000} & \itrtablecell{0.000}{0.000} & \itrtablecell{0.000}{0.000} & \itrtablecell{0.000}{0.000} \\
\multirow{2}{*}{DRF} & Value & \itrtablecell{1.750}{0.004} & \itrtablecell{\textminus1.267}{0.004} & \itrtablecell{\textminus1.927}{0.007} & \itrtablecell{\textminus1.179}{0.004} & \itrtablecell{\textminus0.284}{0.004} \\
 & Err & \itrtablecell{0.120}{0.003} & \itrtablecell{0.113}{0.002} & \itrtablecell{0.112}{0.002} & \itrtablecell{0.110}{0.002} & \itrtablecell{0.122}{0.003} \\
\multirow{2}{*}{GCDS} & Value & \bestcell{1.799}{0.004} & \bestcell{\textminus1.203}{0.004} & \bestcell{\textminus1.832}{0.004} & \bestcell{\textminus1.115}{0.003} & \bestcell{\textminus0.226}{0.004} \\
 & Err & \bestcell{0.049}{0.002} & \bestcell{0.040}{0.001} & \bestcell{0.042}{0.002} & \bestcell{0.040}{0.001} & \bestcell{0.054}{0.002} \\
\bottomrule
\end{tabular*}
\end{table}

\begin{table}[!htbp]
\centering
\caption{Setting~5 (additive Gaussian-mixture noise; $n=1000$, $p=20$): full results for Mean, $\mathrm{Quantile}_{0.1}$, and CVaR. Entries are means (Monte Carlo standard errors) over 100 paired repetitions.}
\label{tab:s5_p20_drf_gcds}
\fontsize{7.5pt}{9pt}\selectfont
\setlength{\tabcolsep}{1pt}
\begin{tabular*}{\textwidth}{@{\extracolsep{\fill}}llccccc@{}}
\toprule
Method & Metric & Mean & $\mathrm{Quantile}_{0.1}$ & $\mathrm{CVaR}_{0.1}$ & $\mathrm{CVaR}_{0.25}$ & $\mathrm{CVaR}_{0.5}$ \\
\midrule
\multirow{2}{*}{Oracle} & Value & \itrtablecell{1.813}{0.004} & \itrtablecell{\textminus1.195}{0.003} & \itrtablecell{\textminus1.816}{0.003} & \itrtablecell{\textminus1.105}{0.003} & \itrtablecell{\textminus0.210}{0.003} \\
 & Err & \itrtablecell{0.000}{0.000} & \itrtablecell{0.000}{0.000} & \itrtablecell{0.000}{0.000} & \itrtablecell{0.000}{0.000} & \itrtablecell{0.000}{0.000} \\
\multirow{2}{*}{DRF} & Value & \bestcell{1.731}{0.005} & \bestcell{\textminus1.293}{0.004} & \bestcell{\textminus1.964}{0.008} & \bestcell{\textminus1.210}{0.004} & \bestcell{\textminus0.306}{0.005} \\
 & Err & \bestcell{0.144}{0.003} & \bestcell{0.138}{0.002} & \bestcell{0.138}{0.003} & \bestcell{0.136}{0.003} & \bestcell{0.147}{0.003} \\
\multirow{2}{*}{GCDS} & Value & \itrtablecell{1.635}{0.010} & \itrtablecell{\textminus1.513}{0.018} & \itrtablecell{\textminus2.247}{0.020} & \itrtablecell{\textminus1.412}{0.016} & \itrtablecell{\textminus0.417}{0.011} \\
 & Err & \itrtablecell{0.207}{0.005} & \itrtablecell{0.226}{0.006} & \itrtablecell{0.237}{0.007} & \itrtablecell{0.224}{0.006} & \itrtablecell{0.214}{0.006} \\
\bottomrule
\end{tabular*}
\end{table}

\clearpage
\subsection{Treatment-Interactive Outcome Distributions}
\label{app:treatment-interactive}

To examine criterion-dependent treatment decisions, we adopt the treatment-interactive error design from Simulation Study II of \href{https://doi.org/10.1002/sim.8102}{Xiao, Zhang, and Lu (2019)}. Let \(X_1,X_2\) be independent \(\mathrm{Uniform}[-1,1]\) variables, let \(B\sim\mathrm{Bernoulli}(0.5)\) independently of \(X\), and generate
\[
\begin{aligned}
Y={}&1+0.5\sin\{\pi(X_1-X_2)\}+0.25(1+X_1+2X_2)^2\\
&+(B-0.5)\theta^\top\widetilde X+(1+B d_0X_1^2)\epsilon,
\end{aligned}
\]
where \(\widetilde X=(1,X_1,X_2)^\top\), \(\theta=(0.5,2,-1)^\top\), \(d_0=5\), and \(\epsilon\sim\mathrm{Gamma}(\mathrm{shape}=1,\mathrm{rate}=1)-1\) independently of \((X,B)\). The implementation uses \(A\in\{-1,+1\}\) with \(B=(A+1)/2\). We use \(p=2\), \(n=800\), 100 Monte Carlo replications, and an independent test sample of size 10,000 per replication. Treatment changes both the conditional location and scale. 

Table~\ref{tab:treatment_interactive} reports the criterion-matched comparisons for Mean, \(Q_{0.1}\), and \(\cvar_{0.1}\).
Conditional regret is the corresponding oracle value minus the learned-rule value. For the mean criterion, \(\ell_1\)-PLS performs best among the reported methods, consistent with the linear mean treatment contrast. For \(Q_{0.1}\) and \(\cvar_{0.1}\), DRF performs best and Kernel shows similar tail-criterion gains. These results illustrate criterion-adaptive decision making with flexible distribution estimation in this nonlinear-tail setting.

\begin{table}[htbp]
\centering
\caption{Treatment-interactive simulation with \(n=800\) and \(p=2\). Entries are Monte Carlo means (standard errors) over 100 replications.}
\label{tab:treatment_interactive}
\fontsize{9}{11}\selectfont
\setlength{\tabcolsep}{5pt}
\renewcommand{\arraystretch}{1.1}
\begin{tabular*}{\linewidth}{@{\extracolsep{\fill}}llccc@{}}
\toprule
Criterion & Method & Value & Conditional regret & OverallErr \\
\midrule
\multirow{4}{*}{Mean} & \(\ell_1\)-PLS & \bestcell{2.229}{0.001} & \bestcell{0.011}{0.001} & \bestcell{0.056}{0.003} \\
& RWL-linear & \itrtablecell{2.219}{0.004} & \itrtablecell{0.021}{0.004} & \itrtablecell{0.074}{0.005} \\
& Kernel & \itrtablecell{2.212}{0.002} & \itrtablecell{0.028}{0.002} & \itrtablecell{0.078}{0.003} \\
& DRF & \itrtablecell{2.202}{0.003} & \itrtablecell{0.038}{0.003} & \itrtablecell{0.104}{0.004} \\
\midrule
\multirow{3}{*}{\(Q_{0.1}\)} & CQR (0.1) & \itrtablecell{0.455}{0.005} & \itrtablecell{0.313}{0.005} & \itrtablecell{0.396}{0.004} \\
& Kernel & \itrtablecell{0.729}{0.002} & \itrtablecell{0.039}{0.002} & \itrtablecell{0.115}{0.002} \\
& DRF & \bestcell{0.730}{0.003} & \bestcell{0.038}{0.003} & \bestcell{0.099}{0.003} \\
\midrule
\multirow{3}{*}{\(\cvar_{0.1}\)} & Qi-CVaR (0.1) & \itrtablecell{0.461}{0.002} & \itrtablecell{0.242}{0.002} & \itrtablecell{0.296}{0.002} \\
& Kernel & \itrtablecell{0.658}{0.002} & \itrtablecell{0.044}{0.002} & \itrtablecell{0.134}{0.002} \\
& DRF & \bestcell{0.666}{0.003} & \bestcell{0.036}{0.002} & \bestcell{0.117}{0.003} \\
\bottomrule
\end{tabular*}
\end{table}

\clearpage
\subsection{Tuning and Preflight Diagnostics}
\label{app:simulation-diagnostics}

\paragraph{Non-generative methods.}
The following protocol applies to Settings~1--3 and their
higher-dimensional extensions. Kernel selects a bandwidth shared by
both treatment arms by ten-fold cross-validation over
\(\{0.15,0.20,0.25,0.30,0.40,0.50,0.70,0.90,1.20,1.50,2.00\}\),
minimizing the mean squared CDF prediction error at empirical outcome
quantiles for 20 equally spaced probability levels in \([0.1,0.9]\),
using all covariates. \(\ell_1\)-PLS selects its penalty along the
\texttt{glmnet} path by ten-fold cross-validation, maximizing the
summed validation-fold policy value. RWL uses a linear kernel and
five-fold policy-value cross-validation over \(\{2^{-10},\ldots,2^{0}\}\).
\(\ell_2\)-CVaR uses the same grid with five-fold cross-validation,
CQR initialization within each training fold, and its original
validation score, which under \(\mu=0\) is the mean outcome among
validation observations whose treatment agrees with the learned rule;
it is then refit on the full training sample. CQR fits linear quantile
regression with treatment interactions at \(\tau\in\{0.1,0.25,0.5\}\)
without tuning. DRF uses 500 trees per treatment arm and otherwise the
defaults of \texttt{drf} 1.3.1.

\paragraph{Generative methods.}
Conditional DDPM and GCDS are tuned on independent pilot data. We
measure predictive accuracy by the continuous ranked probability score
(CRPS; lower is better) on factual validation outcomes, averaged within
and then equally across treatment arms, and define
\(\mathrm{skill}=1-\mathrm{CRPS}_{\mathrm{model}}/\mathrm{CRPS}_{\mathrm{ref}}\),
where the reference is the covariate-free empirical outcome distribution
of the corresponding training arm. Each candidate is fit on five pilot
datasets with two training seeds each, using treatment-stratified 80/20
training--validation splits and 512 generated draws per validation
covariate and arm. A candidate is eligible if at least 9 of its 10 runs
succeed and its scores are finite on all five datasets; Conditional
DDPM additionally requires mean skill of at least \(-0.10\).
Conditional DDPM searches learning rates \(\{10^{-4},2\times10^{-4}\}\)
and hidden widths \(\{64,128\}\), selecting by a one-standard-error rule
that favors smaller networks, with standard errors computed across the
five dataset-level averages. GCDS searches generator and discriminator
learning rates in \(\{5\times10^{-4},10^{-3}\}\times\{10^{-4},5\times10^{-4}\}\)
and, since its candidates share one architecture, selects the eligible
candidate with the lowest mean CRPS. Checkpoints are assessed every
100 epochs up to 2{,}000 for Conditional DDPM and every 500 iterations
up to 7{,}500 for GCDS; the final budget is the median selected
checkpoint of the chosen candidate, rounded to the checkpoint interval.
Formal fits use the full training sample and this budget without
further checkpoint selection. Test outcomes and oracle policies are
not used for tuning.

Both methods use batch size 128. Conditional DDPM uses 200 diffusion
steps, \(v\)-prediction, a cosine noise schedule, EMA decay 0.999, and
50-step deterministic DDIM sampling. GCDS uses three-dimensional
Gaussian noise, generator hidden width 50, discriminator hidden widths
\((50,25)\), and five discriminator updates per generator update.
Selected configurations are listed in Table~\ref{tab:gen-configs}.

\begin{table}[ht]
\centering
\small
\caption{Selected generative configurations. All Conditional DDPM
models use hidden width 64. GCDS learning rates are given as
(generator, discriminator). $^\dagger$Did not pass the preflight check.
--: not evaluated.}
\label{tab:gen-configs}
\begin{tabular}{llcccc}
\toprule
 & & \multicolumn{2}{c}{Conditional DDPM} & \multicolumn{2}{c}{GCDS} \\
\cmidrule(lr){3-4}\cmidrule(lr){5-6}
Setting & \(p\) & LR & Epochs & LR & Iterations \\
\midrule
1 & 10  & \(10^{-4}\)        & 900              & \((10^{-3},\,5{\times}10^{-4})\)      & 500 \\
2 & 10  & \(2{\times}10^{-4}\) & 800            & \((5{\times}10^{-4},\,10^{-4})\)      & 1500 \\
3 & 10  & \(2{\times}10^{-4}\) & 1600\(^\dagger\) & \((5{\times}10^{-4},\,10^{-4})\)    & 2500\(^\dagger\) \\
\midrule
1 & 20  & \(10^{-4}\)        & 1000             & \((5{\times}10^{-4},\,10^{-4})\)      & 1500 \\
1 & 50  & \(2{\times}10^{-4}\) & 300            & \((5{\times}10^{-4},\,10^{-4})\)      & 1000 \\
1 & 100 & \(10^{-4}\)        & 300              & \((10^{-3},\,10^{-4})\)               & 6500\(^\dagger\) \\
1 & 300 & \(10^{-4}\)        & 200              & \((10^{-3},\,10^{-4})\)               & 3000\(^\dagger\) \\
2 & 20  & \(10^{-4}\)        & 700              & \((5{\times}10^{-4},\,10^{-4})\)      & 1000 \\
2 & 300 & \(10^{-4}\)        & 100\(^\dagger\)  & \((10^{-3},\,10^{-4})\)               & 5500\(^\dagger\) \\
4 & 10  & --                 & --               & \((5{\times}10^{-4},\,10^{-4})\)      & 3500 \\
4 & 20  & --                 & --               & \((5{\times}10^{-4},\,10^{-4})\)      & 1000 \\
5 & 10  & --                 & --               & \((5{\times}10^{-4},\,10^{-4})\)      & 4500 \\
5 & 20  & --                 & --               & \((5{\times}10^{-4},\,10^{-4})\)      & 1000 \\
\bottomrule
\end{tabular}
\end{table}

\paragraph{Preflight check.}
Before formal evaluation, each selected configuration was run 20 times
with independent seeds and treatment-stratified 80/20 splits, using no
formal test data. It passed if at least 19 runs completed with finite
validation CRPS and the median skill was nonnegative. This check
assesses execution reliability and predictive quality. Although our protocol
originally required passing it before formal evaluation, we evaluated
all configurations, including those that did not pass
(Table~\ref{tab:gen-configs}), and report all replications without
performance-based exclusion.
In formal evaluation, all Conditional DDPM runs were evaluable. One of 1{,}000 GCDS runs (Setting~2, \(p=300\)) failed a generated-sample collapse check and could not be evaluated, so the GCDS entries in Table~\ref{tab:s2_p300_full} average over 99 replications. Some fitted policies are also
degenerate: in Setting~2 at \(p=300\), DRF's CVaR\(_{0.1}\) rule assigns
treatment \(-1\) to every test subject in all 100 replications
(Err 0.499); Conditional DDPM's mean-targeted rule is constant in 5 of 100 replications in Setting~1 at \(p=300\) and in 48 of 100 in Setting~2 at \(p=300\). All such results are retained in the reported tables.


\clearpage
\newpage
\input{checklist.tex}
\end{document}

%% file: checklist.tex
\section*{NeurIPS Paper Checklist}

\begin{enumerate}

\item {\bf Claims}
    \item[] Question: Do the main claims made in the abstract and introduction accurately reflect the paper's contributions and scope?
    \item[] Answer: \answerYes{} 
    \item[] Justification: The abstract and introduction accurately summarize the paper’s main contributions: the distribution-first framework, the theoretical analysis, and the empirical comparisons. They also reflect the scope of the paper and the criteria considered.
    \item[] Guidelines:
    \begin{itemize}
        \item The answer \answerNA{} means that the abstract and introduction do not include the claims made in the paper.
        \item The abstract and/or introduction should clearly state the claims made, including the contributions made in the paper and important assumptions and limitations. A \answerNo{} or \answerNA{} answer to this question will not be perceived well by the reviewers. 
        \item The claims made should match theoretical and experimental results, and reflect how much the results can be expected to generalize to other settings. 
        \item It is fine to include aspirational goals as motivation as long as it is clear that these goals are not attained by the paper. 
    \end{itemize}

\item {\bf Limitations}
    \item[] Question: Does the paper discuss the limitations of the work performed by the authors?
    \item[] Answer: \answerYes{} 
    \item[] Justification: The paper discusses several limitations, including the current focus on H\"older-continuous treatment criteria, the plug-in nature of the framework, and the dependence of generative-model gains on the outcome regime and target criterion. It also outlines corresponding future directions, such as doubly robust extensions, sharper theory for specific criteria, and broader data settings.
    \item[] Guidelines:
    \begin{itemize}
        \item The answer \answerNA{} means that the paper has no limitation while the answer \answerNo{} means that the paper has limitations, but those are not discussed in the paper. 
        \item The authors are encouraged to create a separate ``Limitations'' section in their paper.
        \item The paper should point out any strong assumptions and how robust the results are to violations of these assumptions (e.g., independence assumptions, noiseless settings, model well-specification, asymptotic approximations only holding locally). The authors should reflect on how these assumptions might be violated in practice and what the implications would be.
        \item The authors should reflect on the scope of the claims made, e.g., if the approach was only tested on a few datasets or with a few runs. In general, empirical results often depend on implicit assumptions, which should be articulated.
        \item The authors should reflect on the factors that influence the performance of the approach. For example, a facial recognition algorithm may perform poorly when image resolution is low or images are taken in low lighting. Or a speech-to-text system might not be used reliably to provide closed captions for online lectures because it fails to handle technical jargon.
        \item The authors should discuss the computational efficiency of the proposed algorithms and how they scale with dataset size.
        \item If applicable, the authors should discuss possible limitations of their approach to address problems of privacy and fairness.
        \item While the authors might fear that complete honesty about limitations might be used by reviewers as grounds for rejection, a worse outcome might be that reviewers discover limitations that aren't acknowledged in the paper. The authors should use their best judgment and recognize that individual actions in favor of transparency play an important role in developing norms that preserve the integrity of the community. Reviewers will be specifically instructed to not penalize honesty concerning limitations.
    \end{itemize}

\item {\bf Theory assumptions and proofs}
    \item[] Question: For each theoretical result, does the paper provide the full set of assumptions and a complete (and correct) proof?
    \item[] Answer: \answerYes{} 
    \item[] Justification: Each theoretical result is stated under explicit assumptions, and full proofs are provided in the appendix. The main text also summarizes the role of the assumptions and the resulting error propagation and regret bounds.
    \item[] Guidelines:
    \begin{itemize}
        \item The answer \answerNA{} means that the paper does not include theoretical results. 
        \item All the theorems, formulas, and proofs in the paper should be numbered and cross-referenced.
        \item All assumptions should be clearly stated or referenced in the statement of any theorems.
        \item The proofs can either appear in the main paper or the supplemental material, but if they appear in the supplemental material, the authors are encouraged to provide a short proof sketch to provide intuition. 
        \item Inversely, any informal proof provided in the core of the paper should be complemented by formal proofs provided in appendix or supplemental material.
        \item Theorems and Lemmas that the proof relies upon should be properly referenced. 
    \end{itemize}

    \item {\bf Experimental result reproducibility}
    \item[] Question: Does the paper fully disclose all the information needed to reproduce the main experimental results of the paper to the extent that it affects the main claims and/or conclusions of the paper (regardless of whether the code and data are provided or not)?
    \item[] Answer: \answerYes{} 
    \item[] Justification: The paper specifies the simulation settings, evaluation criteria, competing methods, and the real-data analysis setup needed to reproduce the main empirical results. It also describes the first-stage estimators and the overall distribution-first pipeline used in each experiment.
    \item[] Guidelines:
    \begin{itemize}
        \item The answer \answerNA{} means that the paper does not include experiments.
        \item If the paper includes experiments, a \answerNo{} answer to this question will not be perceived well by the reviewers: Making the paper reproducible is important, regardless of whether the code and data are provided or not.
        \item If the contribution is a dataset and\slash or model, the authors should describe the steps taken to make their results reproducible or verifiable. 
        \item Depending on the contribution, reproducibility can be accomplished in various ways. For example, if the contribution is a novel architecture, describing the architecture fully might suffice, or if the contribution is a specific model and empirical evaluation, it may be necessary to either make it possible for others to replicate the model with the same dataset, or provide access to the model. In general. releasing code and data is often one good way to accomplish this, but reproducibility can also be provided via detailed instructions for how to replicate the results, access to a hosted model (e.g., in the case of a large language model), releasing of a model checkpoint, or other means that are appropriate to the research performed.
        \item While NeurIPS does not require releasing code, the conference does require all submissions to provide some reasonable avenue for reproducibility, which may depend on the nature of the contribution. For example
        \begin{enumerate}
            \item If the contribution is primarily a new algorithm, the paper should make it clear how to reproduce that algorithm.
            \item If the contribution is primarily a new model architecture, the paper should describe the architecture clearly and fully.
            \item If the contribution is a new model (e.g., a large language model), then there should either be a way to access this model for reproducing the results or a way to reproduce the model (e.g., with an open-source dataset or instructions for how to construct the dataset).
            \item We recognize that reproducibility may be tricky in some cases, in which case authors are welcome to describe the particular way they provide for reproducibility. In the case of closed-source models, it may be that access to the model is limited in some way (e.g., to registered users), but it should be possible for other researchers to have some path to reproducing or verifying the results.
        \end{enumerate}
    \end{itemize}

\item {\bf Open access to data and code}
    \item[] Question: Does the paper provide open access to the data and code, with sufficient instructions to faithfully reproduce the main experimental results, as described in supplemental material?
    \item[] Answer: \answerYes{} 
    \item[] Justification: The paper provides access to the code and specifies the data source, simulation data-generation process, sample settings, and experimental configurations for the main results.
    \item[] Guidelines:
    \begin{itemize}
        \item The answer \answerNA{} means that paper does not include experiments requiring code.
        \item Please see the NeurIPS code and data submission guidelines (\url{https://neurips.cc/public/guides/CodeSubmissionPolicy}) for more details.
        \item While we encourage the release of code and data, we understand that this might not be possible, so \answerNo{} is an acceptable answer. Papers cannot be rejected simply for not including code, unless this is central to the contribution (e.g., for a new open-source benchmark).
        \item The instructions should contain the exact command and environment needed to run to reproduce the results. See the NeurIPS code and data submission guidelines (\url{https://neurips.cc/public/guides/CodeSubmissionPolicy}) for more details.
        \item The authors should provide instructions on data access and preparation, including how to access the raw data, preprocessed data, intermediate data, and generated data, etc.
        \item The authors should provide scripts to reproduce all experimental results for the new proposed method and baselines. If only a subset of experiments are reproducible, they should state which ones are omitted from the script and why.
        \item At submission time, to preserve anonymity, the authors should release anonymized versions (if applicable).
        \item Providing as much information as possible in supplemental material (appended to the paper) is recommended, but including URLs to data and code is permitted.
    \end{itemize}

\item {\bf Experimental setting/details}
    \item[] Question: Does the paper specify all the training and test details (e.g., data splits, hyperparameters, how they were chosen, type of optimizer) necessary to understand the results?
    \item[] Answer: \answerYes{} 
    \item[] Justification: The paper specifies the simulation settings, sample sizes, data splits, evaluation criteria, and the main implementation choices for each experiment. These details are sufficient to understand how the results were obtained.
    \item[] Guidelines:
    \begin{itemize}
        \item The answer \answerNA{} means that the paper does not include experiments.
        \item The experimental setting should be presented in the core of the paper to a level of detail that is necessary to appreciate the results and make sense of them.
        \item The full details can be provided either with the code, in appendix, or as supplemental material.
    \end{itemize}

\item {\bf Experiment statistical significance}
    \item[] Question: Does the paper report error bars suitably and correctly defined or other appropriate information about the statistical significance of the experiments?
    \item[] Answer: \answerYes{} 
    \item[] Justification: Simulation results are reported with Monte Carlo standard errors across replications. ACTG175 utilities and treatment rates are reported with standard deviations across 200 paired splits, and runtimes with medians and interquartile ranges; these descriptive summaries do not establish statistical significance.
    \item[] Guidelines:
    \begin{itemize}
        \item The answer \answerNA{} means that the paper does not include experiments.
        \item The authors should answer \answerYes{} if the results are accompanied by error bars, confidence intervals, or statistical significance tests, at least for the experiments that support the main claims of the paper.
        \item The factors of variability that the error bars are capturing should be clearly stated (for example, train/test split, initialization, random drawing of some parameter, or overall run with given experimental conditions).
        \item The method for calculating the error bars should be explained (closed form formula, call to a library function, bootstrap, etc.)
        \item The assumptions made should be given (e.g., Normally distributed errors).
        \item It should be clear whether the error bar is the standard deviation or the standard error of the mean.
        \item It is OK to report 1-sigma error bars, but one should state it. The authors should preferably report a 2-sigma error bar than state that they have a 96\% CI, if the hypothesis of Normality of errors is not verified.
        \item For asymmetric distributions, the authors should be careful not to show in tables or figures symmetric error bars that would yield results that are out of range (e.g., negative error rates).
        \item If error bars are reported in tables or plots, the authors should explain in the text how they were calculated and reference the corresponding figures or tables in the text.
    \end{itemize}

\item {\bf Experiments compute resources}
    \item[] Question: For each experiment, does the paper provide sufficient information on the computer resources (type of compute workers, memory, time of execution) needed to reproduce the experiments?
    \item[] Answer: \answerYes{} 
    \item[] Justification: The appendix describes the compute environment used for the experiments, including the cluster setting, CPU/GPU usage, software environment, and the main factors affecting computational cost. It also clarifies that the reported compute corresponds to the experiments included in the paper.
    \item[] Guidelines:
    \begin{itemize}
        \item The answer \answerNA{} means that the paper does not include experiments.
        \item The paper should indicate the type of compute workers CPU or GPU, internal cluster, or cloud provider, including relevant memory and storage.
        \item The paper should provide the amount of compute required for each of the individual experimental runs as well as estimate the total compute. 
        \item The paper should disclose whether the full research project required more compute than the experiments reported in the paper (e.g., preliminary or failed experiments that didn't make it into the paper). 
    \end{itemize}
    
\item {\bf Code of ethics}
    \item[] Question: Does the research conducted in the paper conform, in every respect, with the NeurIPS Code of Ethics \url{https://neurips.cc/public/EthicsGuidelines}?
    \item[] Answer: \answerYes{} 
    \item[] Justification: We have reviewed the NeurIPS Code of Ethics and are not aware of any aspect of this work that violates it. The paper uses standard simulation studies and a public clinical trial dataset, with no human-subject interaction or other special ethical concerns beyond those discussed in the paper.
    \item[] Guidelines:
    \begin{itemize}
        \item The answer \answerNA{} means that the authors have not reviewed the NeurIPS Code of Ethics.
        \item If the authors answer \answerNo, they should explain the special circumstances that require a deviation from the Code of Ethics.
        \item The authors should make sure to preserve anonymity (e.g., if there is a special consideration due to laws or regulations in their jurisdiction).
    \end{itemize}

\item {\bf Broader impacts}
    \item[] Question: Does the paper discuss both potential positive societal impacts and negative societal impacts of the work performed?
    \item[] Answer: \answerNo{} 
    \item[] Justification: The paper is primarily a foundational methodological study, and we do not identify an immediate direct path to societal harm from the proposed framework itself. The paper discusses its potential positive relevance to precision medicine.
    \item[] Guidelines:
    \begin{itemize}
        \item The answer \answerNA{} means that there is no societal impact of the work performed.
        \item If the authors answer \answerNA{} or \answerNo, they should explain why their work has no societal impact or why the paper does not address societal impact.
        \item Examples of negative societal impacts include potential malicious or unintended uses (e.g., disinformation, generating fake profiles, surveillance), fairness considerations (e.g., deployment of technologies that could make decisions that unfairly impact specific groups), privacy considerations, and security considerations.
        \item The conference expects that many papers will be foundational research and not tied to particular applications, let alone deployments. However, if there is a direct path to any negative applications, the authors should point it out. For example, it is legitimate to point out that an improvement in the quality of generative models could be used to generate Deepfakes for disinformation. On the other hand, it is not needed to point out that a generic algorithm for optimizing neural networks could enable people to train models that generate Deepfakes faster.
        \item The authors should consider possible harms that could arise when the technology is being used as intended and functioning correctly, harms that could arise when the technology is being used as intended but gives incorrect results, and harms following from (intentional or unintentional) misuse of the technology.
        \item If there are negative societal impacts, the authors could also discuss possible mitigation strategies (e.g., gated release of models, providing defenses in addition to attacks, mechanisms for monitoring misuse, mechanisms to monitor how a system learns from feedback over time, improving the efficiency and accessibility of ML).
    \end{itemize}
    
\item {\bf Safeguards}
    \item[] Question: Does the paper describe safeguards that have been put in place for responsible release of data or models that have a high risk for misuse (e.g., pre-trained language models, image generators, or scraped datasets)?
    \item[] Answer: \answerNA{} 
    \item[] Justification: The paper does not release high-risk models, scraped datasets, or other resources with an obvious dual-use or misuse risk. It is a methodological study based on simulations and an existing clinical trial dataset.
    \item[] Guidelines:
    \begin{itemize}
        \item The answer \answerNA{} means that the paper poses no such risks.
        \item Released models that have a high risk for misuse or dual-use should be released with necessary safeguards to allow for controlled use of the model, for example by requiring that users adhere to usage guidelines or restrictions to access the model or implementing safety filters. 
        \item Datasets that have been scraped from the Internet could pose safety risks. The authors should describe how they avoided releasing unsafe images.
        \item We recognize that providing effective safeguards is challenging, and many papers do not require this, but we encourage authors to take this into account and make a best faith effort.
    \end{itemize}

\item {\bf Licenses for existing assets}
    \item[] Question: Are the creators or original owners of assets (e.g., code, data, models), used in the paper, properly credited and are the license and terms of use explicitly mentioned and properly respected?
    \item[] Answer: \answerYes{} 
    \item[] Justification:  The paper credits the original sources of the ACTG175 dataset and the existing methods used for comparison through citations in the main text. We use publicly available data and standard research baselines, and no third-party assets are redistributed as part of this work.
    \item[] Guidelines:
    \begin{itemize}
        \item The answer \answerNA{} means that the paper does not use existing assets.
        \item The authors should cite the original paper that produced the code package or dataset.
        \item The authors should state which version of the asset is used and, if possible, include a URL.
        \item The name of the license (e.g., CC-BY 4.0) should be included for each asset.
        \item For scraped data from a particular source (e.g., website), the copyright and terms of service of that source should be provided.
        \item If assets are released, the license, copyright information, and terms of use in the package should be provided. For popular datasets, \url{paperswithcode.com/datasets} has curated licenses for some datasets. Their licensing guide can help determine the license of a dataset.
        \item For existing datasets that are re-packaged, both the original license and the license of the derived asset (if it has changed) should be provided.
        \item If this information is not available online, the authors are encouraged to reach out to the asset's creators.
    \end{itemize}

\item {\bf New assets}
    \item[] Question: Are new assets introduced in the paper well documented and is the documentation provided alongside the assets?
    \item[] Answer: \answerNA{} 
    \item[] Justification: The paper does not release new datasets, models, or other standalone assets. It is a methodological study based on simulations and existing data sources.
    \item[] Guidelines:
    \begin{itemize}
        \item The answer \answerNA{} means that the paper does not release new assets.
        \item Researchers should communicate the details of the dataset\slash code\slash model as part of their submissions via structured templates. This includes details about training, license, limitations, etc. 
        \item The paper should discuss whether and how consent was obtained from people whose asset is used.
        \item At submission time, remember to anonymize your assets (if applicable). You can either create an anonymized URL or include an anonymized zip file.
    \end{itemize}

\item {\bf Crowdsourcing and research with human subjects}
    \item[] Question: For crowdsourcing experiments and research with human subjects, does the paper include the full text of instructions given to participants and screenshots, if applicable, as well as details about compensation (if any)? 
    \item[] Answer: \answerNA{} 
    \item[] Justification: The paper does not involve crowdsourcing experiments or research with human subjects. It uses simulation studies and an existing clinical trial dataset.
    \item[] Guidelines:
    \begin{itemize}
        \item The answer \answerNA{} means that the paper does not involve crowdsourcing nor research with human subjects.
        \item Including this information in the supplemental material is fine, but if the main contribution of the paper involves human subjects, then as much detail as possible should be included in the main paper. 
        \item According to the NeurIPS Code of Ethics, workers involved in data collection, curation, or other labor should be paid at least the minimum wage in the country of the data collector. 
    \end{itemize}

\item {\bf Institutional review board (IRB) approvals or equivalent for research with human subjects}
    \item[] Question: Does the paper describe potential risks incurred by study participants, whether such risks were disclosed to the subjects, and whether Institutional Review Board (IRB) approvals (or an equivalent approval/review based on the requirements of your country or institution) were obtained?
    \item[] Answer: \answerNA{} 
    \item[] Justification: The paper does not involve crowdsourcing or prospective research with human subjects. It uses simulation studies and an existing de-identified clinical trial dataset.
    \item[] Guidelines:
    \begin{itemize}
        \item The answer \answerNA{} means that the paper does not involve crowdsourcing nor research with human subjects.
        \item Depending on the country in which research is conducted, IRB approval (or equivalent) may be required for any human subjects research. If you obtained IRB approval, you should clearly state this in the paper. 
        \item We recognize that the procedures for this may vary significantly between institutions and locations, and we expect authors to adhere to the NeurIPS Code of Ethics and the guidelines for their institution. 
        \item For initial submissions, do not include any information that would break anonymity (if applicable), such as the institution conducting the review.
    \end{itemize}

\item {\bf Declaration of LLM usage}
    \item[] Question: Does the paper describe the usage of LLMs if it is an important, original, or non-standard component of the core methods in this research? Note that if the LLM is used only for writing, editing, or formatting purposes and does \emph{not} impact the core methodology, scientific rigor, or originality of the research, declaration is not required.
    \item[] Answer: \answerNA{} 
    \item[] Justification: The core methodology of the paper does not use LLMs as an important, original, or non-standard component. Any LLM use was limited to writing or editing assistance and does not affect the scientific contribution of the work.
    \item[] Guidelines:
    \begin{itemize}
        \item The answer \answerNA{} means that the core method development in this research does not involve LLMs as any important, original, or non-standard components.
        \item Please refer to our LLM policy in the NeurIPS handbook for what should or should not be described.
    \end{itemize}

\end{enumerate}

%% file: ref.bib
@article{qian2011performance,
  title   = {Performance Guarantees for Individualized Treatment Rules},
  author  = {Qian, Min and Murphy, Susan A.},
  journal = {The Annals of Statistics},
  volume  = {39},
  number  = {2},
  pages   = {1180--1210},
  year    = {2011}
}

@article{zhao2012estimating,
  title   = {Estimating Individualized Treatment Rules Using Outcome Weighted Learning},
  author  = {Zhao, Yingqi and Zeng, Donglin and Rush, A. John and Kosorok, Michael R.},
  journal = {Journal of the American Statistical Association},
  volume  = {107},
  number  = {499},
  pages   = {1106--1118},
  year    = {2012}
}

@article{kosorok2019precision,
  title   = {Precision Medicine},
  author  = {Kosorok, Michael R. and Laber, Eric B.},
  journal = {Annual Review of Statistics and Its Application},
  volume  = {6},
  pages   = {263--286},
  year    = {2019}
}

@article{qi2023robustness,
  title   = {On Robustness of Individualized Decision Rules},
  author  = {Qi, Zhengling and Pang, Jong-Shi and Liu, Yufeng},
  journal = {Journal of the American Statistical Association},
  volume  = {118},
  number  = {543},
  pages   = {2143--2157},
  year    = {2023}
}

@article{rockafellar2000optimization,
  title   = {Optimization of Conditional Value-at-Risk},
  author  = {Rockafellar, R. Tyrrell and Uryasev, Stanislav},
  journal = {Journal of Risk},
  volume  = {2},
  number  = {3},
  pages   = {21--41},
  year    = {2000}
}

@article{tversky1992advances,
  title   = {Advances in Prospect Theory: Cumulative Representation of Uncertainty},
  author  = {Tversky, Amos and Kahneman, Daniel},
  journal = {Journal of Risk and Uncertainty},
  volume  = {5},
  number  = {4},
  pages   = {297--323},
  year    = {1992}
}

@article{breiman2001random,
  title   = {Random Forests},
  author  = {Breiman, Leo},
  journal = {Machine Learning},
  volume  = {45},
  number  = {1},
  pages   = {5--32},
  year    = {2001}
}

@article{athey2019generalized,
  title   = {Generalized Random Forests},
  author  = {Athey, Susan and Tibshirani, Julie and Wager, Stefan},
  journal = {The Annals of Statistics},
  volume  = {47},
  number  = {2},
  pages   = {1148--1178},
  year    = {2019}
}

@article{cevid2022distributional,
  title   = {Distributional Random Forests: Heterogeneity Adjustment and Multivariate Distributional Regression},
  author = {\'{C}evid, Domagoj and Michel, Lukas and N{\"a}f, Jeffrey and B{\"u}hlmann, Peter and Meinshausen, Nicolai},
  journal = {Journal of Machine Learning Research},
  volume  = {23},
  number  = {333},
  pages   = {1--79},
  year    = {2022}
}

@article{meinshausen2006quantile,
  title   = {Quantile Regression Forests},
  author  = {Meinshausen, Nicolai},
  journal = {Journal of Machine Learning Research},
  volume  = {7},
  number  = {35},
  pages   = {983--999},
  year    = {2006}
}

@article{hall1999nonparametric,
  title   = {Methods for Estimating a Conditional Distribution Function},
  author  = {Hall, Peter and Wolff, Rodney C. L. and Yao, Qiwei},
  journal = {Journal of the American Statistical Association},
  volume  = {94},
  number  = {445},
  pages   = {154--163},
  year    = {1999}
}

@inproceedings{ho2020denoising,
  title     = {Denoising Diffusion Probabilistic Models},
  author    = {Ho, Jonathan and Jain, Ajay and Abbeel, Pieter},
  booktitle = {Advances in Neural Information Processing Systems},
  volume    = {33},
  pages     = {6840--6851},
  year      = {2020}
}

@inproceedings{song2021scorebased,
  title     = {Score-Based Generative Modeling through Stochastic Differential Equations},
  author    = {Song, Yang and Sohl-Dickstein, Jascha and Kingma, Diederik P. and Kumar, Abhishek and Ermon, Stefano and Poole, Ben},
  booktitle = {International Conference on Learning Representations},
  year      = {2021}
}

@inproceedings{kotelnikov2023tabddpm,
  title     = {{TabDDPM}: Modelling Tabular Data with Diffusion Models},
  author    = {Kotelnikov, Akim and Baranchuk, Dmitry and Rubachev, Ivan and Babenko, Artem},
  booktitle = {International Conference on Machine Learning},
  pages     = {17564--17579},
  volume = {202},
  year      = {2023}
}

@inproceedings{
ma2024diffpo,
title={Diff{PO}: A causal diffusion model for learning distributions of potential outcomes},
author={Yuchen Ma and Valentyn Melnychuk and Jonas Schweisthal and Stefan Feuerriegel},
booktitle={Advances in Neural Information Processing Systems},
  volume={37},
  pages={43663--43692},
year={2024}
}

@inproceedings{
tang2025conditional,
title={Conditional Diffusion Models are Minimax-Optimal and Manifold-Adaptive for Conditional Distribution Estimation},
author={Rong Tang and Lizhen Lin and Yun Yang},
booktitle={International Conference on Learning Representations},
year={2025}
}

@inproceedings{goodfellow2014generative,
  title     = {Generative Adversarial Nets},
  author    = {Goodfellow, Ian and Pouget-Abadie, Jean and Mirza, Mehdi and Xu, Bing and Warde-Farley, David and Ozair, Sherjil and Courville, Aaron and Bengio, Yoshua},
  booktitle = {Advances in Neural Information Processing Systems},
  volume    = {27},
  year      = {2014}
}

@article{kingma2013auto,
  title   = {Auto-Encoding Variational Bayes},
  author  = {Kingma, Diederik P. and Welling, Max},
  journal = {arXiv preprint arXiv:1312.6114},
  year    = {2013}
}

@article{papamakarios2021normalizing,
  title   = {Normalizing Flows for Probabilistic Modeling and Inference},
  author  = {Papamakarios, George and Nalisnick, Eric and Rezende, Danilo Jimenez and Mohamed, Shakir and Lakshminarayanan, Balaji},
  journal = {Journal of Machine Learning Research},
  volume  = {22},
  number  = {57},
  pages   = {1--64},
  year    = {2021}
}

@inproceedings{huang2021offpolicy,
  title     = {Off-Policy Risk Assessment in Contextual Bandits},
  author    = {Huang, Audrey and Leqi, Liu and Lipton, Zachary C. and Azizzadenesheli, Kamyar},
  booktitle = {Advances in Neural Information Processing Systems},
  volume    = {34},
  pages     = {23714--23726},
  year      = {2021}
}

@inproceedings{huang2022offpolicy,
  title     = {Off-Policy Risk Assessment for Markov Decision Processes},
  author    = {Huang, Audrey and Leqi, Liu and Lipton, Zachary C. and Azizzadenesheli, Kamyar},
  booktitle = {International Conference on Artificial Intelligence and Statistics},
  pages     = {5022--5050},
  volume = 	 {151},
  year      = {2022}
}

@article{zhou2017residual,
  title={Residual Weighted Learning for Estimating Individualized Treatment Rules},
  author={Zhou, Xin and Mayer-Hamblett, Nicole and Khan, Umer and Kosorok, Michael R.},
  journal={Journal of the American Statistical Association},
  volume={112},
  number={517},
  pages={169--187},
  year={2017}
}

@article{qi2020multi,
  title={Multi-Armed Angle-Based Direct Learning for Estimating Optimal Individualized Treatment Rules with Various Outcomes},
  author={Qi, Zhengling and Liu, Dacheng and Fu, Haoda and Liu, Yufeng},
  journal={Journal of the American Statistical Association},
  volume={115},
  number={530},
  pages={678--691},
  year={2020}
}

@article{zhou2023deep,
  title   = {A Deep Generative Approach to Conditional Sampling},
  author  = {Zhou, Xingyu and Jiao, Yuling and Liu, Jin and Huang, Jian},
  journal = {Journal of the American Statistical Association},
  volume  = {118},
  number  = {543},
  pages   = {1837--1848},
  year    = {2023}
}

@article{murphy2003optimal,
  title   = {Optimal Dynamic Treatment Regimes},
  author  = {Murphy, Susan A.},
  journal = {Journal of the Royal Statistical Society: Series B (Statistical Methodology)},
  volume  = {65},
  number  = {2},
  pages   = {331--355},
  year    = {2003}
}

@article{wang2018quantile,
  title   = {Quantile-Optimal Treatment Regimes},
  author  = {Wang, Lan and Zhou, Yu and Song, Rui and Sherwood, Ben},
  journal = {Journal of the American Statistical Association},
  volume  = {113},
  number  = {523},
  pages   = {1243--1254},
  year    = {2018}
}

@article{xiao2019robust,
  title   = {Robust Regression for Optimal Individualized Treatment Rules},
  author  = {Xiao, Wei and Zhang, Hao Helen and Lu, Wenbin},
  journal = {Statistics in Medicine},
  volume  = {38},
  number  = {11},
  pages   = {2059--2073},
  year    = {2019}
}

@article{audibert2007fast,
  title={Fast learning rates for plug-in classifiers},
  author={Audibert, Jean-Yves and Tsybakov, Alexandre B.},
  journal={The Annals of Statistics},
  volume={35},
  number={2},
  pages={608--633},
  year={2007}
}

@article{luedtke2017faster,
  title={Faster rates for policy learning},
  author={Luedtke, Alexander and Chambaz, Antoine},
  journal={arXiv preprint arXiv:1704.06431},
  year={2017}
}

@article{hammer1996trial,
  title={A trial comparing nucleoside monotherapy with combination therapy in HIV-infected adults with CD4 cell counts from 200 to 500 per cubic millimeter},
  author={Hammer, Scott M and Katzenstein, David A and Hughes, Michael D and Gundacker, Holly and Schooley, Robert T and Haubrich, Richard H and Henry, Keith WK and Lederman, Michael M and Phair, John P and Niu, Manette and others},
  journal={New England Journal of Medicine},
  volume={335},
  number={15},
  pages={1081--1090},
  year={1996}
}

@article{attema2013prospect,
  title   = {Prospect theory in the health domain: A quantitative assessment},
  author  = {Attema, Arthur E. and Brouwer, Werner B. F. and {l'Haridon}, Olivier},
  journal = {Journal of Health Economics},
  volume  = {32},
  number  = {6},
  pages   = {1057--1065},
  year    = {2013}
}

@article{lipman2019corrective,
  title   = {The Corrective Approach: Policy Implications of Recent Developments in {QALY} Measurement Based on Prospect Theory},
  author  = {Lipman, Stefan A. and Brouwer, Werner B. F. and Attema, Arthur E.},
  journal = {Value in Health},
  volume  = {22},
  number  = {7},
  pages   = {816--821},
  year    = {2019}
}

@article{linn2017outcome,
  author  = {Linn, Kristin A. and Laber, Eric B. and Stefanski, Leonard A.},
  title   = {Interactive {Q}-learning for quantiles},
  journal = {Journal of the American Statistical Association},
  year    = {2017},
  volume  = {112},
  number  = {518},
  pages   = {638--649}
}

@article{prashanth2022risk,
  title={Risk-sensitive reinforcement learning via policy gradient search},
  author={Prashanth, L. A. and Fu, Michael C.},
  journal={Foundations and Trends in Machine Learning},
  volume={15},
  number={5},
  pages={537--693},
  year={2022}
}

@inproceedings{barakat2025global,
  title     = {On the Global Optimality of Policy Gradient Methods in General Utility Reinforcement Learning},
  author    = {Barakat, Anas and Chakraborty, Souradip and Yu, Peihong and Tokekar, Pratap and Bedi, Amrit Singh},
  booktitle = {Advances in Neural Information Processing Systems},
  year      = {2025}
}

@inproceedings{smith2025risk,
  title={Risk-Sensitive Theory of Mind: Coordinating with Agents of Unknown Bias using Cumulative Prospect Theory},
  author={Smith, Mason O and Zhang, Wenlong},
  booktitle={International Conference on Machine Learning},
    pages = 	 {55999--56017},
  year = 	 {2025},
  volume = 	 {267}
}

@article{naf2024causal,
  title={{Causal-DRF}: Conditional kernel treatment effect estimation using distributional random forest},
  author={N{\"a}f, Jeffrey and Park, Junhyung and Susmann, Herbert},
  journal={arXiv preprint arXiv:2411.08778},
  year={2024}
}

@article{liu2021density,
  title={Density estimation using deep generative neural networks},
  author={Liu, Qiao and Xu, Jiaze and Jiang, Rui and Wong, Wing Hung},
  journal={Proceedings of the National Academy of Sciences},
  volume={118},
  number={15},
  pages={e2101344118},
  year={2021}
}

@article{rubin1974estimating,
  title={Estimating causal effects of treatments in randomized and nonrandomized studies.},
  author={Rubin, Donald B},
  journal={Journal of Educational Psychology},
  volume={66},
  number={5},
  pages={688--701},
  year={1974}
}

@book{vanderVaart1996weak,
  title     = {Weak Convergence and Empirical Processes: With Applications to Statistics},
  author    = {Van Der Vaart, Aad W and Wellner, Jon A},
  publisher = {Springer},
  pages={16--28},
  year = {1996}
}

@article{einmahl2005uniform,
  title   = {Uniform in Bandwidth Consistency of Kernel-Type Function
             Estimators},
  author  = {Einmahl, Uwe and Mason, David M.},
  journal = {The Annals of Statistics},
  volume  = {33},
  number  = {3},
  pages   = {1380--1403},
  year    = {2005}
}
